\documentclass[10pt,twocolumn,letterpaper]{article}
\usepackage{cvpr}
\usepackage{times}
\usepackage{epsfig}
\usepackage{graphicx}
\usepackage{amsmath}
\usepackage{amssymb}
\usepackage{amsfonts}
\usepackage{bm}
\usepackage[dvipsnames]{xcolor}
\usepackage{booktabs}
\usepackage{subfig}
\usepackage{xfrac}
\usepackage{threeparttable}
\usepackage{multirow}
\usepackage{rotating}
\usepackage{changepage}
\usepackage{makecell}
\usepackage{floatrow}
\newcolumntype{P}[1]{>{\centering\arraybackslash}p{#1}}
\usepackage{lineno}
\modulolinenumbers[5]
\usepackage[breaklinks=true,bookmarks=false]{hyperref}
\cvprfinalcopy 
\def\cvprPaperID{6246} 

\usepackage{atbegshi}
\AtBeginDocument{\AtBeginShipoutNext{\AtBeginShipoutDiscard}}
\begin{document}
\let\cleardoublepage\clearpage
\title{A deep learning framework for quality assessment and restoration in video endoscopy}
{\author{Sharib Ali\thanks{Corresponding author: Sharib Ali, sharib.ali@eng.ox.ac.uk}, Felix Zhou, Adam Bailey, Barbara Braden, James East, Xin Lu and Jens Rittscher}\thanks{The  research  was  supported by  the  National  Institute  for  Health  Research  (NIHR)  Oxford  Biomedical  Research  Centre  (BRC).  The  views  expressed  are  those  of  the  authors  and  not  necessarily  those  of  the  NHS,  the NIHR or the Department of Health.}\thanks{S. Ali and J. Rittscher are with Institute of Biomedical Engineering, Department of Engineering Science, University of Oxford, Oxford, UK.  S. Ali is supported by the NIHR Oxford BRC and J. Rittscher is supported through the EPSRC funded Seebibyte programme (EP/M013774/1).}\thanks{F. Zhou and Xin Lu are with Ludwig Institute for Cancer Research, University of Oxford, Oxford, UK} \thanks{A. Bailey, B. Braden  and J. East are with Translational Gastroenterology Unit, Oxford University Hospitals NHS Foundation Trust, Oxford, UK}
}
\maketitle
\begin{abstract}
Endoscopy is a routine imaging technique used for both diagnosis and minimally invasive surgical treatment. Artifacts such as motion blur, bubbles, specular reflections, floating objects and pixel saturation impede the visual interpretation and the automated analysis of endoscopy videos. Given the widespread use of endoscopy in different clinical applications, we contend that the robust and reliable identification of such artifacts and the automated restoration of corrupted video frames is a fundamental medical imaging problem. Existing state-of-the-art methods only deal with the detection and restoration of selected artifacts. However, typically endoscopy videos contain numerous artifacts which motivates to establish a comprehensive solution.  

We propose a fully automatic framework that can: 1) detect and classify six different primary artifacts, 2) provide a quality score for each frame and 3) restore mildly corrupted frames. To detect different artifacts our framework exploits fast multi-scale, single stage convolutional neural network detector. We introduce a quality metric to assess frame quality and predict image restoration success. Generative adversarial networks with carefully chosen regularization are finally used to restore corrupted frames. 
 
Our detector yields the highest mean average precision (mAP at 5\% threshold) of 49.0 and the lowest computational time of 88 ms allowing for accurate real-time processing. Our restoration models for blind deblurring, saturation correction and inpainting demonstrate significant improvements over previous methods. On a set of 10 test videos we show that our approach preserves an average of 68.7\% which is 25\% more frames than that retained from the raw videos.
\end{abstract}
%
\section{Introduction}
\noindent{Originally} used to image the esophagus, stomach and colon, miniaturization of hardware and improvement of imaging sensors now enable endoscopy of the ear, nose, throat, heart, urinary tract, joints, and abdomen. Common to these endoscopy applications, presence of different imaging artifacts pose significant challenges in monitoring disease progression. The camera in the endoscope is embedded in a long flexible tube. Any small hand motion can cause severe motion artifacts in recorded videos. The light, required for illumination, can interact with tissue and surrounding fluid generating very bright pixel areas ({\it either due to specularity or pixel saturation}). Different viewing angles and occlusions can result in contrast issues due to underexposure. Additionally, similar to any other complex real-world imaging applications, visual clutters of debris, liquid, bubbles, etc., can limit the visual understanding of the underlying tissue. In this study, we are thus considering the following artifacts: specular reflections, pixel saturation, motion blur, contrast and undesired visual clutter. Not only do such artifacts occlude the tissue/organ of interest during diagnosis and treatment, they also adversely affect any computer assisted endoscopy methods (e.g., video mosaicking for follow-ups and archiving, video-frame retrieval for reporting etc.). 

Chikkerur {\it et al.} \cite{Chikkerur:ITB11} and Menor and colleagues \cite{MENOR20161551} have studied video frame quality assessment methods. While they introduce and review very useful global video quality metrics; neither information regarding the cause of frame quality degradation nor the degraded regions could be identified for frame restoration. In general, utilizing these quality scores~\cite{Chikkerur:ITB11}-\!\!\cite{MENOR20161551} only allows for the removal of frames corrupted with artifacts without considering the severity of each artifact type. Such simple removal of corrupted frames can severely reduce the information content of videos and affect their overall temporal smoothness. One adverse effect of this, for example, can be on mosaicking methods that require at least $60\%$ overlap in successive temporal frames to succeed~\cite{ALI:16PR}. Artifacts are thus the primary obstacles in developing effective and reliable computer assisted endoscopy tools. The precise identification, classification and -if possible- restoration are critical to perform a downstream analysis of the video data.

Detecting multiple artifacts and providing adequate restoration is highly challenging. To date, most research groups have studied only specific artifacts in endoscopic imaging~\cite{deblurringEndoscopy,stehle2006removal,tchoulack2008video,akbari2018adaptive}. For example, deblurring of wireless capsule endoscopy images utilizing a total variational (TV) approach was proposed in~\cite{deblurringEndoscopy}. TV-based de-blurring is however parameter sensitive and requires geometrical features to perform well. Endoscopic images have very sparse features and lack geometrically prominent structures. Both hand-crafted features~\cite{stehle2006removal,tchoulack2008video,akbari2018adaptive,Mohammed:18JIM} and neural networks~\cite{Funke:18SPIE} have been used to restore specular reflections. A major drawback of these existing  restoration techniques is that heuristically chosen image intensities are compared with neighboring (local) image pixels. In general, both local and global information is required for realistic frame restoration. One common limitation of almost all the methods is that they only address one particular artifact class, while naturally various different effects corrupt endoscopy videos. For example, both `specularities' and a water `bubble' can be present in the same frame. Endoscopists also dynamically switch between different modalities during acquisition (e.g., normal brightfield (BF), acetic acid, narrow band imaging (NBI) or fluorescence light (FL)) to better highlight specific pathological features. Finally, inter-patient variation is significant even when viewed under the same modality. Existing methods fail to adequately address all of these challenges. In addition to addressing one type of imaging artifact, only one imaging modality and a single patient video sequence are considered in most of endoscopy-based image analysis literature~\cite{stehle2006removal,tchoulack2008video,akbari2018adaptive,Funke:18SPIE,Mohammed:18JIM}. The use of small size data sets in these studies also raises concern regarding method generalization to image variabilities often present in endoscopic data. For example, in~\cite{akbari2018adaptive} only 100 randomly selected images were used to train the Support Vector Machine (SVM) for detecting specular regions. 

In this paper, we propose a systematic and comprehensive approach to the problem. Our framework addresses the precise detection and localisation of six different artifacts and introduces artifact type specific restoration of mildly affected frames. Unlike previous methods~\cite{deblurringEndoscopy,stehle2006removal,tchoulack2008video,akbari2018adaptive,Funke:18SPIE,Mohammed:18JIM} that require manual adjustment of parameter settings or the use of hand-crafted features only suitable for specific artifacts, we propose to use~\textit{multiple class artifact} detection and restoration methods utilizing \textit{multi-patient} and \textit{multi-modal} video frames. Such an approach decreases false classification rate and better generalizes both detection and frame restoration methods. Reliable multi-class detection is made possible through a multi-scale and deep convolutional neural network based object detection which can efficiently generalize multi-class artifact detection in cross patients and cross modality present in endoscopic data. Realistic frame restoration is achieved using Generative adversarial networks (GANs,~\cite{Goodfellow:NIPS14}). While our work is built on these approaches, substantial additional work has been necessary to avoid the introduction of additional artifacts and disruptions to the the overall visual coherence. In order to achieve this we introduce artifact type dependent regularization. A novel edge-based regularization and restoration is proposed for de-blurring. Restoration of large saturated pixel areas using GAN is of its first kind and have never been addressed in literature. In order to handle the color shift we introduce a novel  color-transfer technique in this scheme. For tackling with artifacts like debris, bubbles, and other misc. artifacts we apply a complete restoration of pixels based on  global contextual regularization scheme. Additionally, we condition each of GAN model with prior image information. We demonstrate that such carefully chosen models can lead to both high-quality and very realistic frame restoration. 

To our knowledge, this is the first attempt to propose a systematic and general approach that handles cross modality, and inter-patient video data for both automatic detection of multiple artifacts present in endoscopic data and their subsequent restorations. We use 7 unique  patient videos (gastroesophageal, selected from large cohort of 200 videos) for training and 10 different videos for extensive validations. Our experiments utilizing well-established video quality assessment metrics illustrate the effectiveness of our approaches. In addition, quality of the restored frames has also been evaluated by two experienced endoscopists. A score based on visual improvement, importance, and presence or absence of any artificially introduced artifact in our restored frames were provided by these experts.

The remainder of this article is organized as follows. In Section 2, we introduce our endoscopy data set for artifact detection. Section 3 details our proposed approaches for artifact detection and endoscopic video frame restoration. In this section we also review closely related works associated with each process. In Section 4, we present experiments and results for each step of our framework to show the efficacy of individual methods. Finally, in Section 5 we conclude the paper and outline directions for future work.
\section{Material}{\label{sec:materials}}
\begin{figure}[t!]
\centering
\begin{minipage}[b]{0.48\linewidth}
\includegraphics[width=\linewidth]{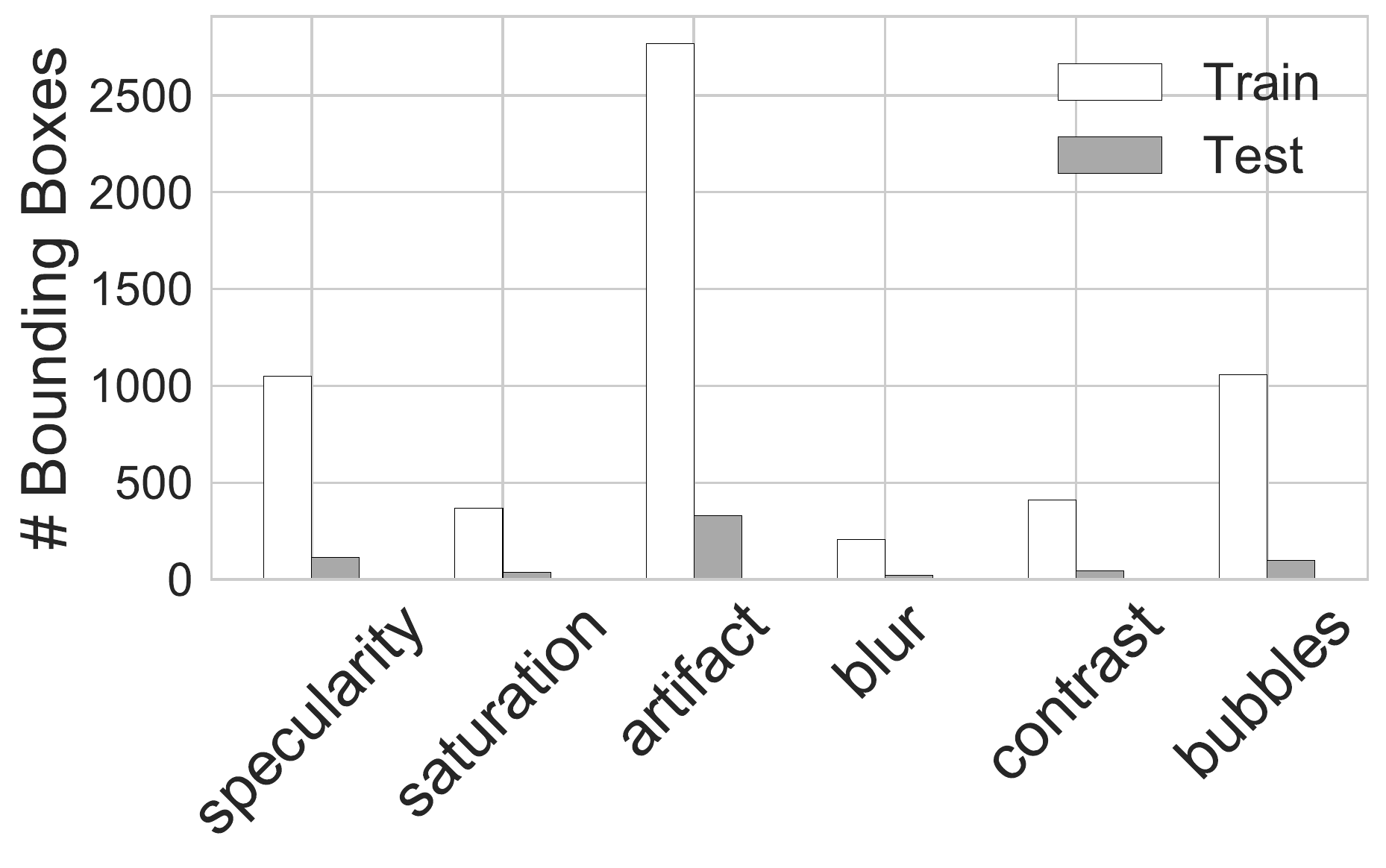}
\end{minipage}
\begin{minipage}[b]{0.48\linewidth}
\includegraphics[width=\linewidth]{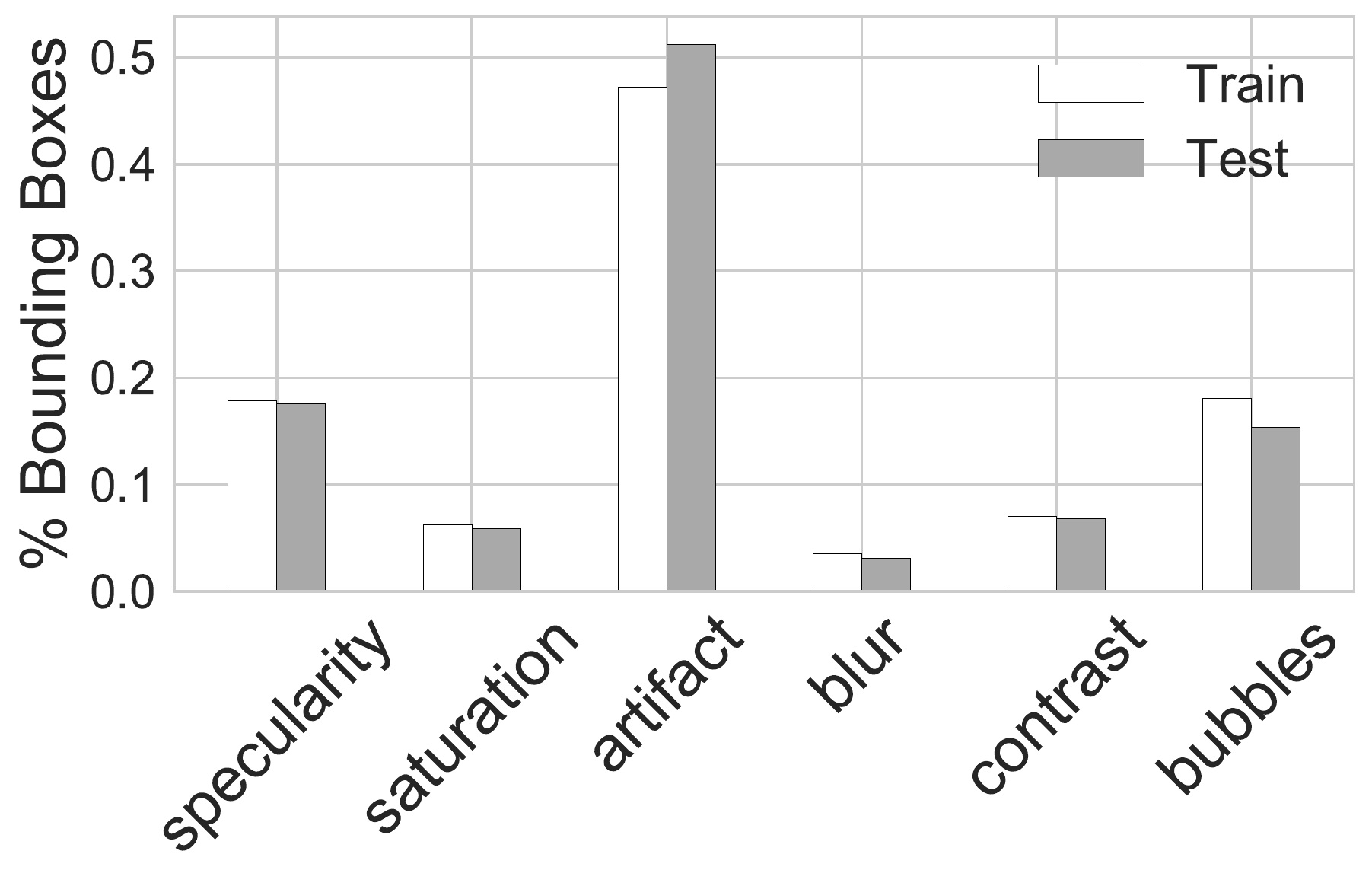}
\end{minipage}
\begin{minipage}[b]{0.24\linewidth}
\includegraphics[width=\linewidth]{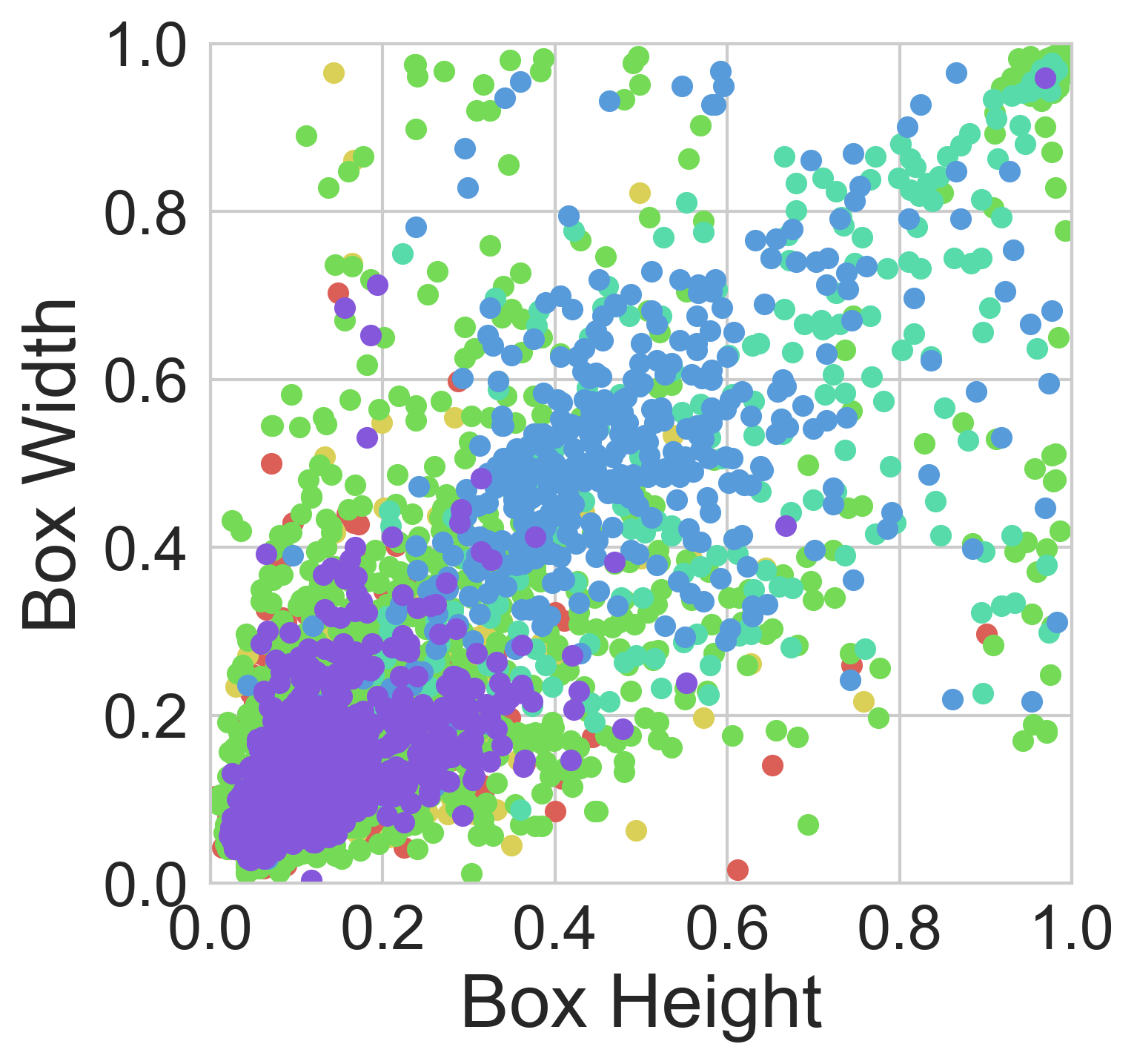}
\end{minipage}
\begin{minipage}[b]{0.24\linewidth}
\includegraphics[width=\linewidth]{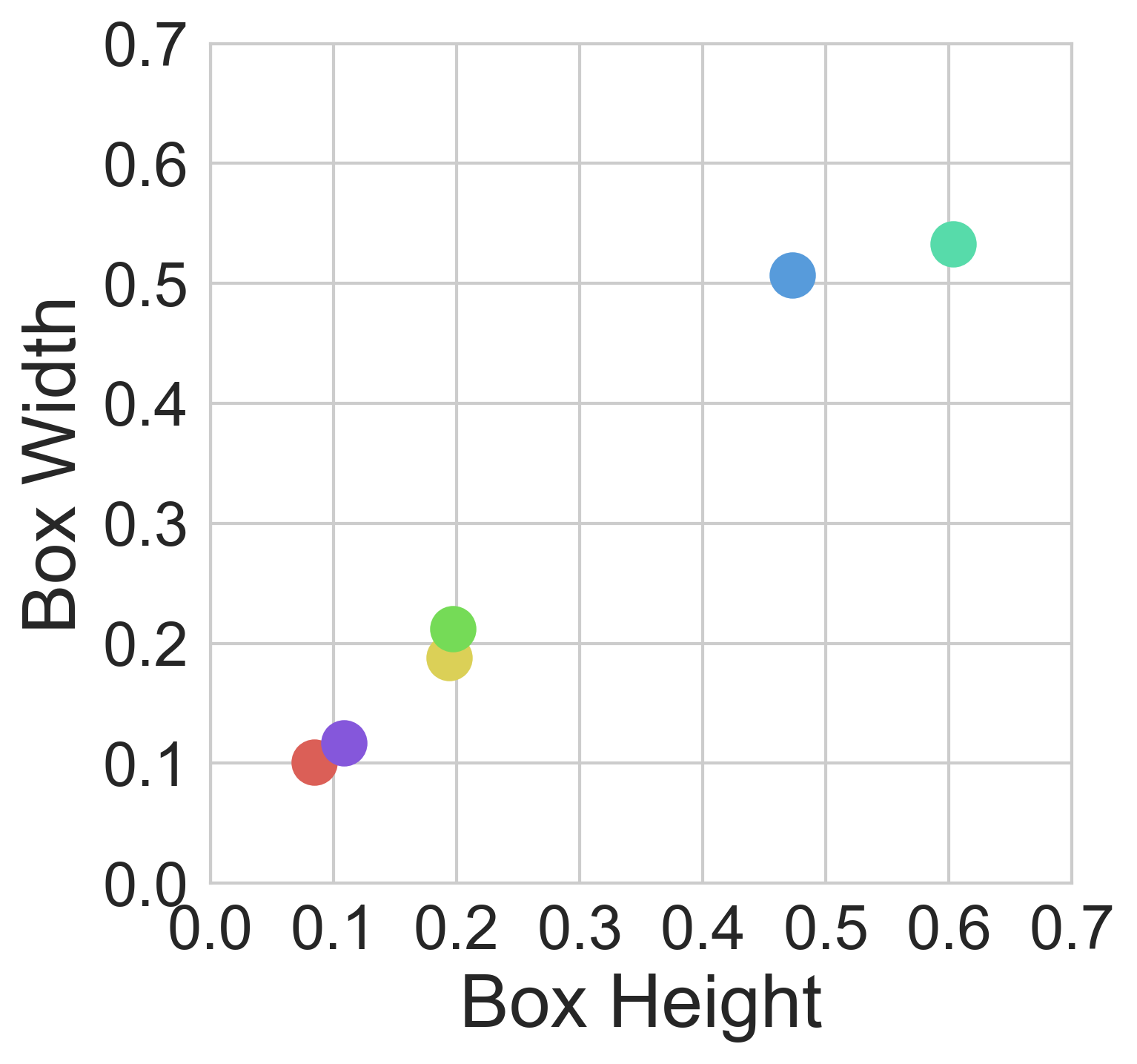}
\end{minipage}
\begin{minipage}[b]{0.24\linewidth}
\includegraphics[width=\linewidth]{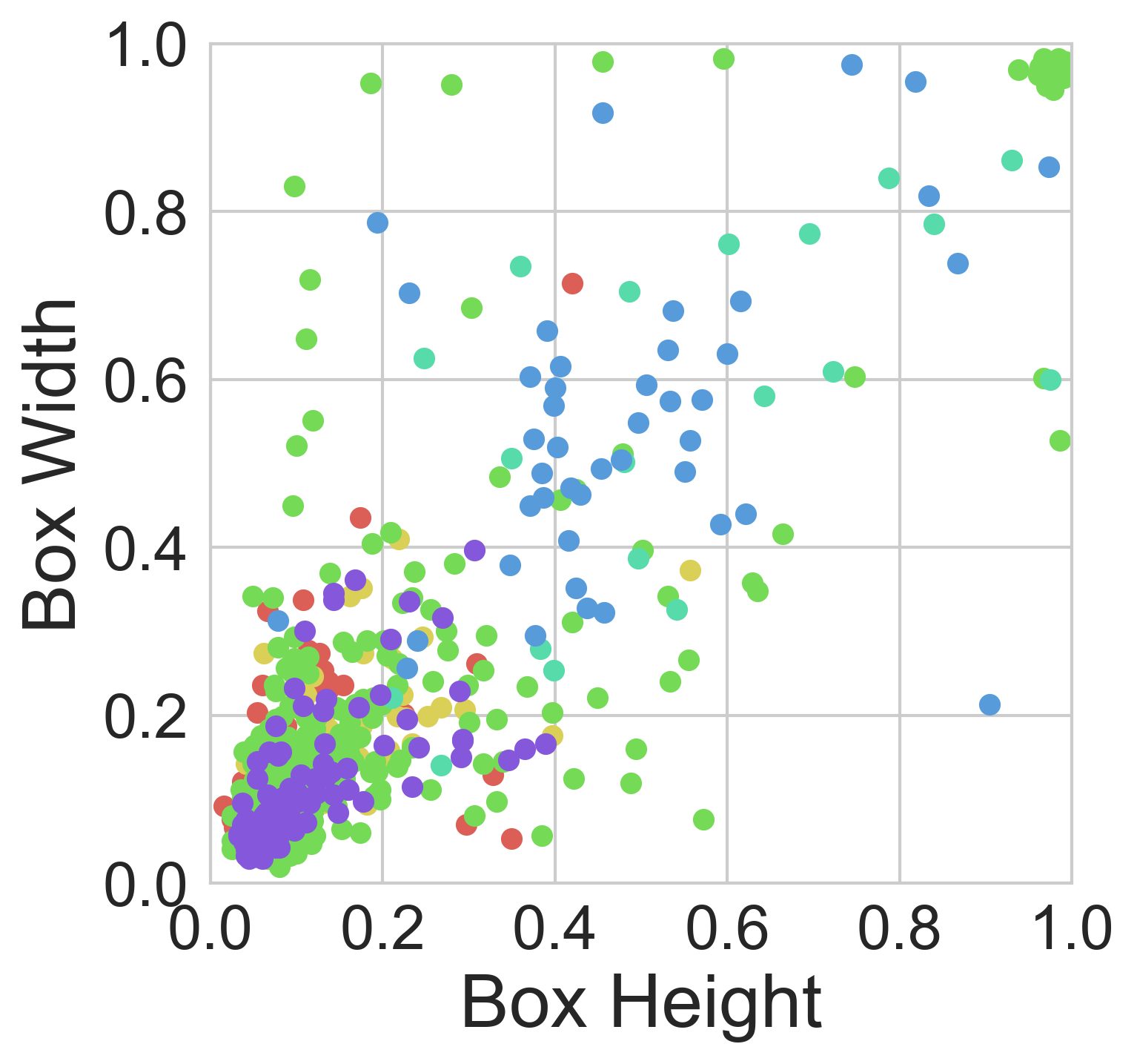}
\end{minipage}
\begin{minipage}[b]{0.24\linewidth}
\includegraphics[width=\linewidth]{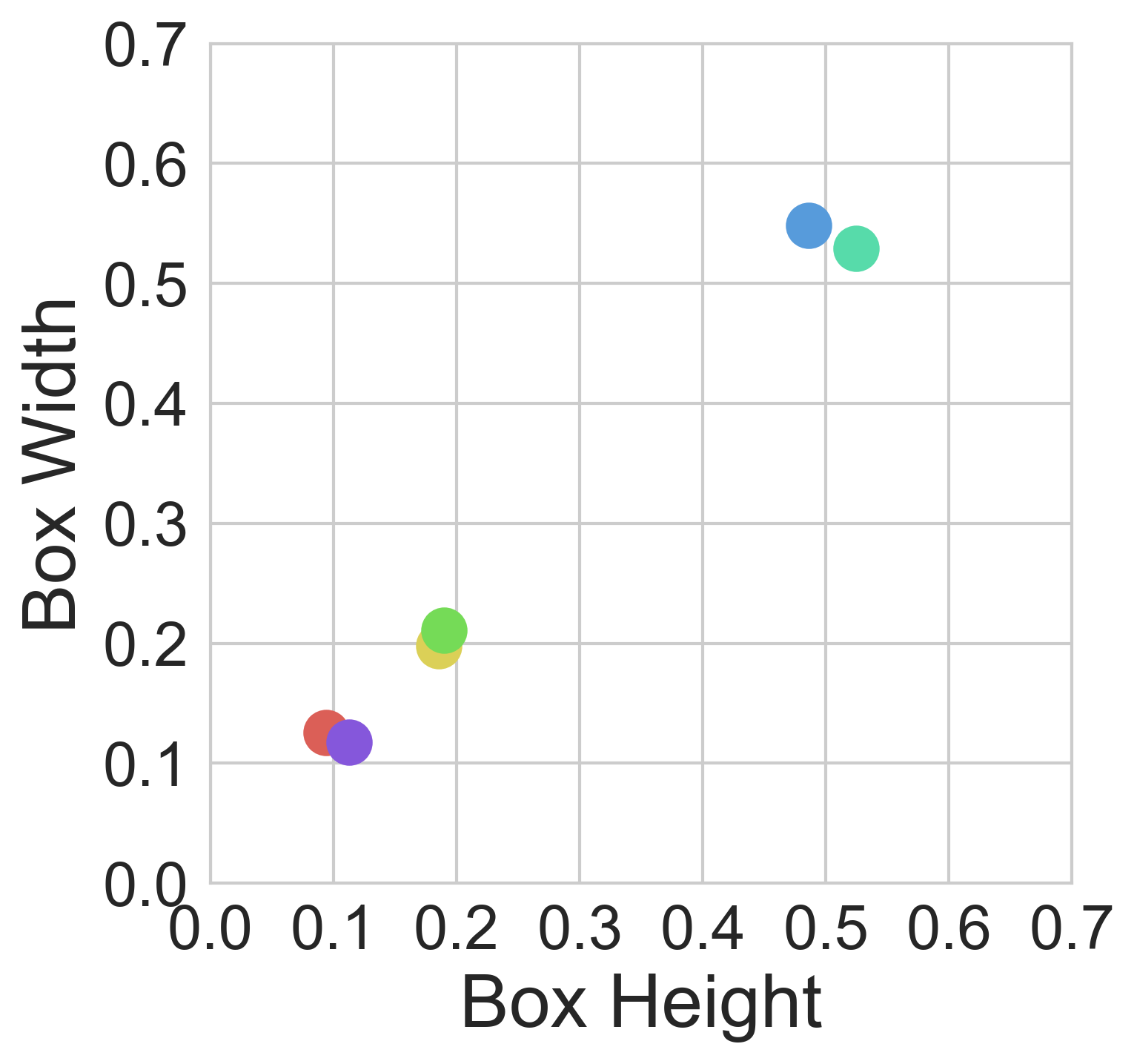}
\end{minipage}

\begin{minipage}[b]{\linewidth}
\includegraphics[width=\linewidth]{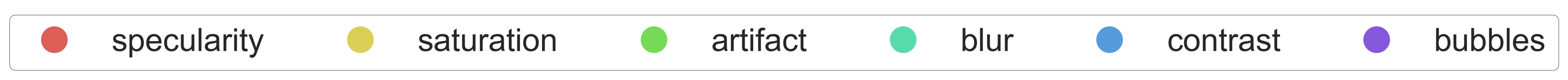}
\end{minipage}
\caption{Top row: artifact type distribution in the training and testing enbdoscopy image data set in terms of number of bounding boxes, left and percentage of the total number of bounding boxes, right. Bottom row: sizes of annotated boxes normalised by the image dimensions of the training set, left and the test set, right.}\label{fig:data set_distribution_YOLO_endo}
	
\end{figure}

Our artifact detection data set consists of a total of 1290 endoscopy images (resized to 512 x 512 pixels) from two operating modalities; normal bright field (BF), and narrow-band imaging (NBI) sampled from 7 unique patient videos selected from a cohort of 200 endoscopic videos for training data. The selection was based on number of representative artifacts present in these videos and  texture variability of the underlying esophagus. Two experts annotated a total of 6504 artifacts using bounding boxes where each annotation is classified as:
\begin{enumerate}
	\item \textbf{blur} - streaking from fast camera motion 
	\item \textbf{bubbles} - water bubbles that distorts appearance of the underlying tissue
	\item \textbf{specularity} - mirror-like surface reflection
	\item \textbf{saturation} - overexposed bright pixel areas
	\item \textbf{contrast} - low contrast areas from underexposure or occlusion
		\item \textbf{misc. artifact} (also referred as `artifact` in this paper)- miscellaneous artifacts; e.g., chromatic aberration, debris, imaging artifacts etc.
\end{enumerate}
A 90\%-10\% split was used to construct the train-test set for object detection resulting in 1161 and 129 images and 5860 and 644 bounding boxes, respectively. In general, the training and testing data exhibits the same class distribution (see Fig.\ref{fig:data set_distribution_YOLO_endo} (top row)) and similar bounding boxes (roughly square) but either small with average widths less than~0.2 or large with widths greater than~0.5 (see Fig.\ref{fig:data set_distribution_YOLO_endo} (bottom row)). Multiple annotations are used in case a given region contains multiple artifacts.
\section{Method}{\label{sec:method}}
\begin{figure}[t!]
	\centering
	\includegraphics[scale=0.25]{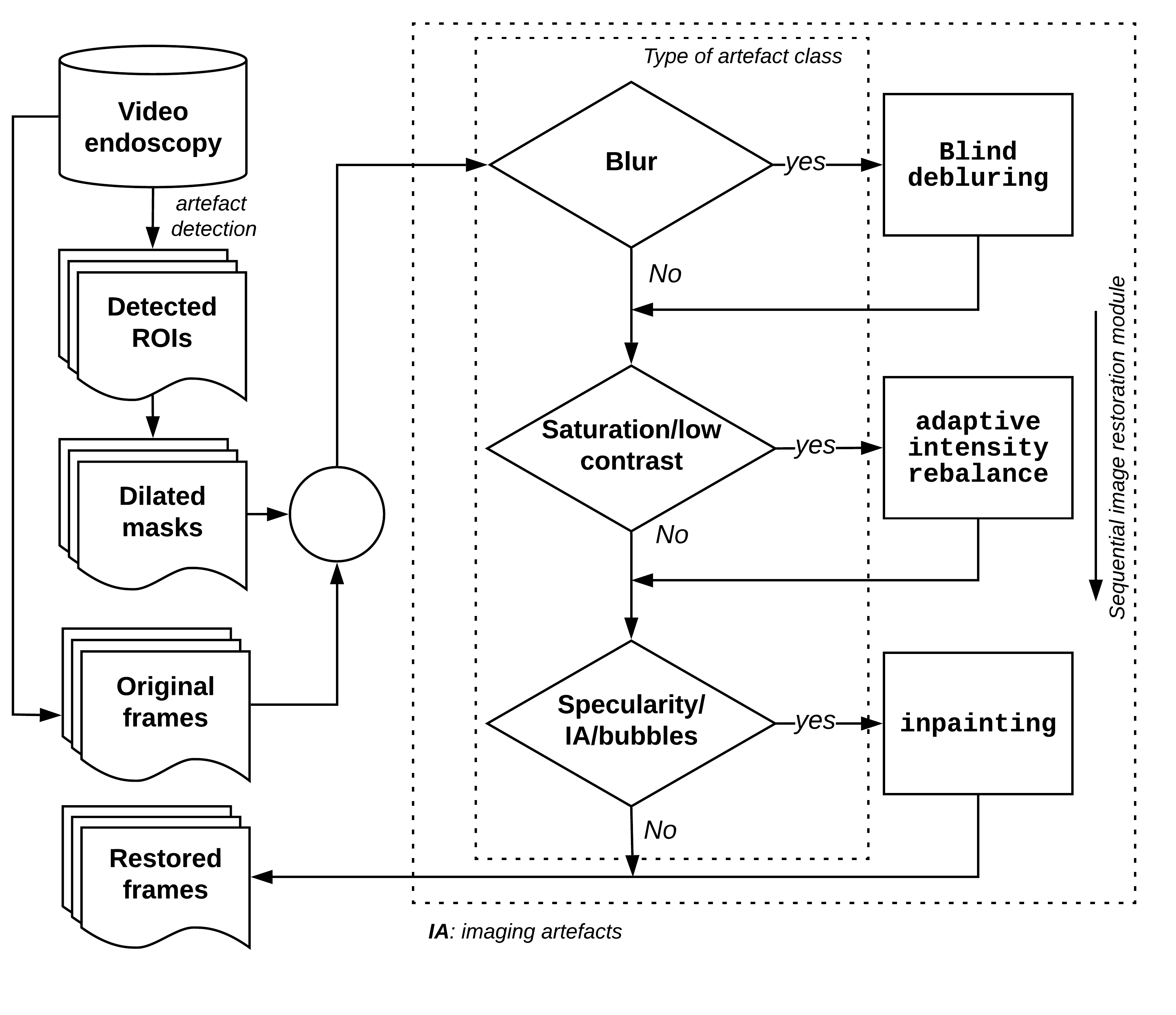}
	\caption{Sequential processes for endoscopic image restoration from detected region-of-interests (ROI) of 6 different artifacts. First, masks of generarted ROIs are dilated and then only these regions are used for restoration. Unlike, in case of blur, the entire image is used. {\label{fig:sequentialProcess}}}
\end{figure}
%
\subsection{Overall approach}
The step-by-step procedure for automatic detection of multiple artifacts and frame restoration of endoscopic videos is presented in Fig.~\ref{fig:sequentialProcess}. It is to be noted that a single frame can be corrupted by multiple artifacts and each artifact class can affect endoscopic frames differently. Therefore, their restoration process is very likely to affect the final restoration result. 

Multiple instance object detection is used to discriminate between the six different types of artifacts (see Section~\ref{sec:materials}) and normal appearance. For each frame a quality score (QS, refer Section~\ref{sec:QS}) is computed based on the type, area and location of the identified artifacts to reflect the feasibility of complete image restoration via the sequential restoration process depicted in Fig.~\ref{fig:sequentialProcess}. The scaling of our QS score is set such that we differentiate between severely corrupted frames ($QS < 0.5$), mildly corrupted frames ($0.5 \leq OS \leq 0.95$), and frames of high quality ($QS > 0.95$). Severely corrupted frames are discarded without any further processing. The proposed image restoration methods are applied to mildly corrupted frames only. In order to guarantee a faithful restoration, mildly corrupted frames go through our proposed sequential framework. All remaining frames are directly concatenated into the final list without any processing. 
%
\subsection{Artifact region detection}
Recent research in computer vision provides us with object detectors that are both robust and suitable for real-time applications. Here, we propose to use a multi-scale deep object detection model for identifying the different artifacts in real-time. Even by itself, real-time artifact detection is already of practical value. For example, the detection results can be used to provide endoscopists with feedback during data acquisition. After detection, additional post-processing using traditional image processing methods can be used to determine the precise boundary of a corrupted region in the image. 
%
\begin{figure}[t!]
\begin{minipage}[b]{0.24\linewidth}
\includegraphics[scale=0.12]{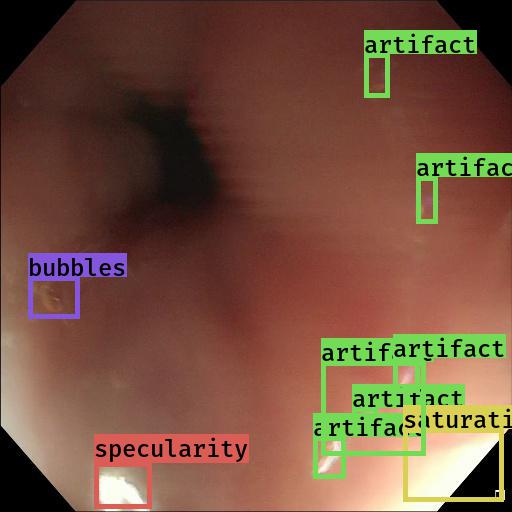}
\end{minipage}
\begin{minipage}[b]{0.24\linewidth}
\includegraphics[scale=0.12]{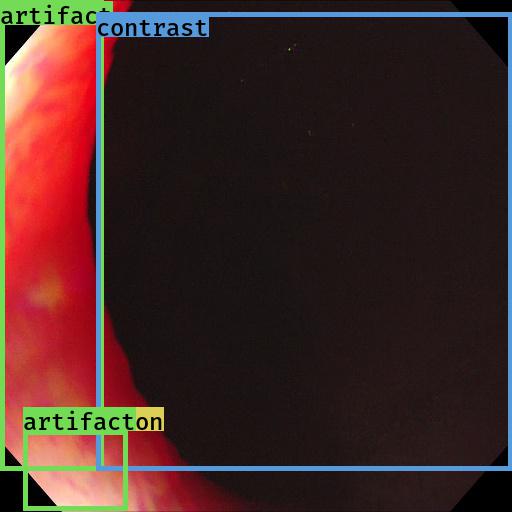}
\end{minipage}
\begin{minipage}[b]{0.24\linewidth}
\includegraphics[scale=0.12]{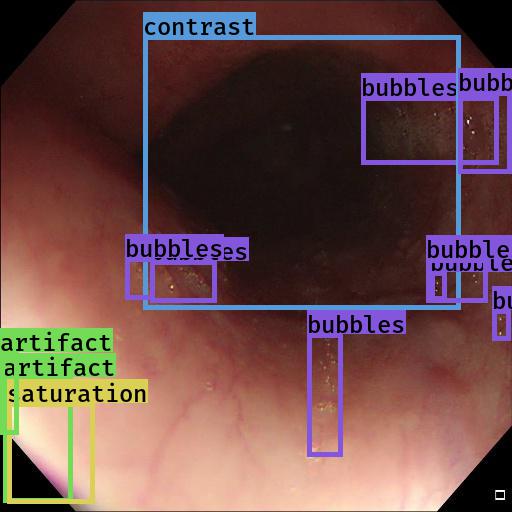}
\end{minipage}
\begin{minipage}[b]{0.24\linewidth}
\includegraphics[scale=0.12]{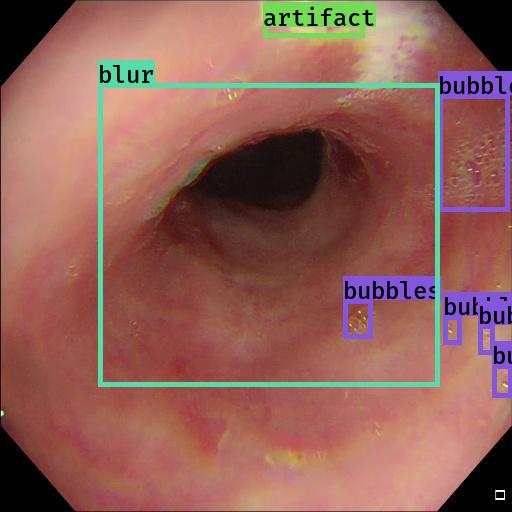}
\end{minipage}
\caption{Examples of detected bounding boxes for some artifact class labels using YOLOv3-spp.}{\label{fig:yolov3DetectedBoxes}}
\end{figure}

Today, deep learning enables us to construct object detectors that generalise traditional hand-crafted `sliding-window' object classification approaches (e.g., Viola-Jones~\cite{viola2001rapid}). Earlier attempts of including OverFeat~\cite{sermanet2013overfeat} and R-CNN~\cite{girshick2014rcnn} demonstrated the power of convolutional neural networks (CNNs) to learn relevant features and detect objects using a fixed number of pre-generated candidate object region proposals~\cite{uijlings2013selectivesearch}. Faster R-CNNs~\cite{ren2015faster} first introduced a fully trainable end-to-end network yielding an initial region proposal network and successive classifications of the proposed regions without intermediate processing. Since region proposal generation precedes bounding box detection sequentially, this architecture is known as a two-stage detector. Though very accurate, a primary drawback is its slow inference and extensive training. You Only Look Once (YOLO,~\cite{redmon2016yolov1}) simplified Faster R-CNNs to predict simultaneously class and bounding box coordinates using a single CNN and a single loss function with good performance and significantly faster inference time. This simultaneous detection is known as a one-stage detector. Compared to two-stage detectors, single-stage detectors mainly suffer two issues: high false detection due to 1) presence of varied size objects and 2) high initial number of anchor boxes requirement that necessitates more accurate positive box mining. The former is corrected by predicting bounding boxes at multiple scales using feature pyramids~\cite{he2014spp}-\!\!\cite{lin2017FPN}. To address the latter, RetinaNet \cite{lin2017focal} introduced a new focal loss which adjusts the propagated loss to focus more on hard, misclassified samples. Recently, YOLOv3~\cite{redmon2018yolov3} simplified the RetinaNet architecture with further speed improvements. Bounding boxes are predicted only at 3 different scales (unlike 5 in RetinaNet) utilizing objectness score and an independent logistic regression to enable the detection of objects belonging to multiple classes  unlike focal loss in RetinaNet. Collectively, Faster R-CNN, RetinaNet and YOLOv3 define the current state-of-the-art detection envelope of accuracy vs speed on the popular natural images benchmark COCO data set~\cite{cocochallenge2014}. 

We investigated the Faster R-CNN, RetinaNet and YOLOv3 architectures for artifact detection. Validated open source codes are available for all of these architectures. Experimentally, we chose to incorporate YOLOv3 with spatial pyramid pooling (YOLOv3-spp) for robust detection and improved inference time for endoscopic artifacts detection. Spatial pyramid pooling allowed to pool features from sub-image regions utilizing computed single-stage CNN features at multiple-scales from YOLOv3 architecture. In addition to the boost in the inference speed, incorporating spatial pyramid pooling decreased false positive detections compared to classical YOLOv3 method (see Section~\ref{sec:artifact_exp}). YOLOv3-spp provided an excellent feature for accuracy-speed trade-off which are the main requirements for usage in clinical settings. Examples of the detected boxes using YOLOv3-spp are shown in Fig.~\ref{fig:yolov3DetectedBoxes}. 
\subsection{Quality score}{\label{sec:QS}}
\begin{figure}[t!]
\centering
\begin{minipage}[b]{0.48\linewidth}
\includegraphics[trim=0.0cm 3.0cm 0.0cm 0.0cm, clip=true,width=\linewidth]{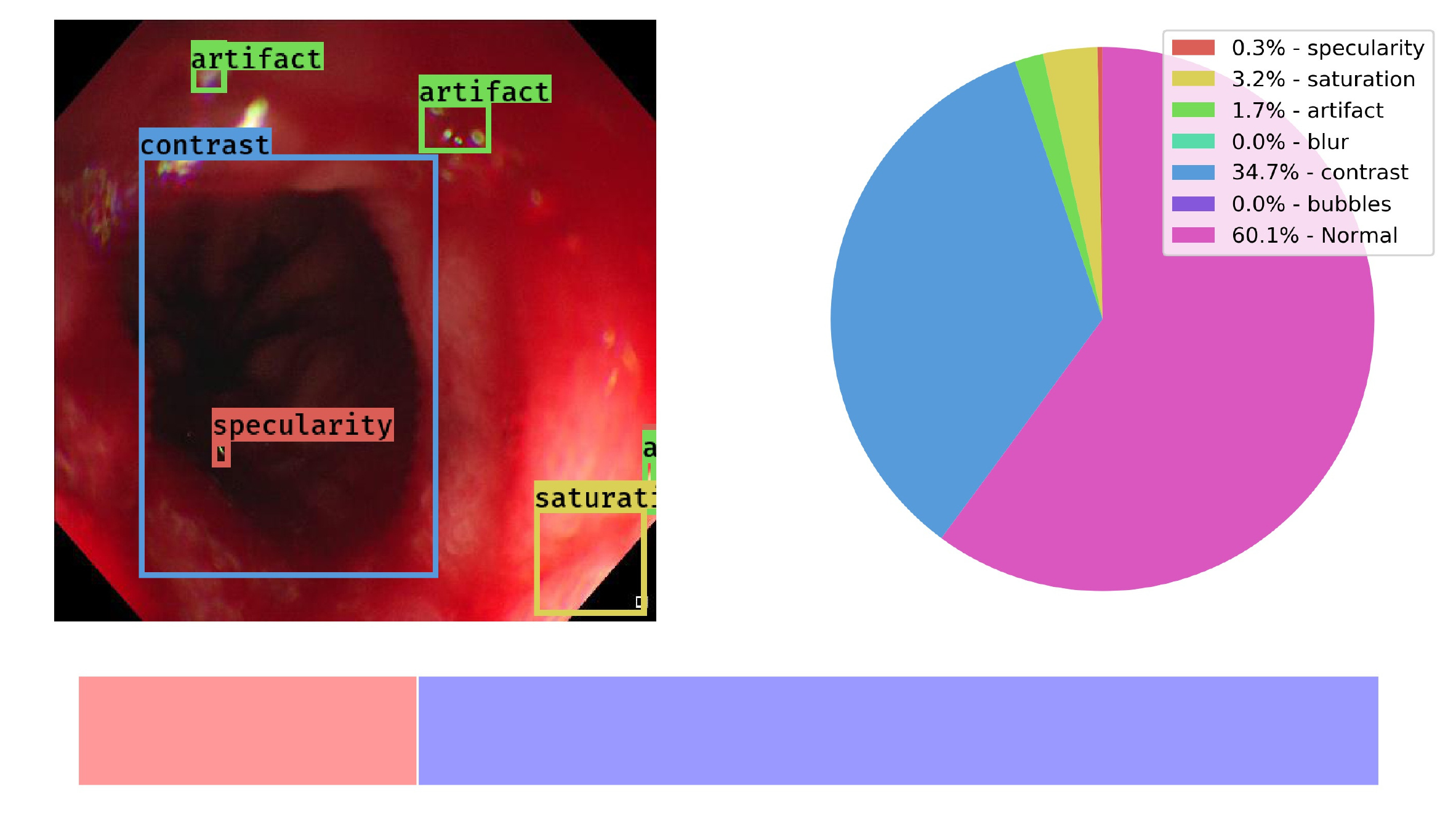}
\centerline{\footnotesize{QS = 0.75}}\medskip
\end{minipage}
\begin{minipage}[b]{0.48\linewidth}
\includegraphics[trim=0.0cm 3.0cm 0.0cm 0.0cm, clip=true,width=\linewidth]{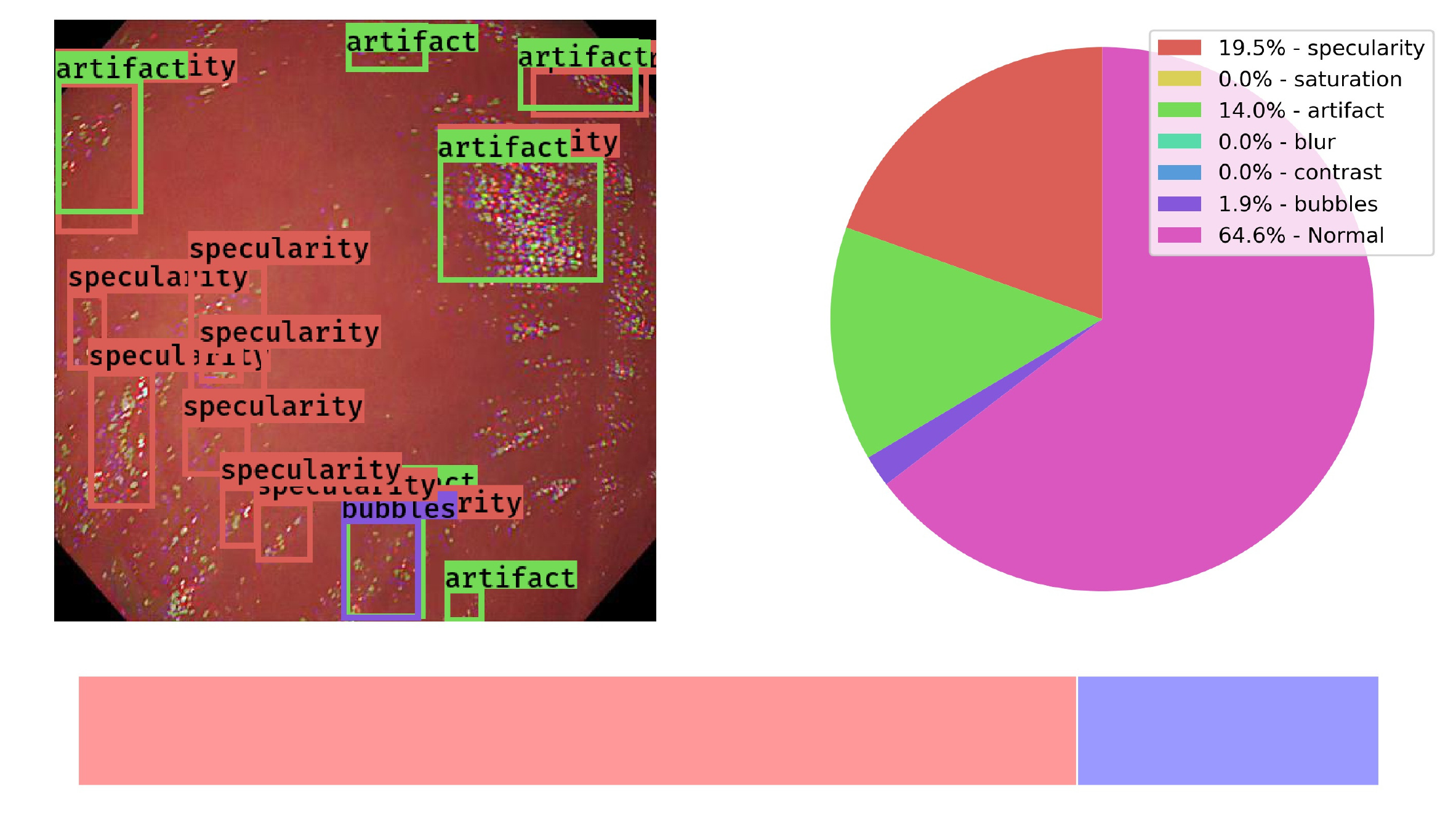}
\centerline{\footnotesize{QS = 0.23}}\medskip
\end{minipage}
\caption{Quality assessment based on class weight, area and location. Images with detection boxes and their corresponding area fraction are shown. On left: shows image with mostly contrast problem and on right: shows that with multiple misc. artifacts and specularities. Below are their calculated quality scores.~\label{fig:QS}}
\end{figure}
%
Quality assessment is important in video endoscopy as image corruption largely affects image analysis methods. However, it is likely that not all frames are corrupted in same proportion. Depending on the amount and type of artifact present in frames realistic frame restoration can be possible. However, such frame grading needs to be carefully determined. Here, we propose a frame quality score (QS) based on: a) type, b)  area and c) location of the detected artifacts. Weights are assigned to each of these categories and a mean weight is computed as the quality score. Weights are assigned to each type based on the ease of restoration, e.g., an entire blurred image can still be restored but the same would not apply with misc. artifacts. Thus, misc. artifacts are assigned a higher weight than blur. Similarly, the area and location of detected artifacts in each frame are important. A centrally located imaging artifact with large area detrimentally degrades image information beyond restoration. Below we describe our weighting scheme:
\begin{itemize}
\item Class weight ($W_C$): misc. artifact (0.50), specularity (0.20), saturation (0.10), blur (0.05), contrast (0.05), bubbles (0.10)
\item Area weight ($W_A$): percentage of the total image area occupied by all detected artifact areas and normal areas 
\item Location weight ($W_L$): center (0.5), left (0.25), right (0.25), top (0.25), bottom (0.25), top-left (0.125), top-right (0.125), bottom-left (0.125), bottom-right (0.125). 
\end{itemize}

The final QS is computed as:
\begin{equation}{\label{eq:QS}}
QS = \lfloor{ 1 -  \sum\limits_{\mathcal B} (\lambda_A W_C W_A + \lambda_L W_C W_L)} \rfloor_0,
\end{equation}
where $\mathcal B$ denotes the set of bounding boxes associated to each detected artifact, $\lambda_A$, $\lambda_L$ are constants that weight the relative contributions of area and location. 
%
We have used $\lambda_{{A}}=0.5,\lambda_{{L}}=0.5$ in our experiments. However, for frames with few detected artifacts (less than 5) such weighting scheme underscores (especially if large area artifacts are present) thus $\lambda_{{A}}=1,\lambda_{{L}}=1$ is used for these cases. Note that QS score in Eq.~(\ref{eq:QS}) is lower-bounded by 0. 

A fixed threshold is user-specified to determine the frames kept for image restoration. Examples of the proposed quality score applied to real data are shown in Fig.~\ref{fig:QS}.  The video frame in Fig.~\ref{fig:QS} (left) has mostly a contrast problem (i.e., low $W_C$) so despite its central location (see blue box) and large area the frame intensity can be restored ($\therefore$QS$=$0.75). However Fig.~\ref{fig:QS} (right) has many misc. artifacts (high $W_C$) and specular areas located centrally centrally (c.f. green, red boxes, $\therefore$ QS=0.23) which inhibits realistic frame restoration so the frame is discarded.
\subsection{Image restoration}{\label{sec:restoration_sec}}
Formulating the reconstruction of the true signal given the noisy and corrupted input image $I$ as an  optimization or estimation problem demands a well-motivated mathematical model. Unfortunately, the various different types of artifacts induce a level of complexity that
make this endeavor very challenging. Assuming image noise to be additive and approximating motion blur as a linear convolution with an unknown kernel is reasonable and in line with previous attempts to the problem. In addition, contrast and pixel saturation problems can be formulated as a non-linear gamma correction. Other remaining artifacts (e.g., specularities, bubbles and imaging artifacts) which are due to combined processes of these phenomena can be assumed as a function of the entire process. The corrupted noisy video frame can thus be approximated as:
\begin{equation}{\label{eq:corruptedImage}}
	I(t) = F\left[(h\,*\,f(t) + \eta)^\gamma\,\right],
\end{equation}
where $\eta$ denotes the additive noise induced by the imaging system, the convolution with $h$ the approximation to the induced motion blur, $\gamma$ captures the over- and under-exposed regions and $F$ is a generalized non-linear function that models capturing other artifacts as well (including specularities, bubbles and imaging artifacts) or a combination of them. This model motivates why the restoration of the video frames is structured into separate processing steps, which are implemented as deep learning models. 
 
Image restoration is the process of generating realistic and noise free image pixels from corrupted image pixels. In endoscopic frame restoration, depending upon the artifact type, the goal is either the generation of an entire noise-free image or pixel inpainting of undesirable pixels using surrounding pixel information~\cite{Celia:IVC07}. For multi-class endoscopic artifact restoration, we require 1) frame deblurring when $h(.)$ is unknown, i.e. a blind deblurring task, 2) minimize the effect of contract imbalance (correction for over- and under-exposed regions) in frames, i.e. $\gamma$ correction, 3) replace specular pixels and those with imaging artifacts or debris with inpainting, i.e. correction for additive noise $\eta(.)$ or a combined non-linear function $F(.)$. Due to the higher likelihood of the presence of multiple artifacts in a single frame, unordered restoration of these artifacts can further annihilate frame quality. We therefore propose an adaptive sequential restoration process that account for the nature of individual artifact types (see Fig.~\ref{fig:sequentialProcess}). 

Recently, GANs~\cite{Goodfellow:NIPS14} have been successfully applied to image-to-image translation problems using limited training data. Here, a generator $G$ `generates' a sample $G(z)$ from a random noise distribution ($p_{noise}(z)$ with $z\!\!\sim \!\!\mathcal{N}(0, \sigma^2\textit{I})$) while a separate discriminator network tries to distinguish between the real target images ($p_{data}(x)$ with assumed $x\!\sim\!$ non-zero mean Gaussian) and the fake image generated by the generator. The objective function $V$ is therefore a min-max problem in this case:
\begin{align}{\label{eq:GAN}}
\min_{G}\max_{D} V(D,G) = \mathbb{E}_{x\sim p_{data}(x)}[\log D(x)] \nonumber~+\\
\mathbb{E}_{z\sim p_{noise}(z)}[\log (1-D(G(z)))] 
\end{align}
In practice, the generator model in Eq.~(\ref{eq:GAN}) is highly non-convex, unstable and slow to train as samples are generated from random input noise. Various groups~\cite{Mirza2014ConditionalGA,CycleGAN2017,pix2pix2017,Karras:NIPS18} have provided ways to address this problem and achieved improvements in reconstruction quality and numerical stability as well as a reduction in computation time. 
One popular way to ensure the stability of the generator output is by conditioning the GAN on prior information (e.g., the class label `$y$` in CGAN,~\cite{Mirza2014ConditionalGA}). The objective function  $V_{cond}$ for CGAN can be written as:
\begin{align}{\label{eq:CGAN}}
\min_{G}\max_{D} V_{cond}(D,G) = \mathbb{E}_{x, y\sim p_{data}(x,y)}[\log D(x|y)]\nonumber~+\\
\mathbb{E}_{y\sim p_{y}, z\sim p_{z}}[\log (1-D(G(z|y), y))] 
\end{align}
Another efficient method is regularizing the generator using contextual losses~(e.g., pix2pix~\cite{pix2pix2017}, deblurGAN~\cite{deblurGAN}). In~\cite{Bang:CoRR18} regularizing the discriminator and generator significantly helped to improve visual quality. We train such conditional generative adversarial models~\cite{Mirza2014ConditionalGA} (CGAN) embedding artifact class dependent contextual losses, see Table~\ref{tab:frameworkTable} for effective restoration. When restoring frames, the detected artifact types are used to decide which CGAN model is applied and in what order (also see Fig.~\ref{fig:sequentialProcess}). 
\begin{table}[t!]
	\centering
	\begin{tabular}{|m{2.6cm}|>{\centering\arraybackslash}m{3.2cm}|}
		\hline
		\textbf{artifact type} & \textbf{Restoration method}\\
		\hline\hline
		Motion blur & CGAN + $l2$-contextual + high-frequency losses\\
		\hline
		specularity/bubbles/misc. artifacts & CGAN + $l1$-contextual loss\\
		\hline
		saturation & CGAN + $l2$-contextual loss + CRT transform\\
		\hline
		Low contrast & same as saturation (reversed training set)\\
		\hline
	\end{tabular}
	\caption{Computational models used for individual artifact classes.}
	\label{tab:frameworkTable}
\end{table}
%
\subsubsection{Motion blur}
Motion blur is a common problem in endoscopy videos. Unlike static images, motion blur is often non-uniform with unknown kernels $h(.)$ (see Eq.~(\ref{eq:corruptedImage})) in video frame data. Several blind-deconvolution have been applied to motion deblurring. These range from classical optimization methods~\cite{tong2004blur,Xu:CVPR13,ipoltvdc} to neural network-based methods~\cite{Chakrabarti:ECCV16,Nah:CVPR17}. Despite good performance of convolutional neural networks (CNNs) over classical methods, a major drawback of CNNs is that they require tuning a large number of hyper-parameters and large training data sets. Blind deconvolution can be posed as an image-to-image translation problem where the blurred image is transformed into its matching unblurred image. 
\begin{figure}[t!]
	\centering
	\includegraphics[scale=0.3]{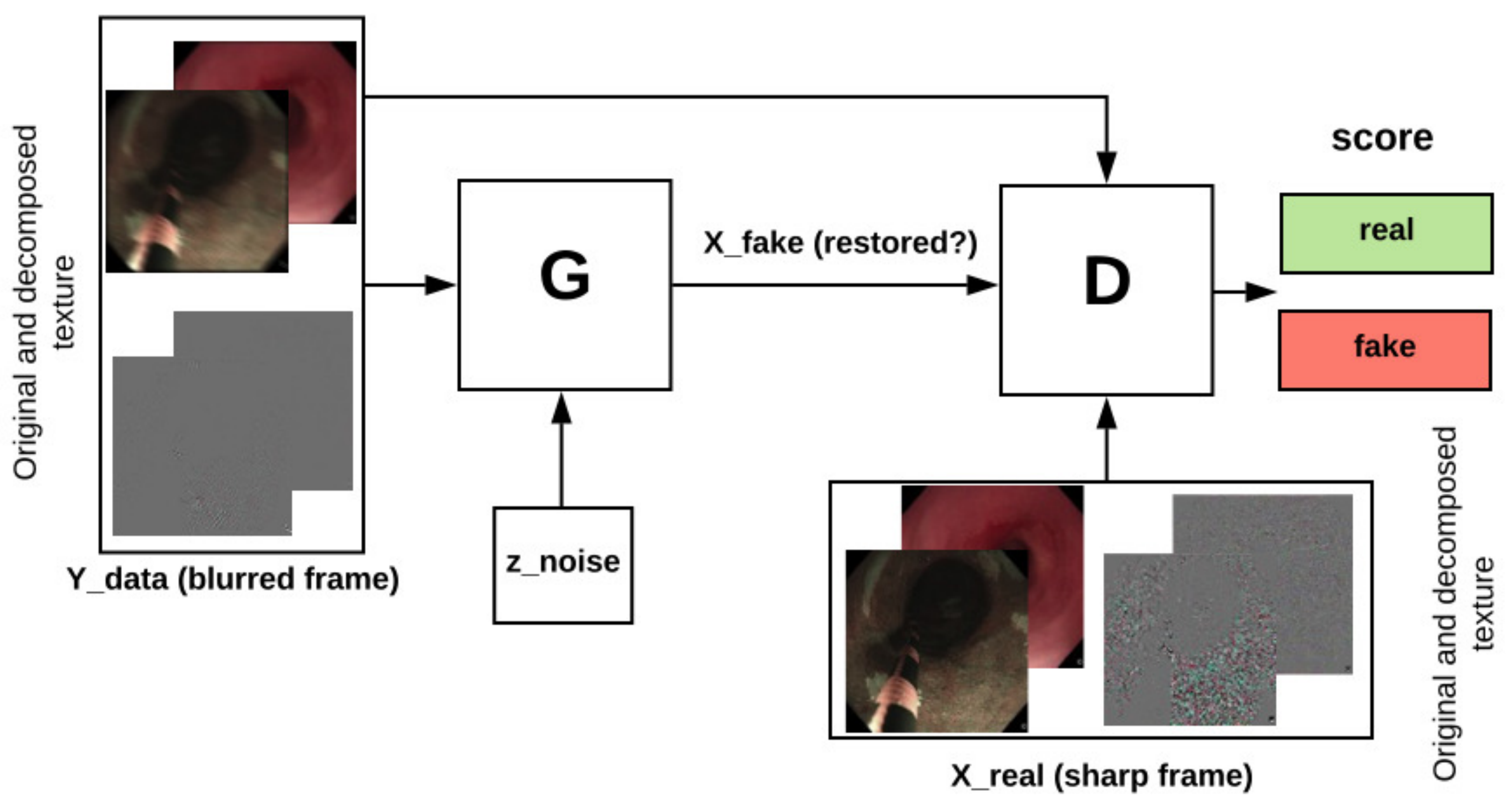}
	\caption{Blind deblurring using CGAN with added contexual high-frequency feature loss.{\label{fig:imageDebluringCGAN}}}
\end{figure}
In this work, we use CGAN with a $l2$-contextual loss (squared difference between generated and target/sharp image) and an additional $l2$ high-frequency loss as regularization. This is motivated by the fact motion blur primarily affects image edges, a few discriminative image pixels compared to the entire image. The high-frequency images are first computed both for blurred and sharp images in the training data using iterative low pass-high pass filtering at 4 different scales~\cite{Buades:IPOL11}. These images are then used to provide additional information to the discriminator regarding the generator's behavior (also see Fig.~\ref{fig:imageDebluringCGAN}). Equation~\eqref{eq:CGAN} becomes:
\begin{align}{\label{eq:CGAN2}}
\min_{G}\max_{D} V_{cond}^{'}(D,G) = V_{cond} + \sum_{i}{\lambda \parallel x_{real_{i}} - G(z_{i}|y_{i}) \parallel}_{l},
\end{align}
where $\lambda = 50$, $i = \left[0, 1\right]$ refer to an original and high-frequency image pair and $l=2$. $x_{real}$ is the ground truth image for restoration (i.e. sharp images in our case). Minimization of Eq.~(\ref{eq:CGAN}) using Jensen-Shannon (JS) divergence as in~\cite{Goodfellow:NIPS14} can lead to problems like mode collapse, vanishing gradients. Consequently,~\cite{Arjovsky:ICML17} proposed to use Wasserstein distance with gradient penalty (WGAN-GP). We propose thus to use CGAN with a critic network based on WGAN-GP~\cite{deblurGAN}. The proposed model was trained for 300 epochs on a paired blur-sharp data set consisting of 10,710 (715 unique sharp images) multi-patient and multi-modal images with 15 different simulated motion trajectories for blur (see~\cite{deblurGAN}). 
\subsubsection{Saturation or low contrast}
\label{subsec:saturation}
The small or larger distances between the light source and the imaged tissue can lead to large illumination changes which can result in saturation or low contrast. This motivates the role of the variable $\gamma$ in Eq.~(\ref{eq:corruptedImage})
Saturated or low contrast image pixels often occur across large image areas compared to specularities and affect the entire image globally. In addition, these illumination changes are more prominently observed in normal brightfield (BF) modality compared to other modalities. Compensation of affected image pixels is a difficult problem depending on the size of the affected image area.  We pose the saturation restoration task as an image-to-image translation problem and apply the same end-to-end CGAN approach used for motion deblur described above with $\mathit{l}_2$ contextual loss only to train a generator-discriminator network for saturation removal. Here, $\mathit{l}_2$ contextual loss is more suitable as we want to capture the deviation between normal illumination condition w.r.t saturation and low contrast conditions.

\begin{figure}[t!]
	\centering
	\includegraphics[scale=0.35]{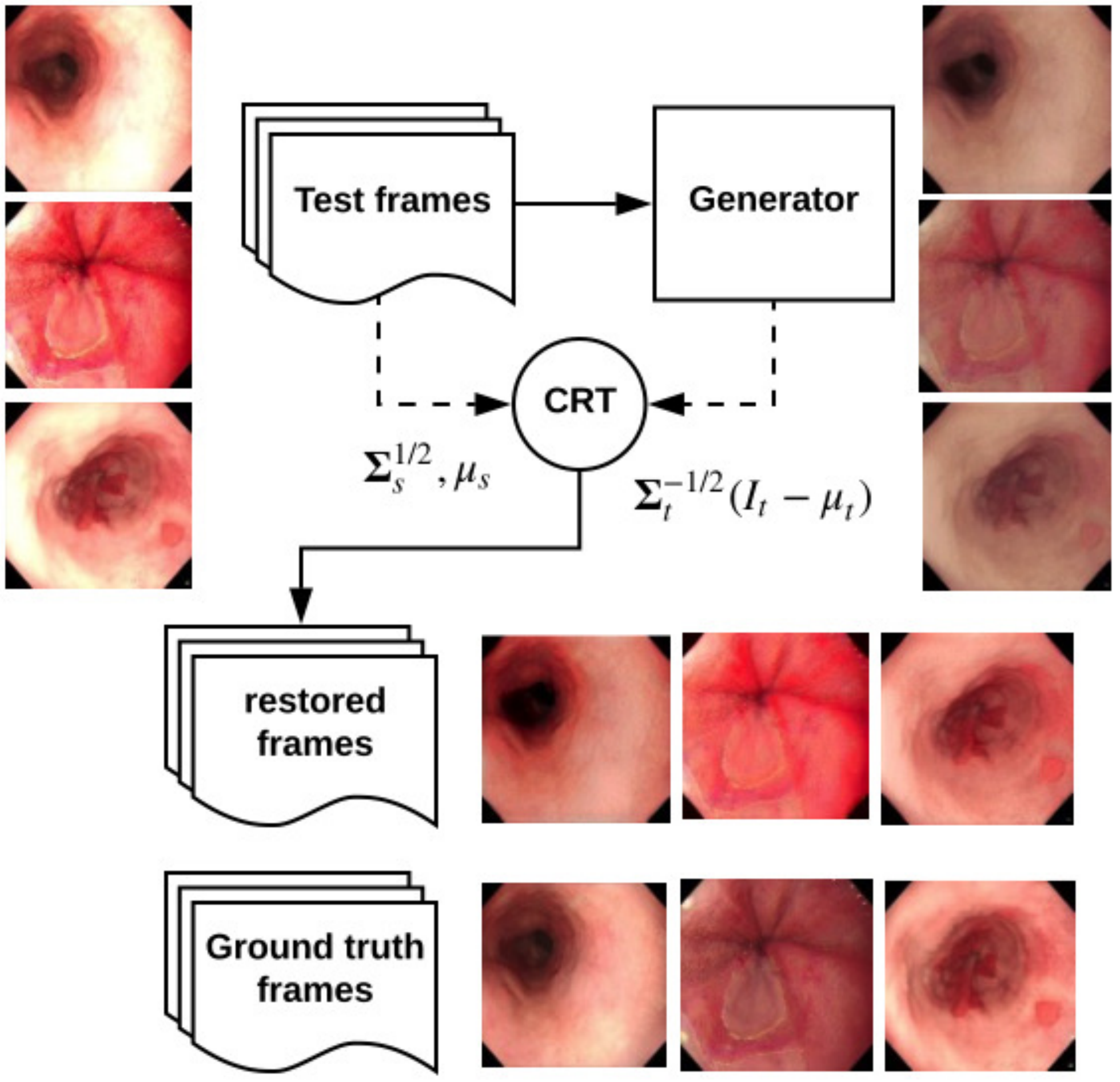}
	\caption{CRT color correction. Saturation corrected frames generation by our trained generator (right), and color transfer comparison with ground truth (bottom). {\label{fig:crtCorrection}}}
\end{figure}

Due to lack of any ground truth data for two different illumination conditions, we created a fused data set that included: 200 natural scene images containing diffuse (scattered light) and ambient (additional illumination to natural light giving regions with pixel saturation) illuminations \footnote{\url{https://engineering.purdue.edu/RVL/Database/specularity_database/index.html}}; and 200 endoscopic image pairs simulated using cycleGAN-based style transfer~\cite{CycleGAN2017} (separately trained on another 200 images with saturated and normal BF images with from 7 unique patients). To correct coloration shift due to the incorporation of natural images in our training set, color transfer (CRT) is applied to the generated frames. Given a source image, $I_{s}$ and a target image, $I_{t}$ to recolor, the mean $(\mathbf{\mu}_s, \mathbf{\mu}_t)$ and covariance matrix $(\mathbf{\Sigma}_s, \mathbf{\Sigma}_t)$ of the respective pixel values (in RGB channels) can be matched through a linear transformation~\cite{hertzmann2001algorithms}:
\begin{align}{\label{eq:color_transfer}}
I_t^{'} =  \mathbf{\Sigma}_s^{1/2} \mathbf{\Sigma}_t^{-1/2} (I_t - \mathbf{\mu}_t) + \mathbf{\mu}_s,
\end{align}
where $I_t^{'}$ is the recolored output. To avoid re-transfer of color from saturated pixel areas in the source, the mean and covariance matrix are computed from image intensities $<$90\% of the maximum intensity value. Fig.~\ref{fig:crtCorrection} shows the generated results using our trained GAN-based network (on the right) and after color shift correction (bottom) showing very close to ground-truth results. To recover low contrast frames, the CGAN-saturation network was trained with a reverse image pair of the same training data set. 
\subsubsection{Specularity, and other misc. artifacts removal}
\label{subsec:specularity}
Illumination inconsistencies and view point changes cause strong bright spots due to reflections from bubbles and shiny organ surfaces, and water-like substances can create multi-colored chromatic artifacts (referred to as `imaging or mixed artifact` in this paper). These inconsistencies appear as  a combination of linear (e.g., additive noise $\eta$) and non-linear noise (function $F(.)$) in Eq.~(\ref{eq:corruptedImage}). A process referred to \textit{inpainting} that uses the information of the surrounding pixels as prior information is used to replace the saturated pixels in affected regions. 
TV-inpainting methods are popular for restoring images with geometrical structures~\cite{Shen:SIAM02} and patch-based methods~\cite{AlexeiFreeman:SIGGRAPH01} for texture synthesis. However, these methods are computationally expensive. Recent advances in deep neural networks have proven to recover visually plausible image structures and textures~\cite{Kohler:GCPR14} with almost real-time performance. However, they are limited to the size of the mask or the number of unknown pixels in an image. In this context, GANs~\cite{pathak:CVPR16,Iizuka:SIGGRAPH17,Jiahui:arXiv18} have been shown to be more successful in providing faster and more coherent reconstructions even with larger masks. Both contextual and generative losses have been used in these methods. Iizuka et al.~\cite{Iizuka:SIGGRAPH17} and Yu et al. \cite{Jiahui:arXiv18} used local and global discriminators to improve the reconstruction quality. To enlarge the network receptive field~\cite{Jiahui:arXiv18} further used a coarse-to-fine network architecture using WGAN-GP instead of the DCGAN in~\cite{Iizuka:SIGGRAPH17}. Additionally, a $\mathit{l}_1$ discounted contextual (reconstruction) loss using a distance-based weight mask was used for added regularization~\cite{Jiahui:arXiv18}. Due to the reduced training time and better reconstruction quality compared to~\cite{Iizuka:SIGGRAPH17}, we use the network proposed in~\cite{Jiahui:arXiv18} for inpainting. 

We use a bottleneck approach to retrain the model initialised with the pretrained weights of the places2 data set~\cite{zhou:PAMI17places}. To capture the large visual variations present in endoscopy images, 1000 images from 7 different patient endoscopy videos with a quality score $>$95\% were used as the `clean' images (see Section~\ref{sec:QS}). We used 172 images as a validation set during the training. Both training and validation sets included multimodal endoscopic video frames.  During training and validation masks of different patch sizes $\{(5\times 5), (7\times, 7), (11\times 11), (13\times 13),..., (33\times33)\}$ were randomly generated and were used for restoration. A single image can have one or multiple generated masks for restoration.
\section{Experiments}{\label{sec:experiments}}
\subsection{Quality assessment metrics}
To evaluate our artifact detection we use the standard mean average precision (mAP) and intersection-over-union (IoU) metrics. We quantitatively compare the detection results of all architectures using the mAP at IoU thresholds for a positive match of 5\%, 25\% and 50\% denoted mAP$_5$, mAP$_{25}$ and mAP$_{50}$ respectively, the mean IoU between positive matches, the number of predicted boxes relative to the number of annotated boxes and the average inference time for one image as quantitative measures. For the quality assessment of deblurring methods we use peak signal-to-noise ratio (PSNR) and structural similarity (SSIM) measures. To overcome the limitations of PSNR for quantification of saturation and specularity restoration tasks we include more sophisticated visual information fidelity (VIF,~\cite{Sheikh:TIP06}) and relative edge coherence (RECO,~\cite{Baroncini:ESPC09}) quality assessment metrics that are independent of the distortion type. 
\subsection{Artifact detection}
\begin{figure}[t!]
    \centering
    \begin{minipage}[b]{0.32\linewidth}
    \includegraphics[width=\linewidth]{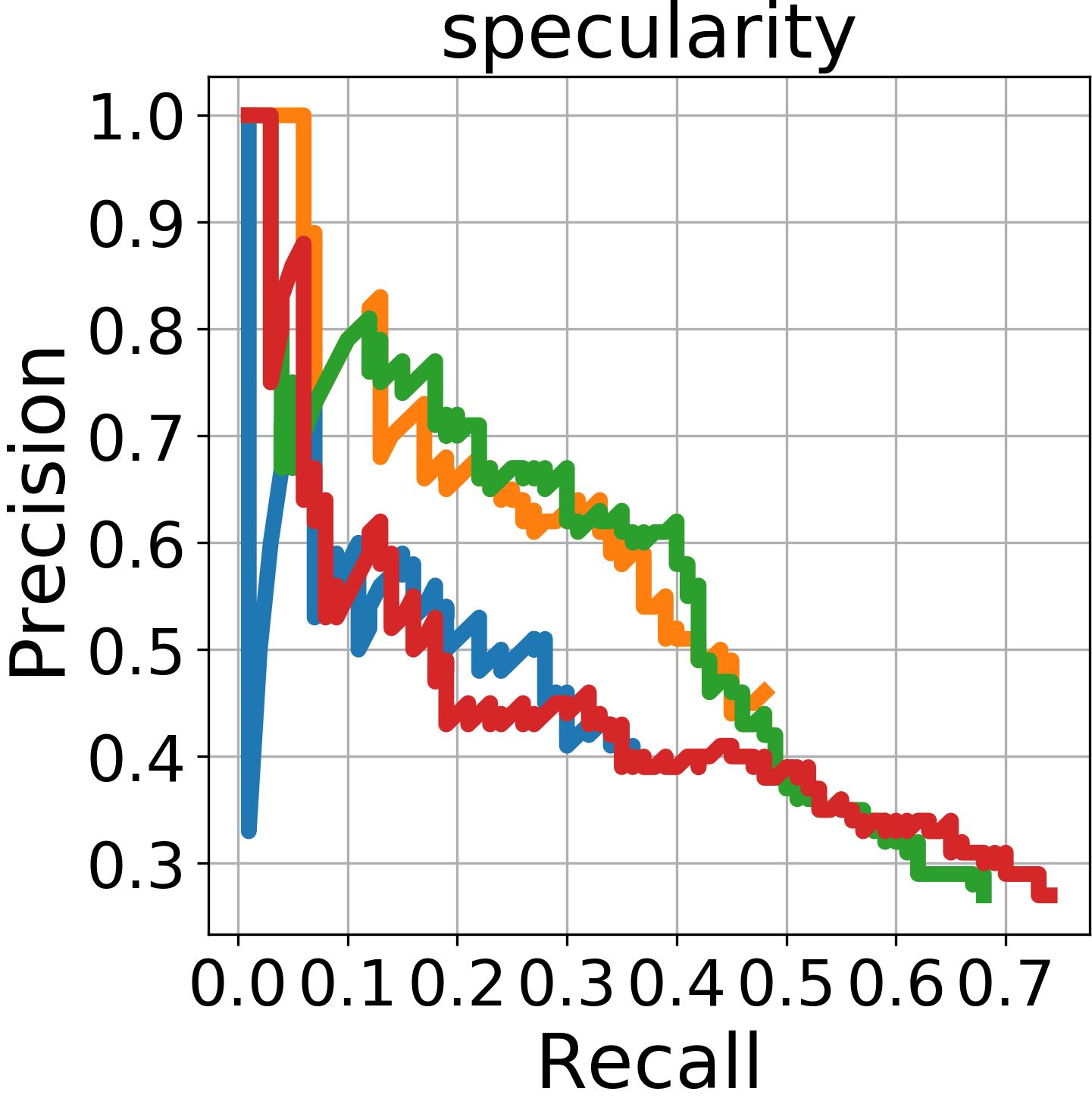}
    \centerline{}
    \end{minipage}
    \begin{minipage}[b]{0.32\linewidth}
    \includegraphics[width=\linewidth]{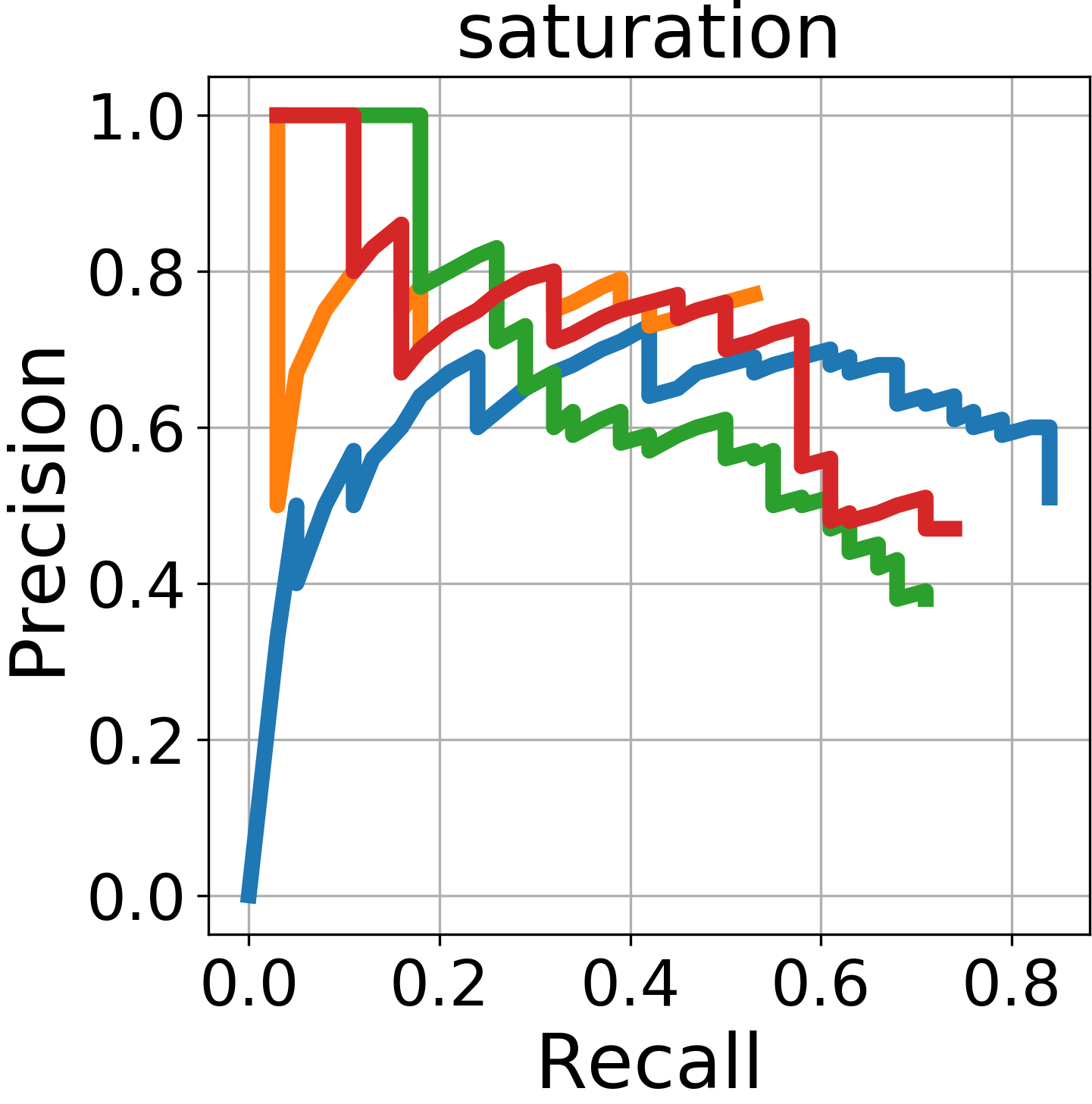}
    \centerline{}
    \end{minipage}
    \begin{minipage}[b]{0.32\linewidth}
    \includegraphics[width=\linewidth]{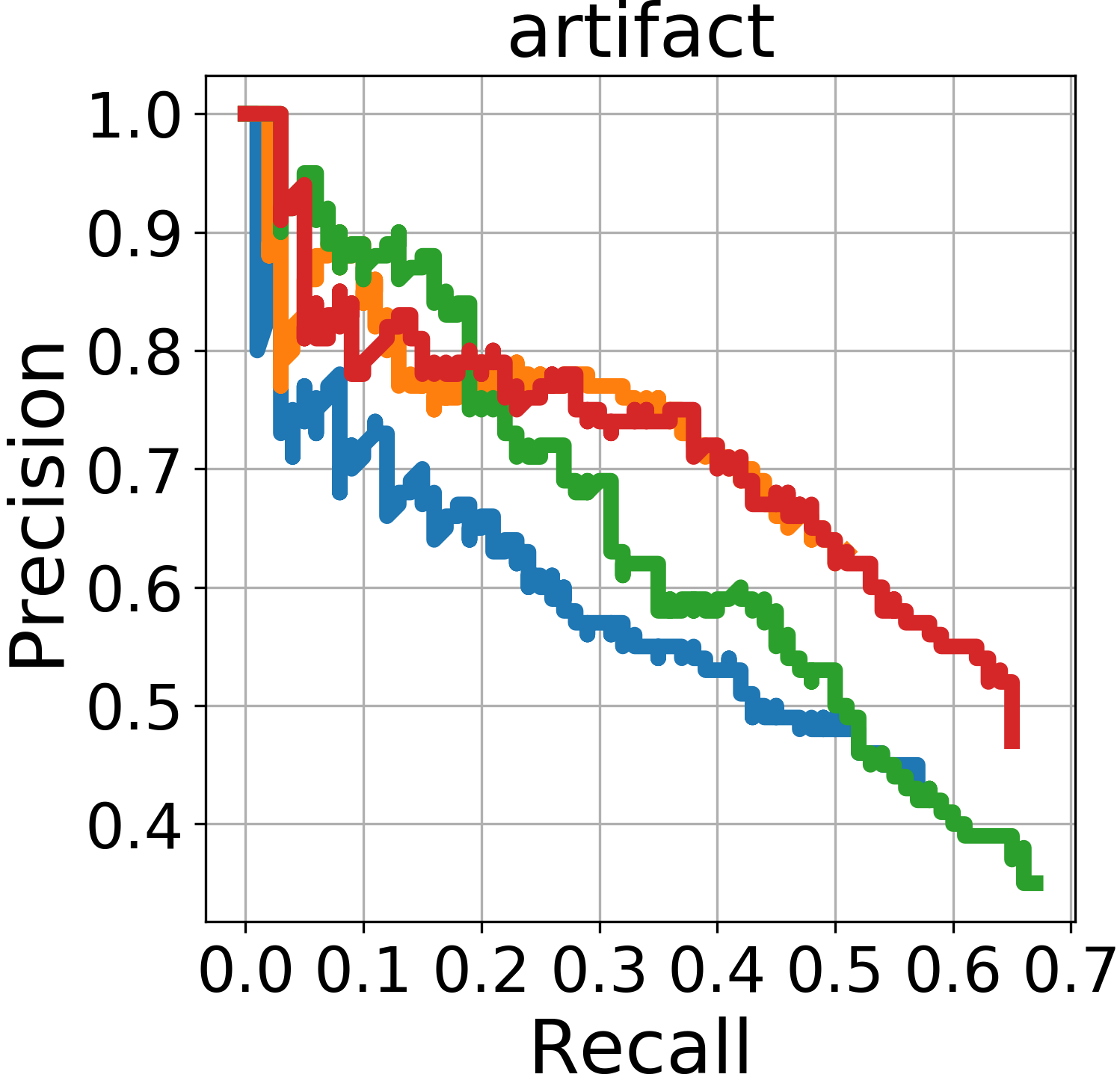}
    \centerline{}
    \end{minipage}
    \begin{minipage}[b]{0.32\linewidth}
    \includegraphics[width=\linewidth]{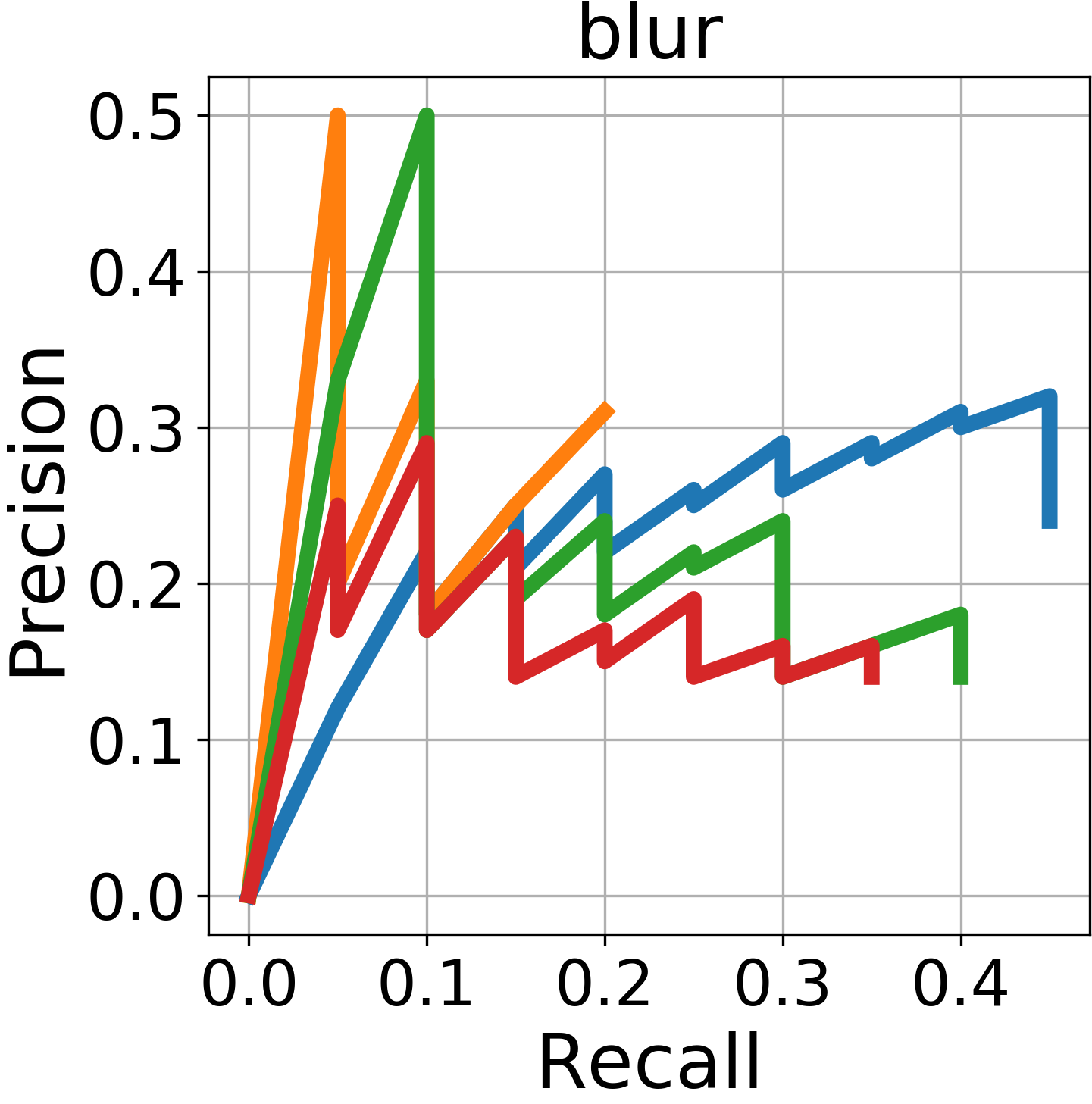}
    \end{minipage}
    \begin{minipage}[b]{0.32\linewidth}
    \includegraphics[width=\linewidth]{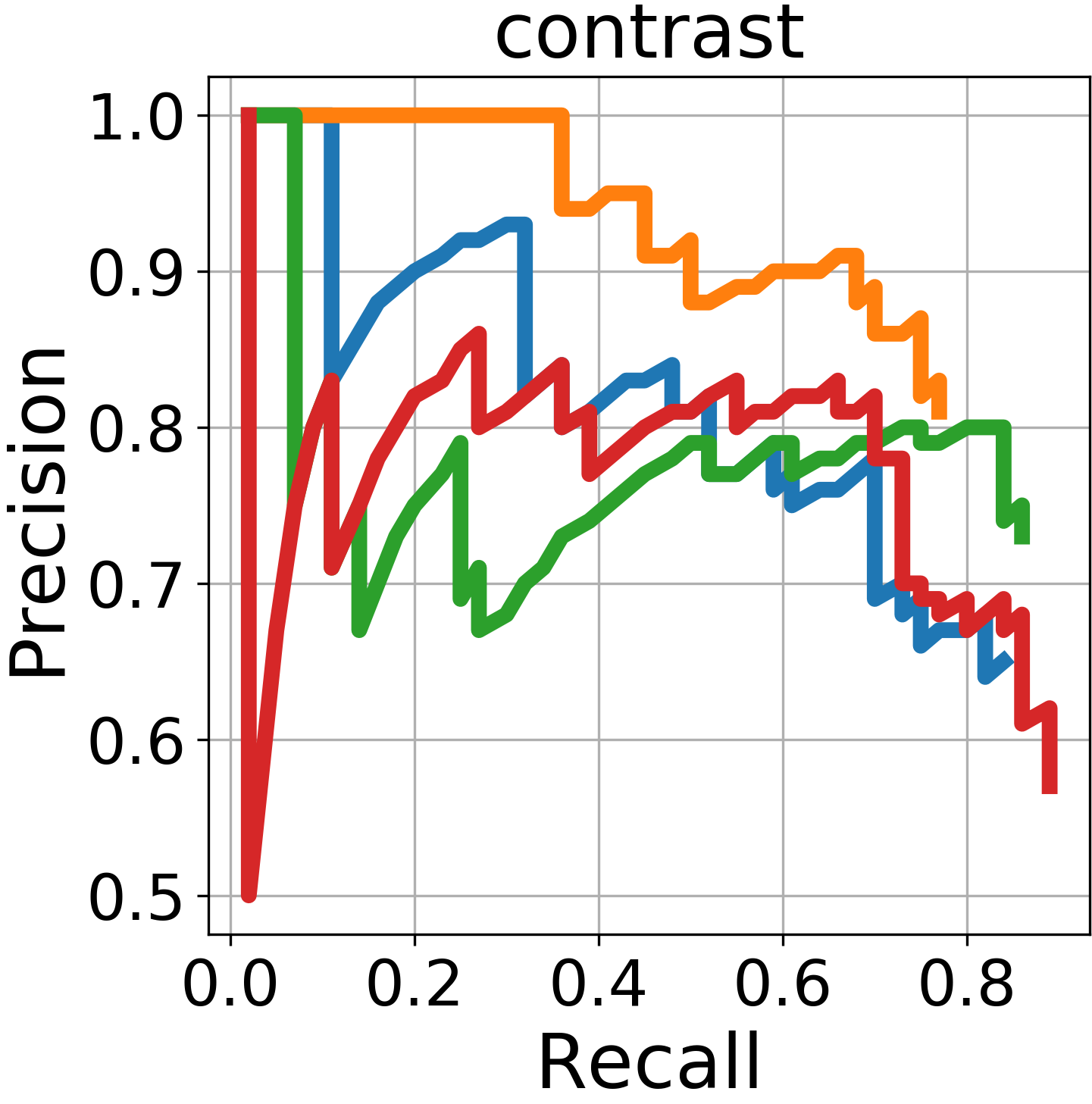}
    \end{minipage}
    \begin{minipage}[b]{0.32\linewidth}
    \includegraphics[width=\linewidth]{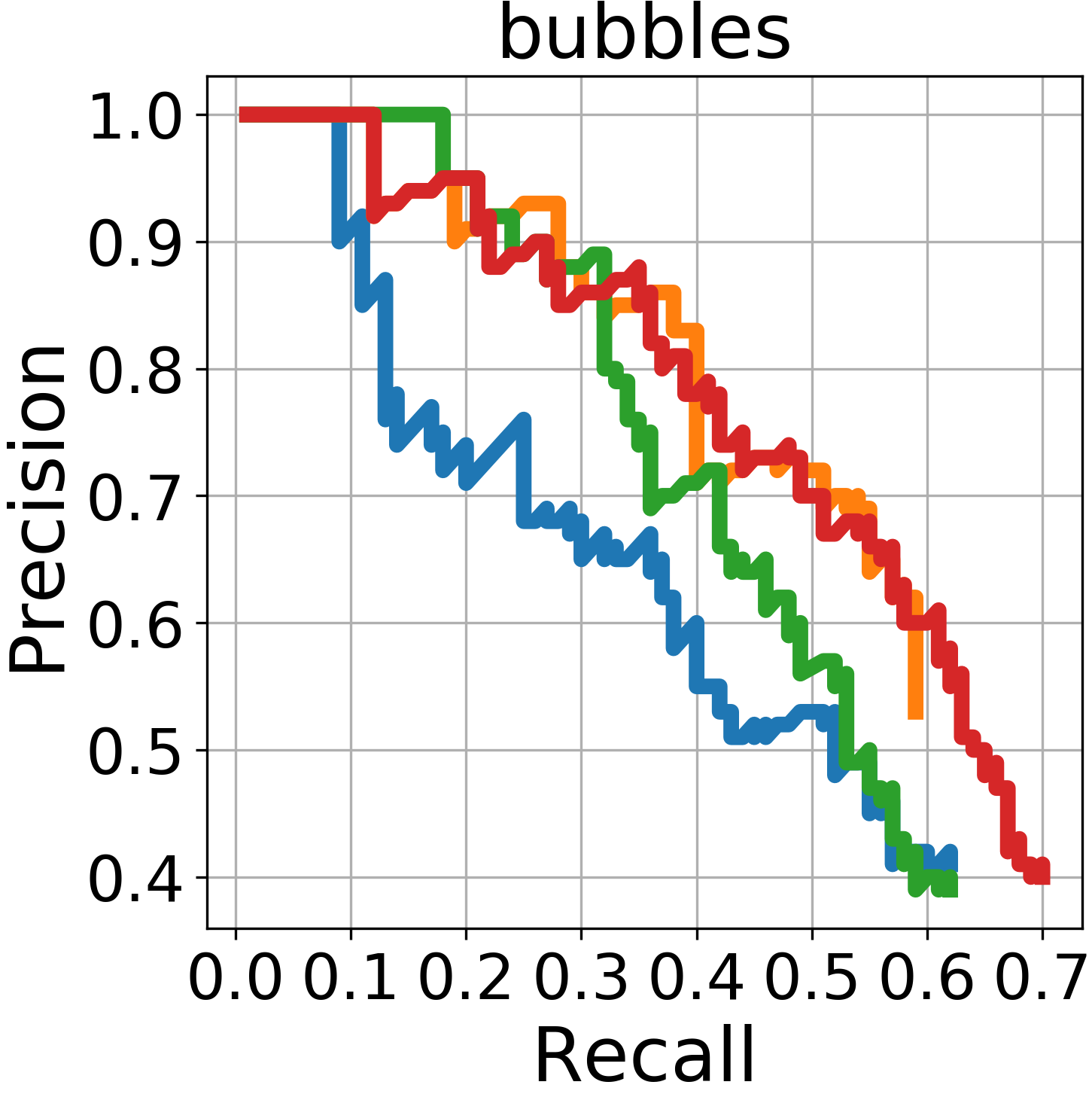}
    \end{minipage}
    \begin{minipage}[b]{.8\linewidth}
    	\includegraphics[width=\linewidth]{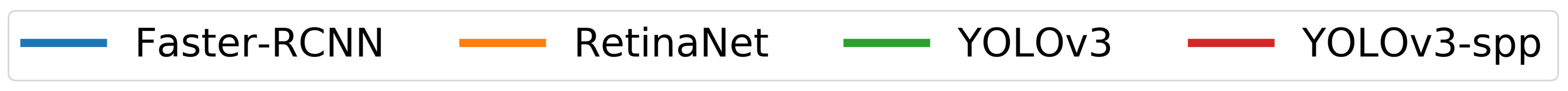}
    \end{minipage}
    
	\caption{Class specific precision-recall curves for artifact detection.}
	\label{fig:precision-recall_YOLO_endo}
\end{figure}

{\label{sec:artifact_exp}}
Table~\ref{tab:map_object_detection} shows that YOLOv3 variants outperform both Faster R-CNN and Retinanet. YOLOv3-spp (proposed) yields the best mAP of 49.0 and 45.7 at IoU thresholds of 0.05 and 0.25 respectively at a detection speed $\approx 6 \times$ faster than Faster R-CNN~\cite{ren2015faster}.  Even though Retinanet exhibits the best IoU of 38.9, it is to be noted that IoU is sensitive to annotator variances in bounding box annotation which might not resemble the performance of detectors. In terms of class-specific performance, from Fig.~\ref{fig:precision-recall_YOLO_endo} and Table.~\ref{tab:average_precision} proposed YOLOv3-spp is the best across detecting misc. artifacts and bubbles (both are predominantly present in endoscopic videos) with average precision of 48.0 and 55.9, respectively. Faster R-CNN yielded the highest average precision for saturation (71.0) and blur (14.5) while RetinaNet and YOLOv3 outperformed respectively for contrast (73.6) and specularity detection (40.0). It is worth noting that proposed YOLOv3-spp yielded second best average precision scores for speculariy (34.7), saturation (55.7) and contrast (72.1). 
\begin{table*}[t!]
	\centering
	\begin{tabular}{|m{2.6cm}|>{\centering\arraybackslash}m{1.4cm}|>{\centering\arraybackslash}m{1.2cm}|>{\centering\arraybackslash}m{1cm}|>{\centering\arraybackslash}m{1cm}|>{\centering\arraybackslash}m{1cm}|>{\centering\arraybackslash}m{1cm}|>{\centering\arraybackslash}m{1cm}|>{\centering\arraybackslash}m{0.8cm}|}
		\hline
		Method & Backbone & Input Size & mAP$_{5}$ & mAP$_{25}$ & mAP$_{50}$ & IoU$_{25}$ & Predict Boxes & Time (ms)\\
		\hline\hline
		Faster R-CNN\cite{ren2015faster} & Resnet50 & 600x600 & {44.9} & 40.4 & 29.5 & 28.3 & 835 & 555\footnote{Python Keras 2.0, (Tensorflow 1.2 backend) Code.}\\
		RetinaNet\cite{lin2017focal} & Resnet50 & 608x608 & 43.8 & 41.2 & 34.7 & {38.9} & {576} & 103\footnote{PyTorch 0.4 Code.} \\ 
		\hline
		YOLOv3\cite{redmon2018yolov3} & darknet53 & 512x512 & 47.4 & 44.3 &  {35.1} & 24.2 & 1252 & 95\footnote{Python call of Darknet trained network.}\\ 
		YOLOv3 &darknet53&608x608 & 48.1 & 45.2 & 33.2 & 21.4 & 1300 &  116$^4$\\
		YOLOv3-spp& darknet53 &512x512 & {49.0} & {45.7} & 34.7 & 24.4 & 1120 & {88}\\
		\hline
	\end{tabular}
	\caption{Artifact detection results on test set with different neural network architectures. All timings are reported on a single 6GB NVIDIA GTX Titan Black GPU and is the average time for a single 512x512 image (possibly rescaled on input as indicated) evaluated over all 129 test images. Total number of ground truth boxes = 644 boxes.}
	\label{tab:map_object_detection}
\end{table*}
%

\begin{table*}[t!h!]
	\begin{center}
		\begin{tabular}{|l|c|c|c|c|c|c|}
			\hline
			Method & Spec. & Sat. & Arte. & Blur & Cont. & Bubb.\\
			\hline\hline
			Faster R-CNN\cite{ren2015faster} & 20.7 & \textbf{71.0} & 35.1 & \textbf{14.5} & 58.7 & 42.4 \\
			RetinaNet\cite{lin2017focal} & 33.1 & 42.9 & 39.8 & 7.2 & \textbf{73.6} & 50.6 \\ 
			\hline
			YOLOv3\cite{redmon2018yolov3} & \textbf{40.0} & 50.4 & 44.3 & 11.6 & 70.8 & 48.9 \\ 
			YOLOv3-spp & 34.7 & 55.7 & \textbf{48.0} & 7.5 & 72.1 & \textbf{55.9} \\ 
			\hline
		\end{tabular}
	\end{center}
	\caption{Class-specific average precision (AP) of the different object detection networks.}
	\label{tab:average_precision}
\end{table*}
\subsection{Frame restoration}
%
\begin{table*}[t!h!]
	\centering
	\begin{tabular}{ |p{2.0cm}|p{0.7cm}|p{0.6cm}|p{0.6cm}p{0.6cm}|p{0.6cm}|p{0.6cm}|  }
		\hline
		Method & Metric &\multicolumn{5}{c|}{Images with varying motion blur}\\
		\cline{3-7}
		&&\multicolumn{1}{c|}{\#80}&\multicolumn{1}{c|}{\#99}&\multicolumn{1}{c|}{\#102}&\multicolumn{1}{c|}{\#113}&\multicolumn{1}{c|}{\#116}\\
		\hline
		\hline
		CGAN$+$cont. \&   &  \footnotesize{PSNR} &\multicolumn{1}{c|}{\textbf{25.22}} &\multicolumn{1}{c|}{\textbf{28.14}} &\multicolumn{1}{c|}{\textbf{27.28}} &\multicolumn{1}{c|}{\textbf{23.41}} &\multicolumn{1}{c|}{\textbf{24.81}} \\
		\cline{2-7}
		\small{HF feature loss} &  \footnotesize{SSIM} &\multicolumn{1}{c|}{\textbf{0.998}} &\multicolumn{1}{c|}{\textbf{0.997}} &\multicolumn{1}{c|}{\textbf{0.993}} &\multicolumn{1}{c|}{\textbf{0.980}} &\multicolumn{1}{c|}{\textbf{0.992}} \\
		\hline
		deblur& \footnotesize{PSNR} &\multicolumn{1}{c|}{{25.17}} &\multicolumn{1}{c|}{{27.93}} &\multicolumn{1}{c|}{{26.96}} &\multicolumn{1}{c|}{{23.40}} &\multicolumn{1}{c|}{\textbf{24.81}} \\
		\cline{2-7}
		GAN \cite{deblurGAN}&  \footnotesize{SSIM} &\multicolumn{1}{c|}{\textbf{0.998}} &\multicolumn{1}{c|}{\textbf{0.997}} &\multicolumn{1}{c|}{{0.992}} &\multicolumn{1}{c|}{{0.979}} &\multicolumn{1}{c|}{\textbf{0.992}} \\
		\hline
		SRN&  \footnotesize{PSNR} &\multicolumn{1}{c|}{24.61} &\multicolumn{1}{c|}{27.50} &\multicolumn{1}{c|}{25.02} &\multicolumn{1}{c|}{22.23} &\multicolumn{1}{c|}{22.00} \\
		\cline{2-7}
		DeblurNet~\cite{tao2018srndeblur}&  \footnotesize{SSIM} &\multicolumn{1}{c|}{0.995} &\multicolumn{1}{c|}{0.996} &\multicolumn{1}{c|}{{0.990}} &\multicolumn{1}{c|}{0.970} &\multicolumn{1}{c|}{0.970} \\
		\hline
		TV-deconv&  \footnotesize{PSNR} &\multicolumn{1}{c|}{24.25} &\multicolumn{1}{c|}{26.72} &\multicolumn{1}{c|}{24.75} &\multicolumn{1}{c|}{21.69} &\multicolumn{1}{c|}{22.20} \\
		\cline{2-7}
		\cite{ipoltvdc}&  \footnotesize{SSIM} &\multicolumn{1}{c|}{0.966} &\multicolumn{1}{c|}{0.994} &\multicolumn{1}{c|}{0.988} &\multicolumn{1}{c|}{0.966} &\multicolumn{1}{c|}{0.983} \\
		\hline
	\end{tabular}
	\vspace{2mm}
	\caption{Peak signal-to-noise ratio (PSNR) and the structural similarity
		measure (SSIM) for randomly selected images with different motion blur.}{\label{tab:blurImage}}
\end{table*}
\subsubsection{Blind deblurring}
We compare our proposed conditional generative adversarial network with added contextual and high-frequency feature losses with deblurGAN~\cite{deblurGAN},  scale-recurrent network-based SRN-DeblurNet~\cite{tao2018srndeblur}, and traditional TV-based method~\cite{ipoltvdc}. TV regularizion weight $\lambda$ and the blur kernel $r$ affects the quality of recovered deblurred images~\cite{ipoltvdc}. We chose $\lambda = 10^3$ and $r=2.3$ after a few iterative parameter setting experiments for our data set. We performed retraining for  SRN-DeblurNet~\cite{tao2018srndeblur} and deblurGAN~\cite{deblurGAN} on the same data set used by our proposed deblurring model.
\begin{table}[t!]
	\centering
	\begin{tabular}{ |p{2.5cm}|p{1.5cm}|p{1.0cm}|p{1.0cm}p{1.0cm}|  }
		\hline
		Method & Metric &\multicolumn{3}{c|}{Image sequences}\\
		\cline{3-5}
		&&\multicolumn{1}{c|}{\#1}&\multicolumn{1}{c|}{\#2}&\multicolumn{1}{c|}{\#3}\\
		\hline
		\hline
		CGAN$+$cont. \& & \footnotesize{$\overline{PSNR}$} &\multicolumn{1}{c|}{\textbf{25.80}} &\multicolumn{1}{c|}{\textbf{24.65}} &\multicolumn{1}{c|}{\textbf{21.25}}  \\
		\cline{2-5}
		 HF feature-loss&  \footnotesize{$\overline{SSIM}$} &\multicolumn{1}{c|}{\textbf{0.997}} &\multicolumn{1}{c|}{\textbf{0.980}} &\multicolumn{1}{c|}{\textbf{0.970}}  \\
		\hline
		deblur& \footnotesize{$\overline{PSNR}$} &\multicolumn{1}{c|}{{25.68}} &\multicolumn{1}{c|}{{24.37}} &\multicolumn{1}{c|}{{21.08}}  \\
		\cline{2-5}
		CGAN \cite{deblurGAN}&  \footnotesize{$\overline{SSIM}$} &\multicolumn{1}{c|}{{0.996}} &\multicolumn{1}{c|}{{0.977}} &\multicolumn{1}{c|}{{0.968}}  \\
		\hline
	\end{tabular}
	\vspace{0.5cm}
	\caption{Average PSNR ($\overline{PSNR}$) and average SSIM ($\overline{SSIM}$) for image sequences in test trajectories both with added high-frequency (HF) feature loss (proposed) and only contextual loss~\cite{deblurGAN} in conditional GAN model. {\label{tab:blurSeq}}}
\end{table}
\begin{figure}[t!]
	\centering
	\begin{minipage}[b]{0.150\linewidth}
		\includegraphics[scale=0.08]{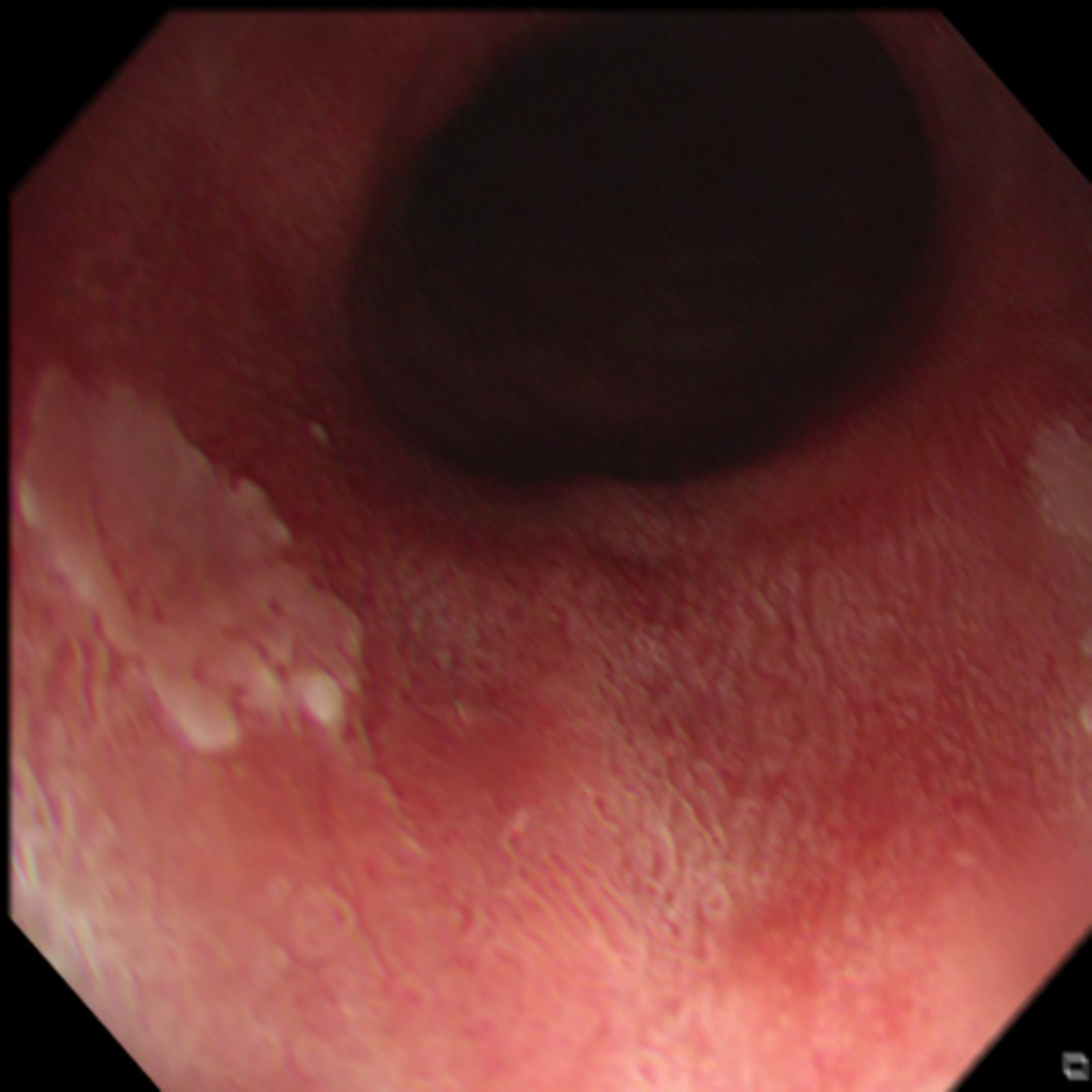}
	\end{minipage}
	\begin{minipage}[b]{0.150\linewidth}
		\includegraphics[scale=0.08]{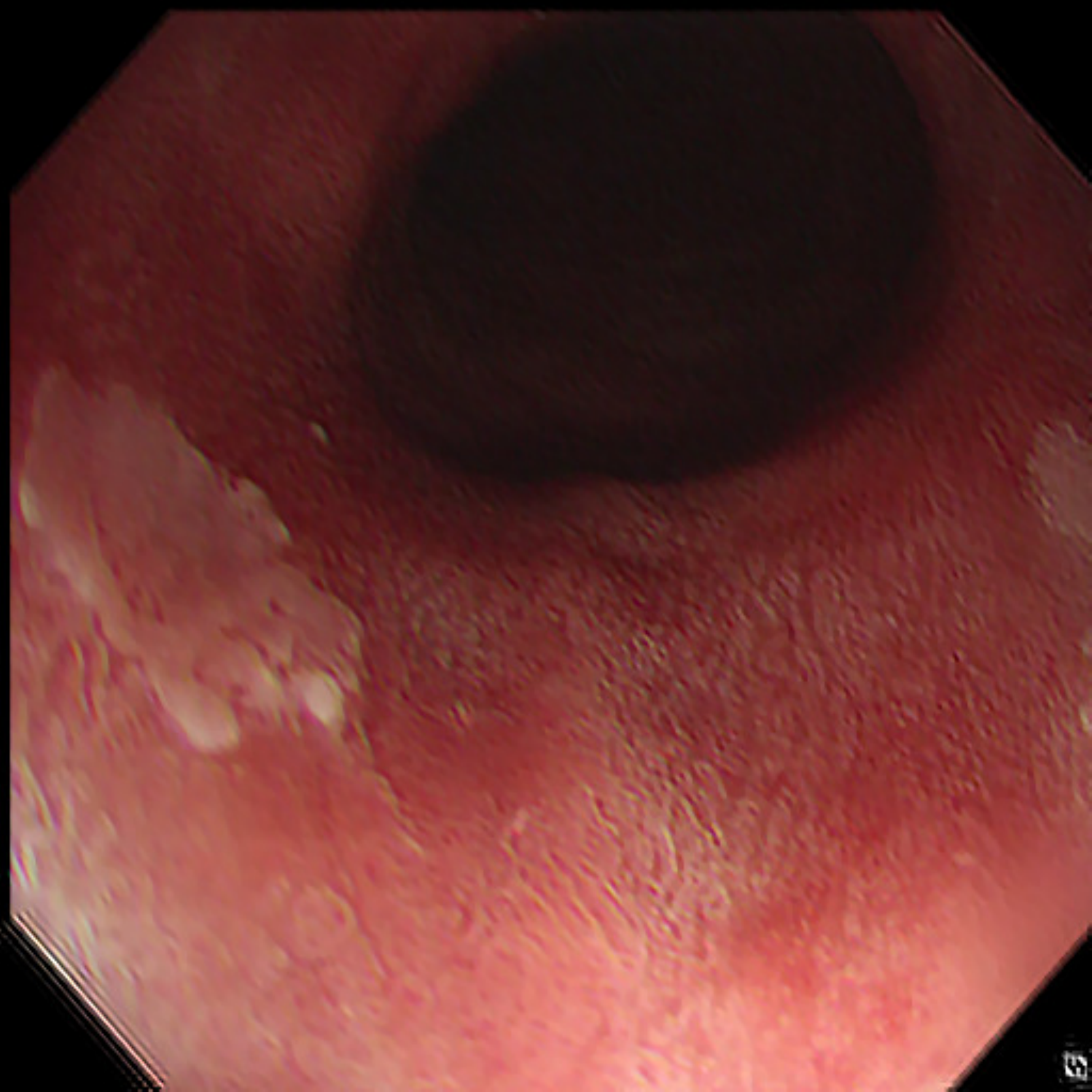}
	\end{minipage}
	\begin{minipage}[b]{0.150\linewidth}
		\includegraphics[scale=0.08]{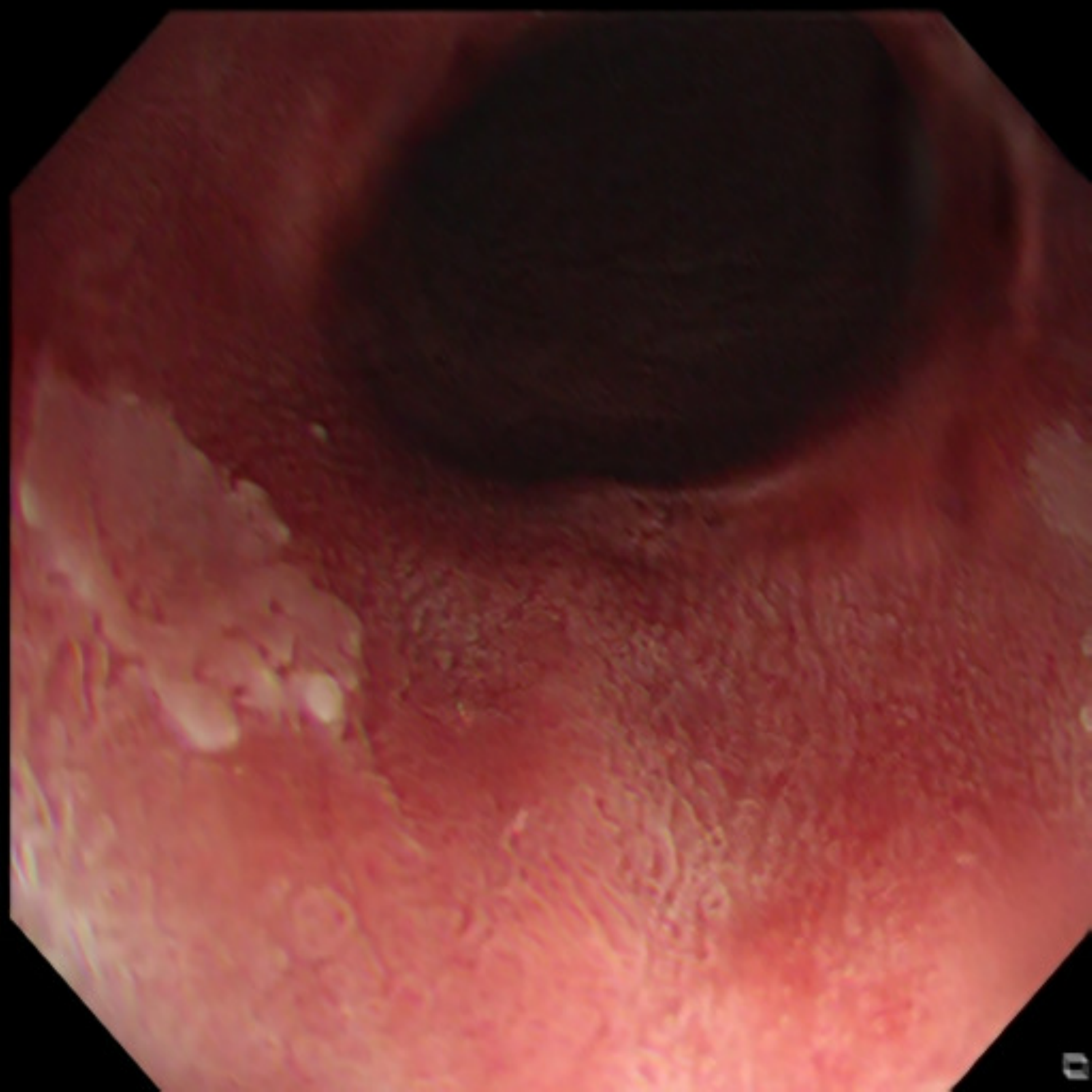}
	\end{minipage}
	\begin{minipage}[b]{0.150\linewidth}
		\includegraphics[scale=0.08]{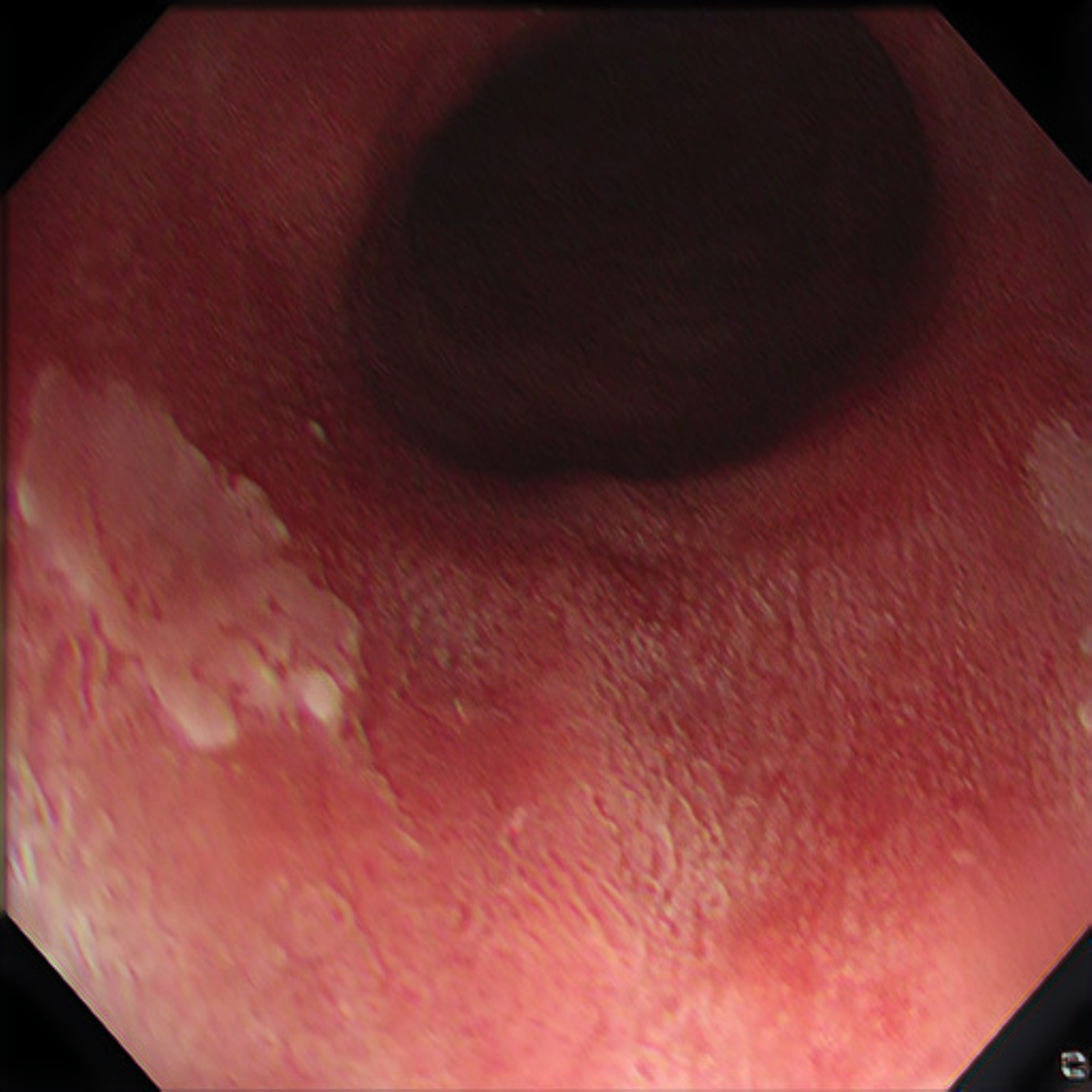}
	\end{minipage}
	\begin{minipage}[b]{0.150\linewidth}
		\includegraphics[scale=0.16]{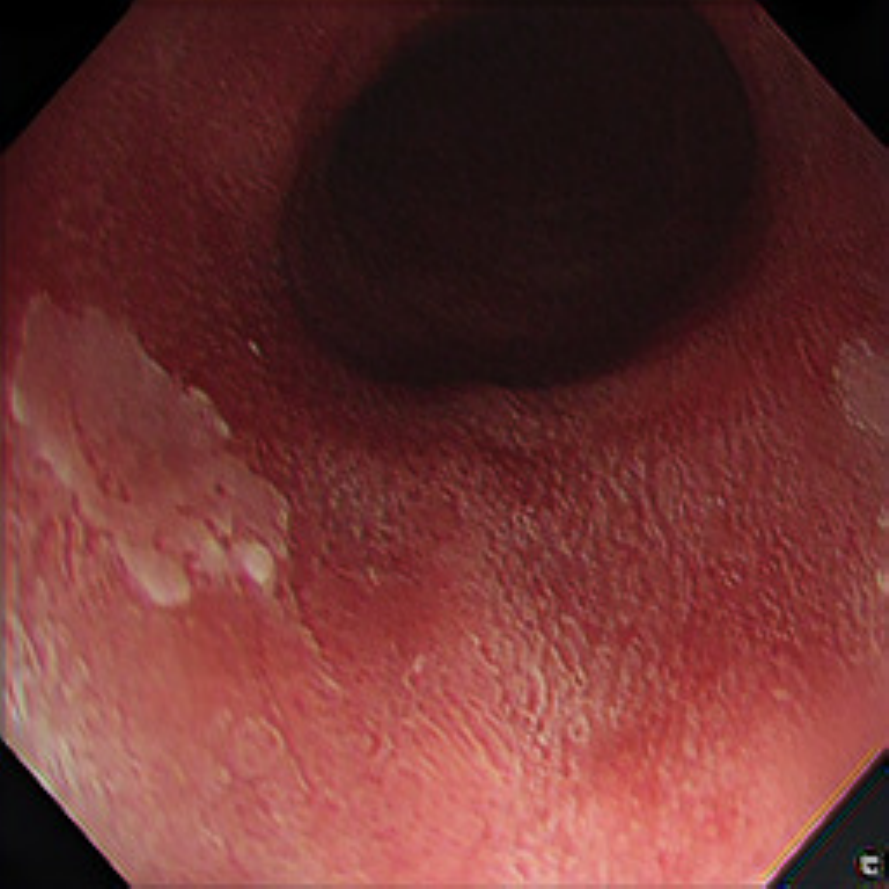}
	\end{minipage}
	\begin{minipage}[b]{0.150\linewidth}
		\includegraphics[scale=0.08]{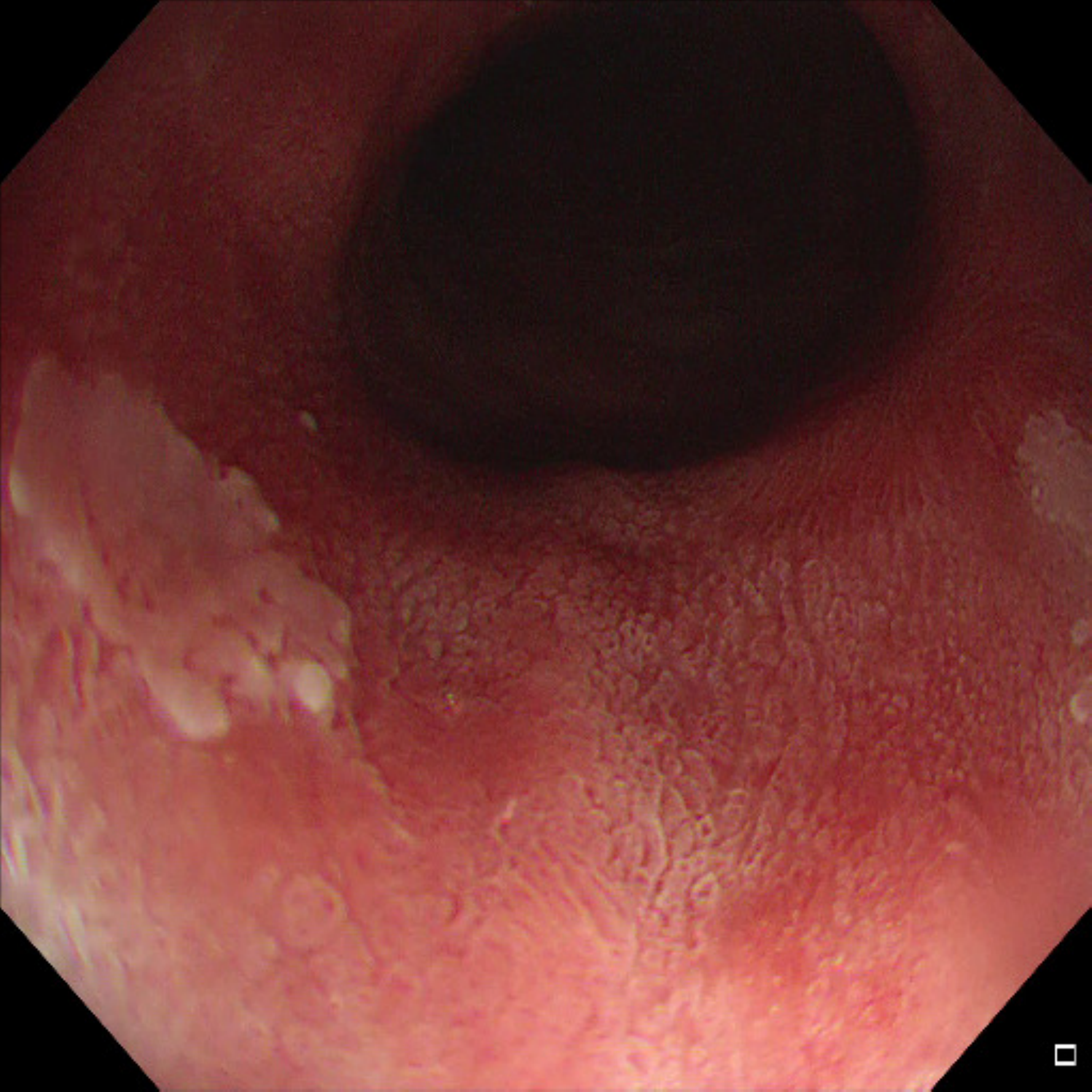}
	\end{minipage}
	
	\vspace{0.15cm}
	\begin{minipage}[b]{0.15\linewidth}
		\includegraphics[scale=0.08]{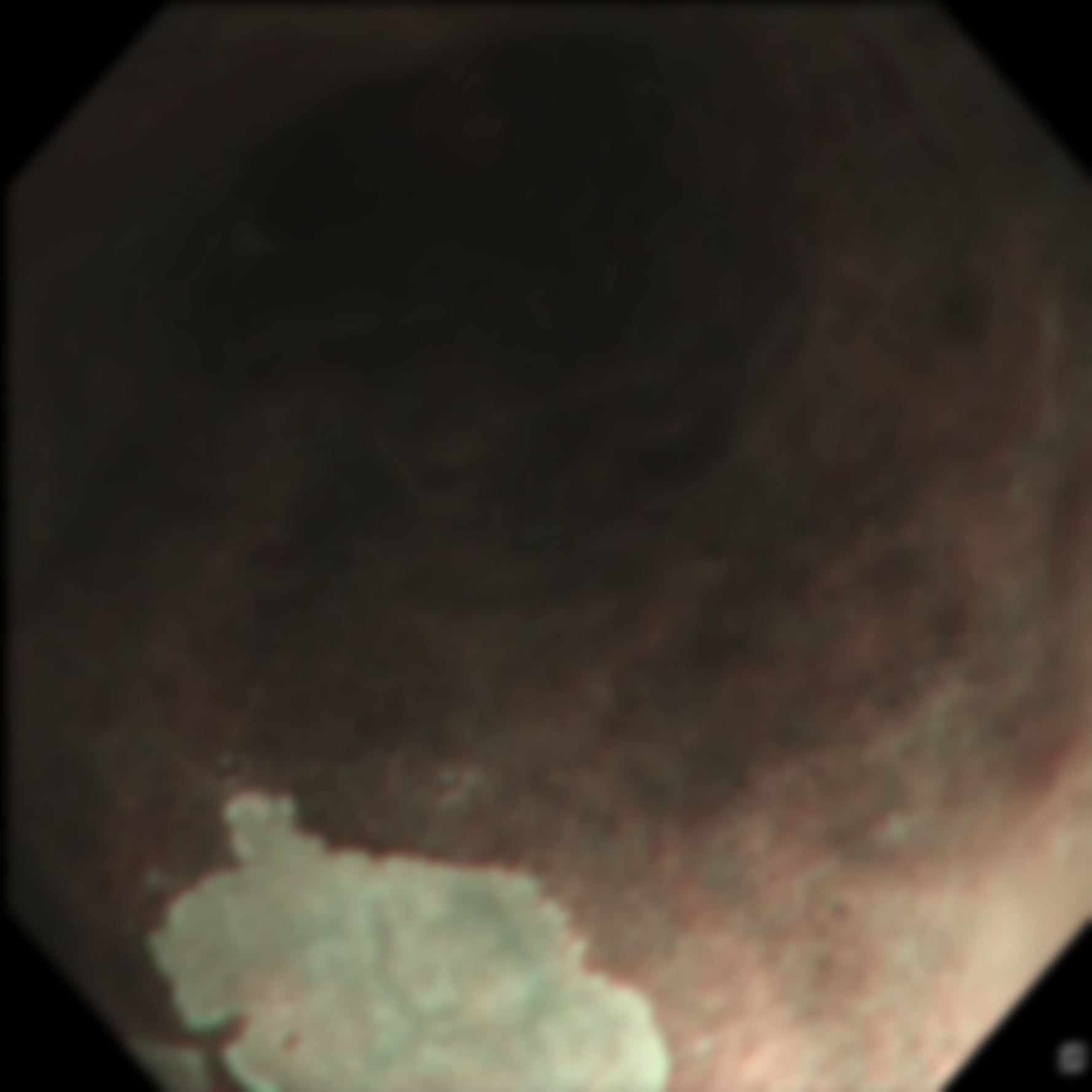}
	\end{minipage}
	\begin{minipage}[b]{0.15\linewidth}
		\includegraphics[scale=0.08]{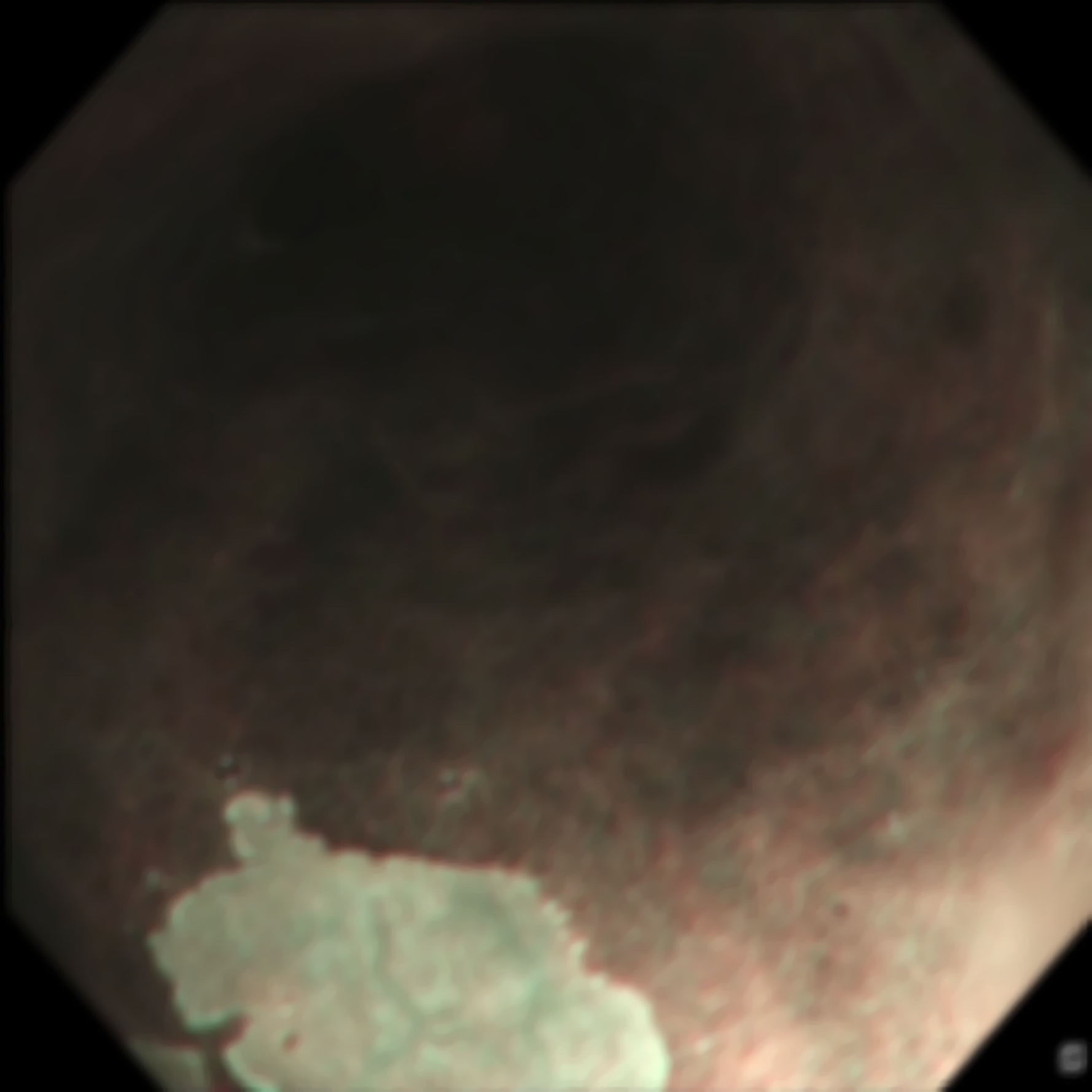}
	\end{minipage}
	\begin{minipage}[b]{0.15\linewidth}
		\includegraphics[scale=0.08]{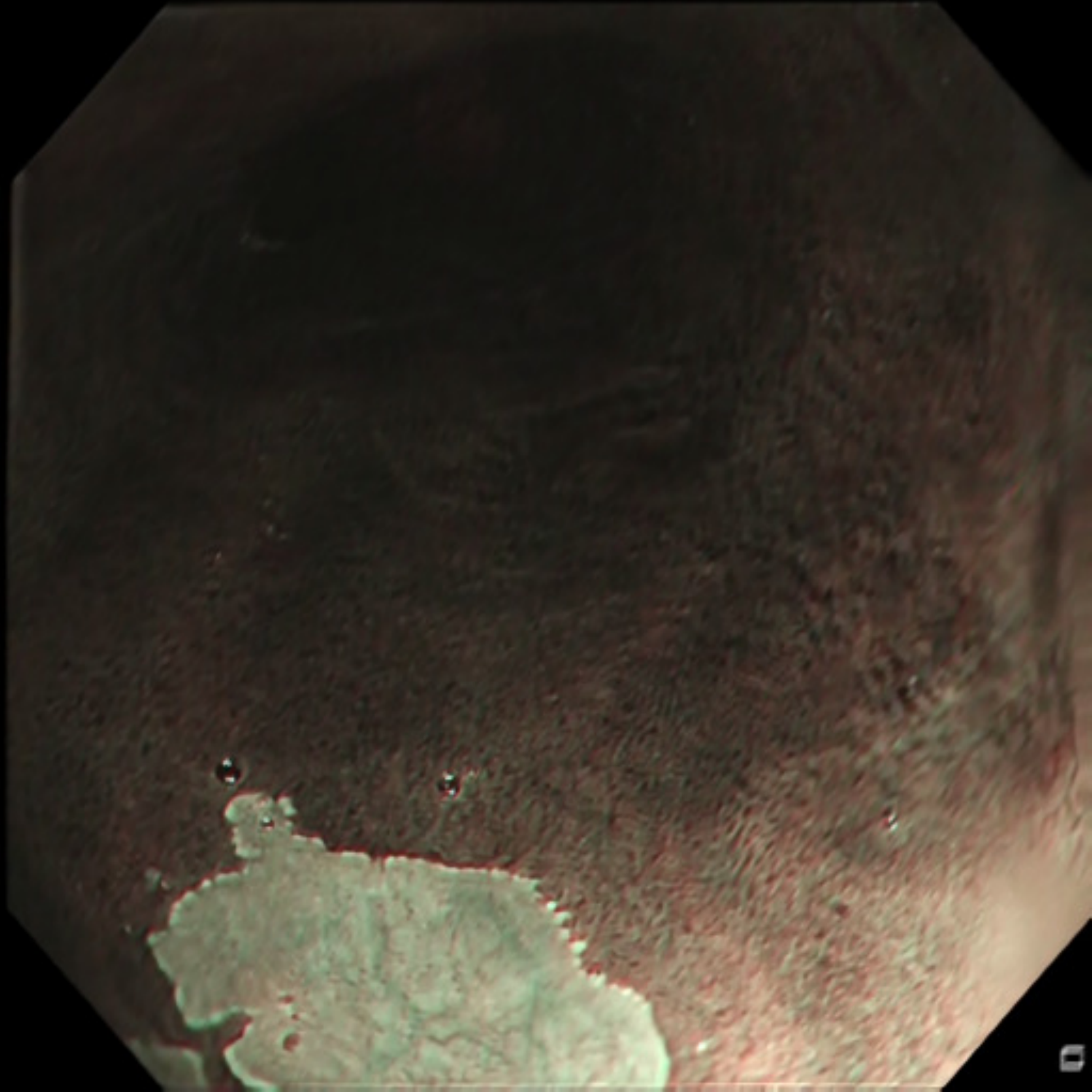}
	\end{minipage}
	\begin{minipage}[b]{0.15\linewidth}
		\includegraphics[scale=0.08]{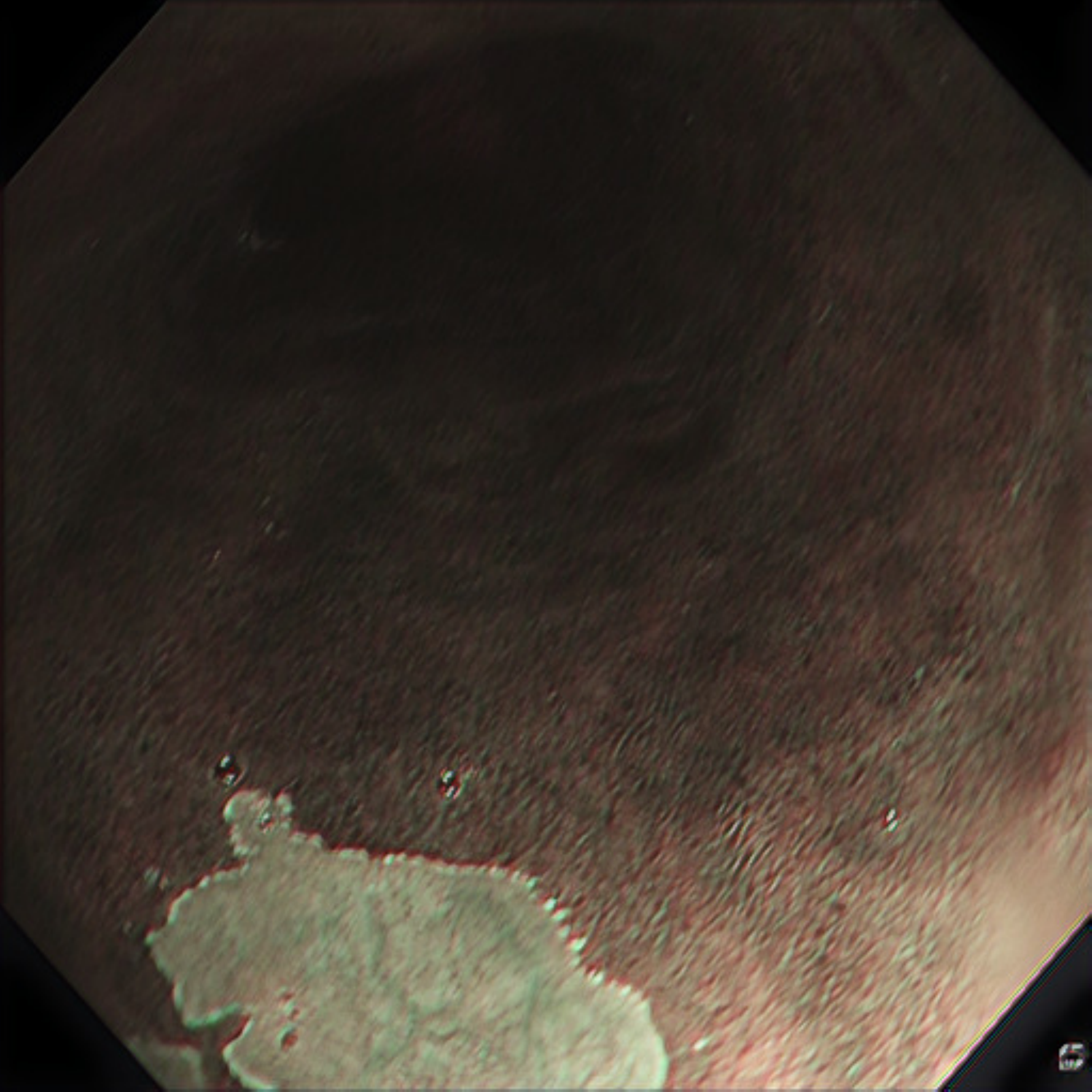}
	\end{minipage}
	\begin{minipage}[b]{0.15\linewidth}
		\includegraphics[scale=0.16]{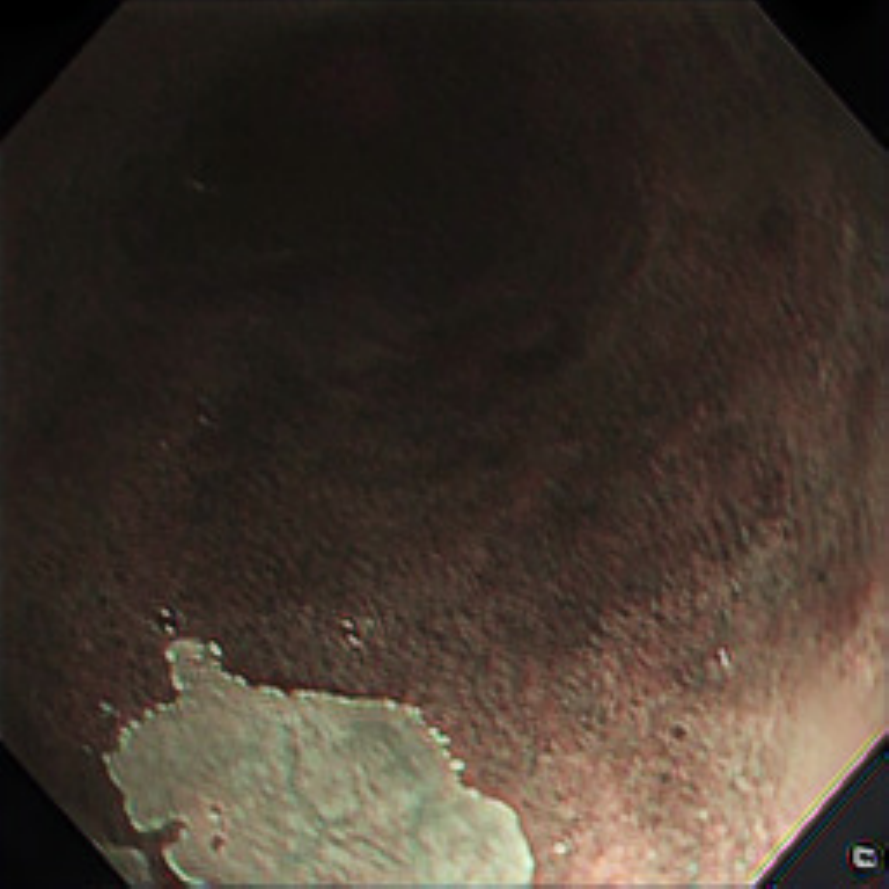}
	\end{minipage}
	\begin{minipage}[b]{0.15\linewidth}
		\includegraphics[scale=0.08]{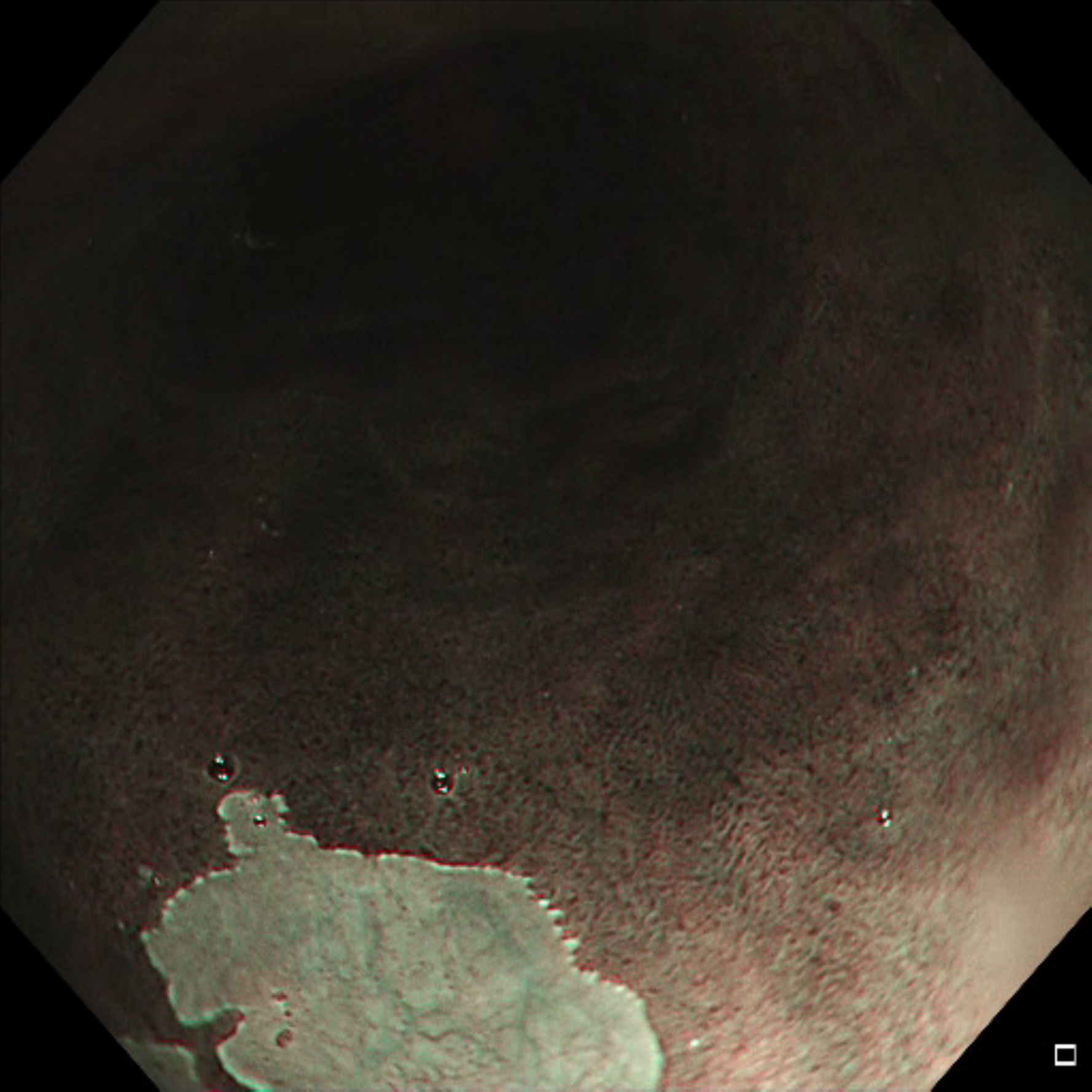}
	\end{minipage}
	
	\begin{minipage}[b]{0.15\linewidth}
		\centerline{\footnotesize{Blurred}}
	\end{minipage}
	\begin{minipage}[b]{0.15\linewidth}
		\centerline{\footnotesize{TV}}
	\end{minipage}
	\begin{minipage}[b]{0.15\linewidth}
		\centerline{\footnotesize{SRN}}
	\end{minipage}
	\begin{minipage}[b]{0.15\linewidth}
		\centerline{\footnotesize{deblur}}
	\end{minipage}
	\begin{minipage}[b]{0.15\linewidth}
		\centerline{\footnotesize{Proposed}}
	\end{minipage}
	\begin{minipage}[b]{0.15\linewidth}
		\centerline{\footnotesize{GT}}
	\end{minipage}
	
	\begin{minipage}[b]{0.15\linewidth}
		\centerline{\footnotesize{}}
	\end{minipage}
	\begin{minipage}[b]{0.15\linewidth}
		\centerline{\footnotesize{deconv}}
	\end{minipage}
	\begin{minipage}[b]{0.15\linewidth}
		\centerline{\footnotesize{DeblurNet}}
	\end{minipage}
	\begin{minipage}[b]{0.15\linewidth}
		\centerline{\footnotesize{GAN}}
	\end{minipage}
	\begin{minipage}[b]{0.16\linewidth}
		\centerline{\footnotesize{}}
	\end{minipage}
	\begin{minipage}[b]{0.15\linewidth}
		\centerline{\footnotesize{}}
	\end{minipage}
	\caption{Qualitative results for different de-blurring methods on WL and NBI frames.}{\label{fig:deblur}}
\end{figure}
We quantitatively evaluated the frame deblurring methods using 5 images with visually large blur and our simulated test trajectories (see Table~\ref{tab:blurImage}) and on 3 different test sequences (simulated motion blur trajectories, see Table~\ref{tab:blurSeq}) each with 30 images. Table~\ref{tab:blurImage} shows that CGAN with $l2$-contextual loss and added high-frequency (HF) feature loss score the highest PSNR and SSIM values for all blurred frames while TV-based deconvolution method~\cite{ipoltvdc} resulted in the least PSNR and SSIM values over all frames. Nearly, 1~\textbf{dB} increase can be seen against the deblurGAN method~\cite{deblurGAN} for frames \#80, \#99, and \#103 while  $\approx$ 2~\textbf{dB} gain can be seen for \#102, \#116 against SRN-DeblurNet~\cite{tao2018srndeblur} using the proposed model. Overall the proposed model yields the best result compared to second best deblurGAN for the blurred image sequences in Table~\ref{tab:blurSeq}. This is also seen qualitatively in Fig.~\ref{fig:deblur}. It can be observed that SRN-DeblurNet deforms the image at upper right in both WL and NBI frames.
\vspace{2mm}
\subsubsection{Saturation removal}
\begin{center}
	\begin{figure}[t!]
		\centering
		\begin{minipage}[b]{0.15\linewidth}
			\includegraphics[width=\linewidth]{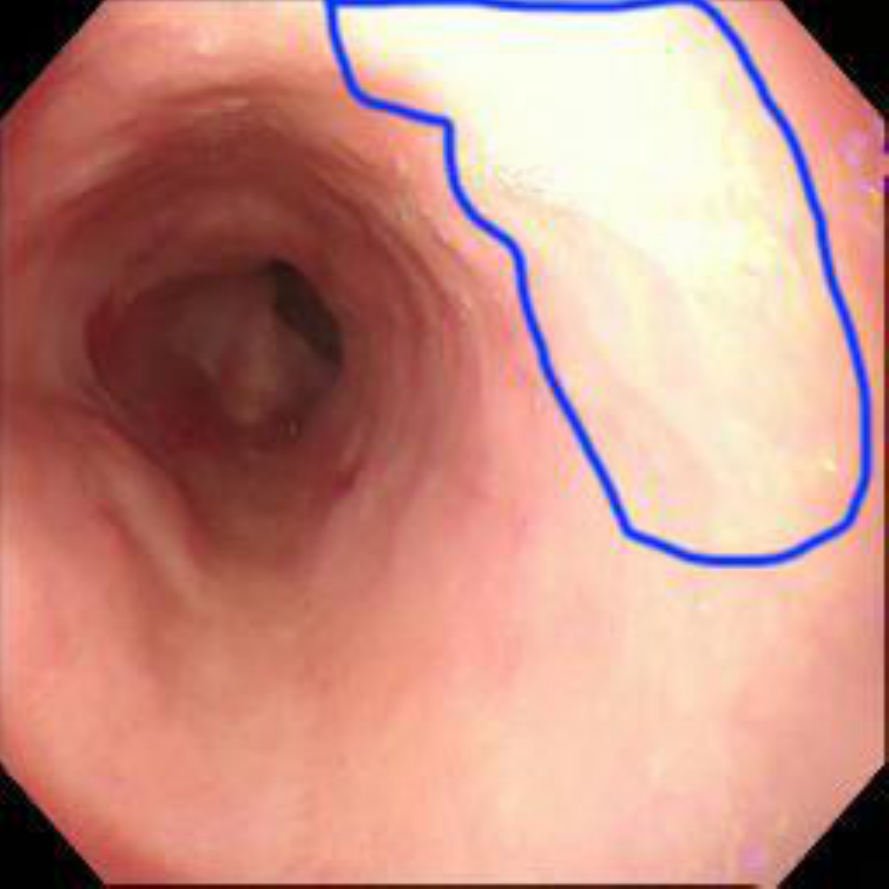}
		\end{minipage}
		\begin{minipage}[b]{0.16\linewidth}
			\includegraphics[width=\linewidth]{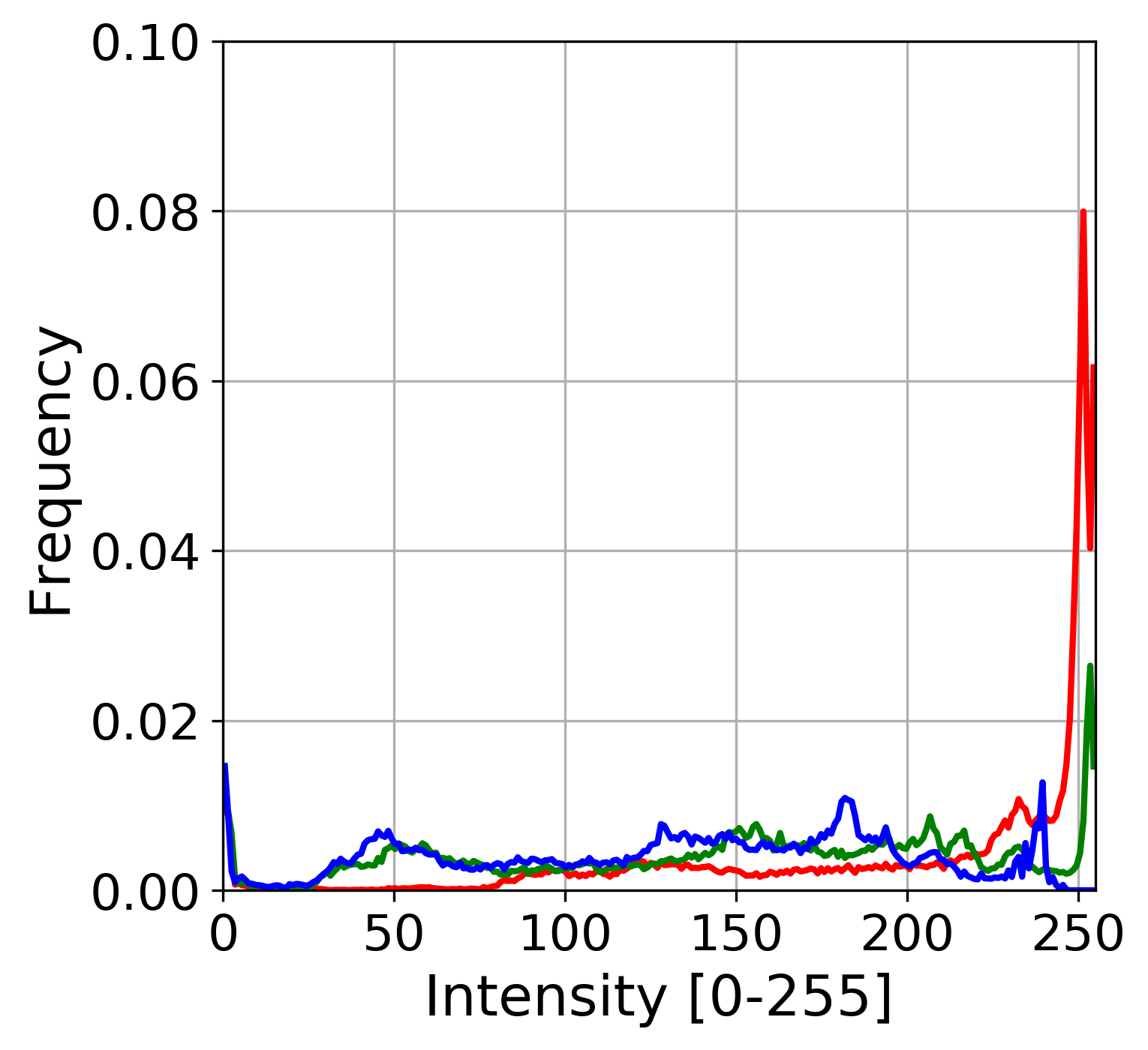}
		\end{minipage}
		\begin{minipage}[b]{0.15\linewidth}
			\includegraphics[width=\linewidth]{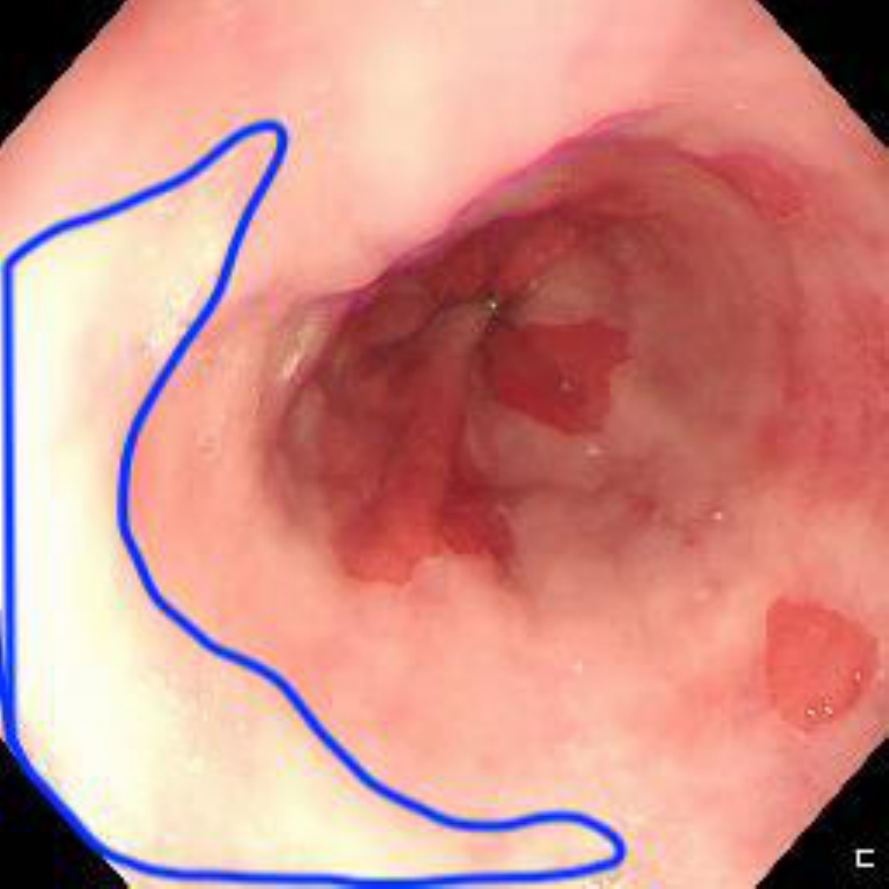}
		\end{minipage}
		\begin{minipage}[b]{0.16\linewidth}
			\includegraphics[width=\linewidth]{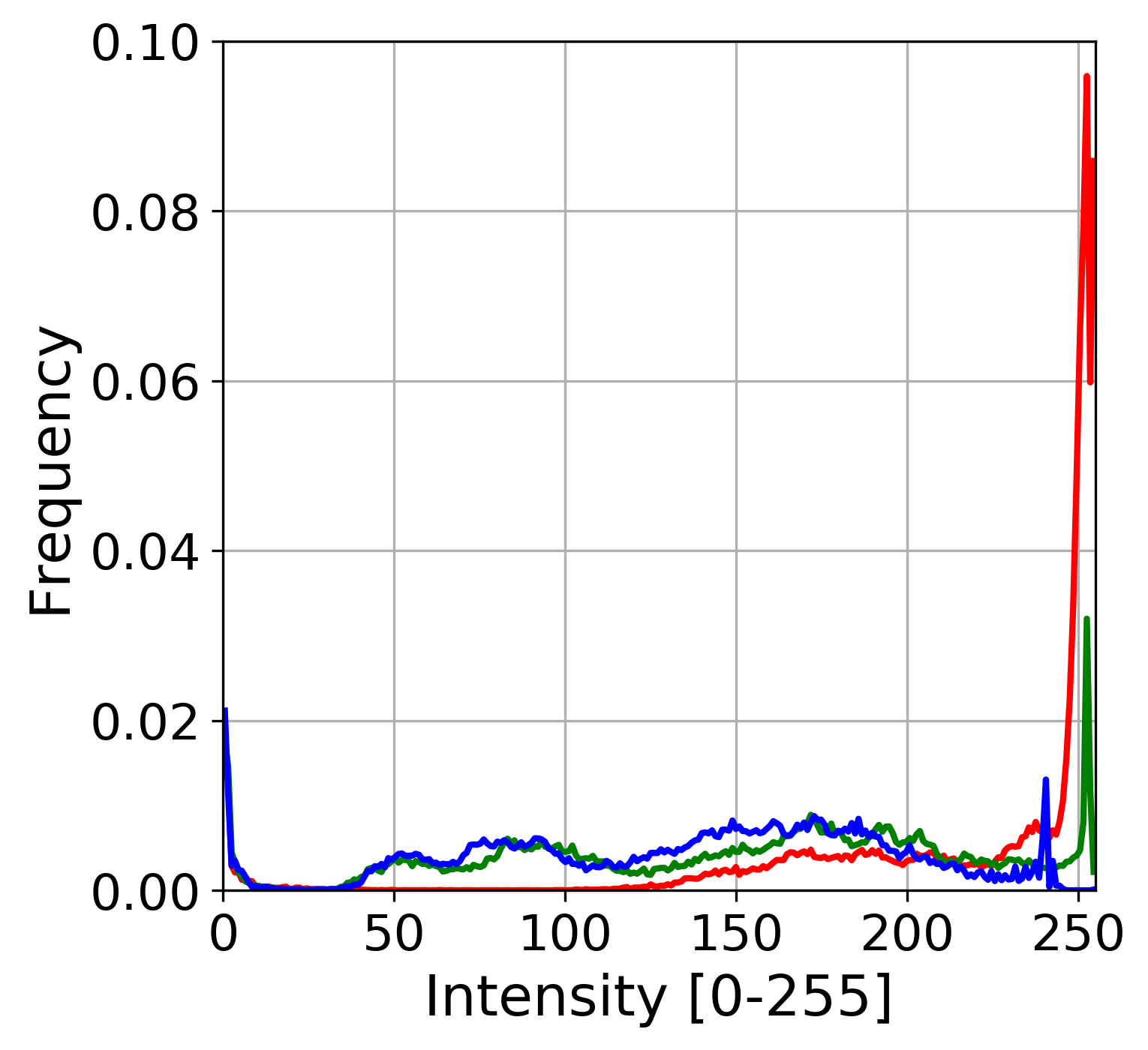}
		\end{minipage}
		\begin{minipage}[b]{0.15\linewidth}
			\includegraphics[width=\linewidth]{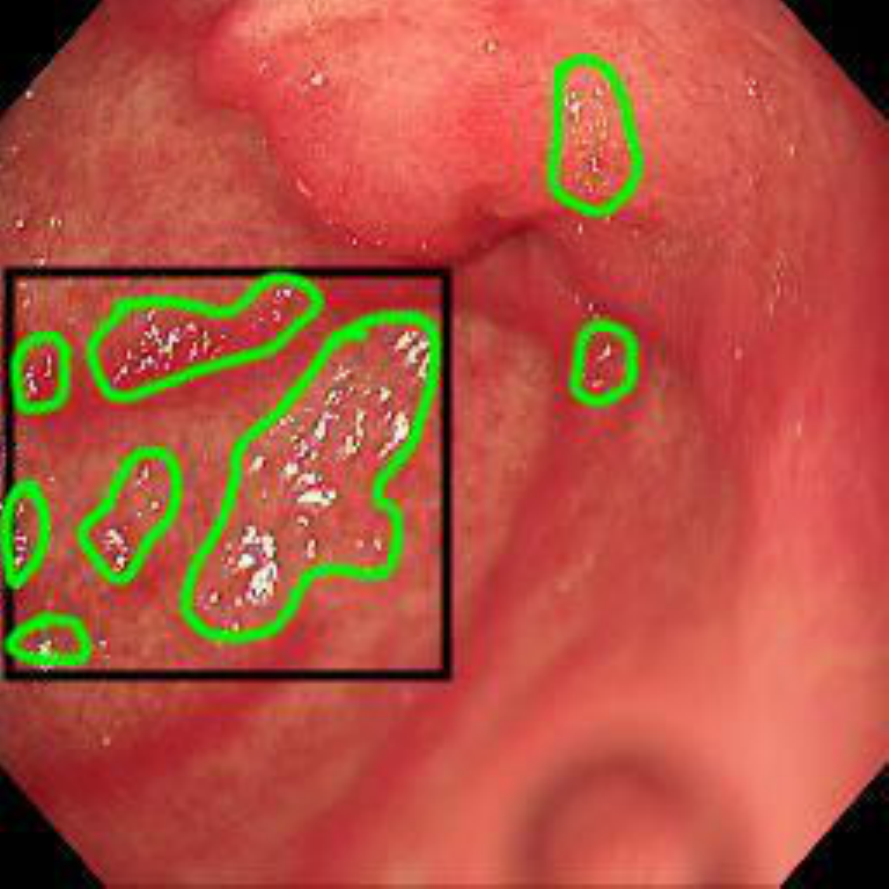}
		\end{minipage}
		\begin{minipage}[b]{0.16\linewidth}
			\includegraphics[width=\linewidth]{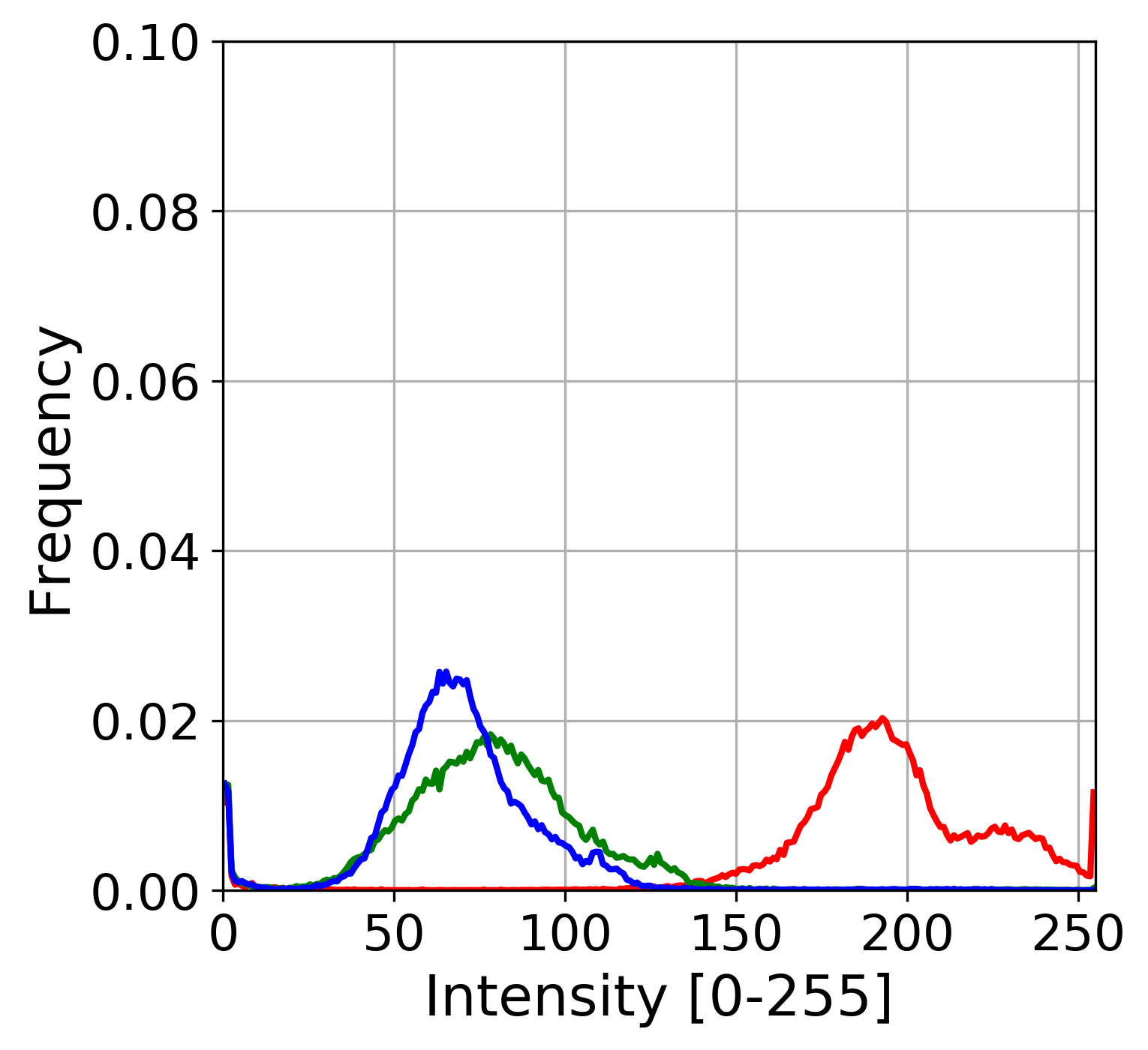}
		\end{minipage}
		
		\centering
		\begin{minipage}[b]{0.15\linewidth}
			\includegraphics[width=\linewidth]{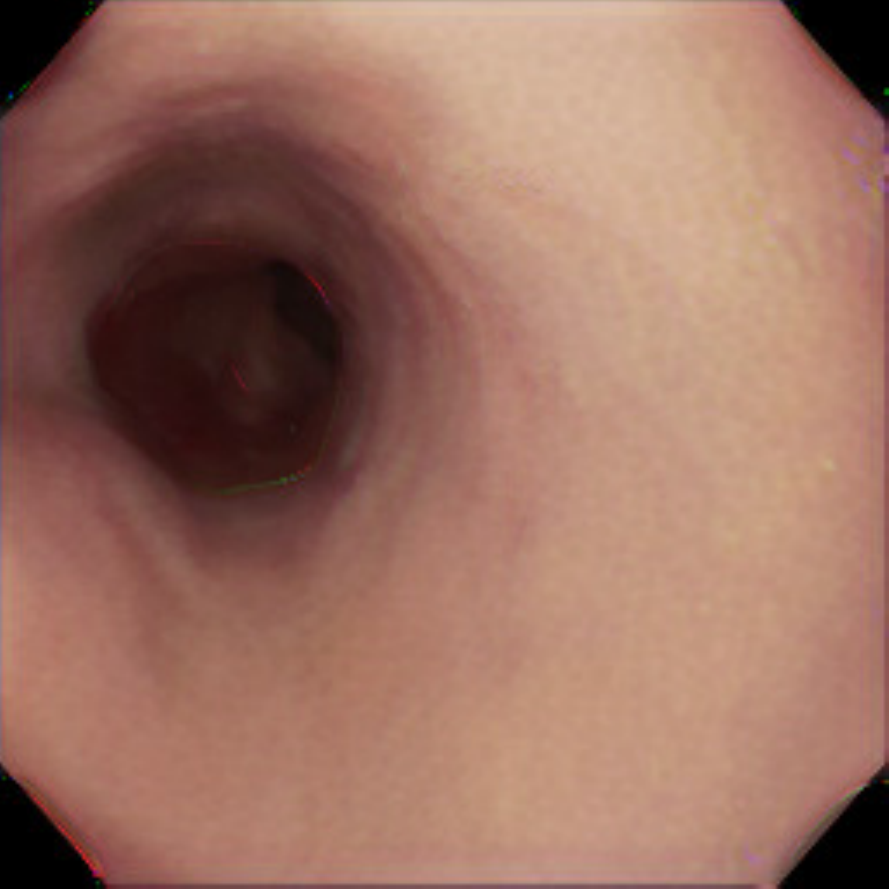}
		\end{minipage}
		\begin{minipage}[b]{0.16\linewidth}
			\includegraphics[width=\linewidth]{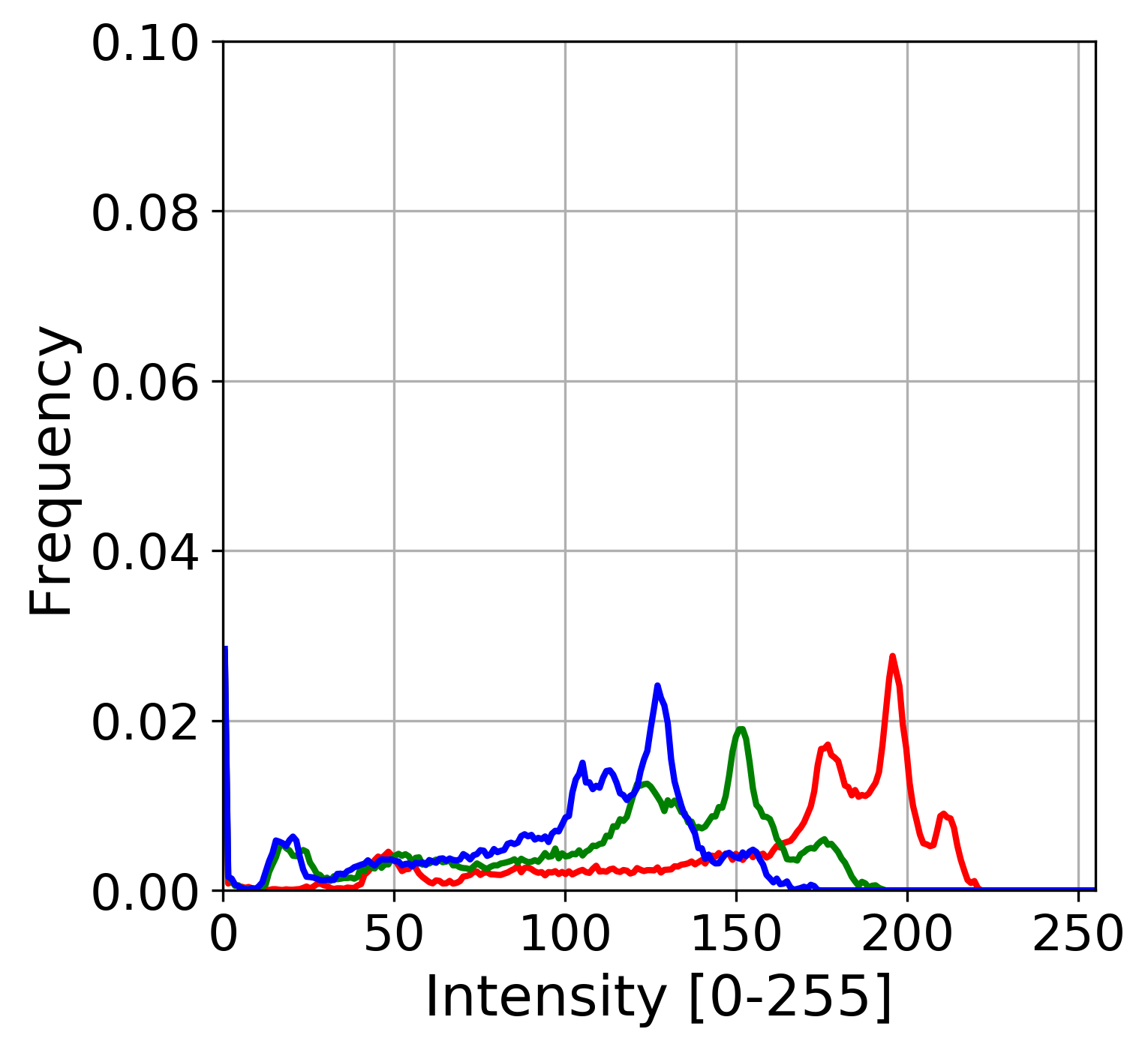}
		\end{minipage}
		\begin{minipage}[b]{0.15\linewidth}
			\includegraphics[width=\linewidth]{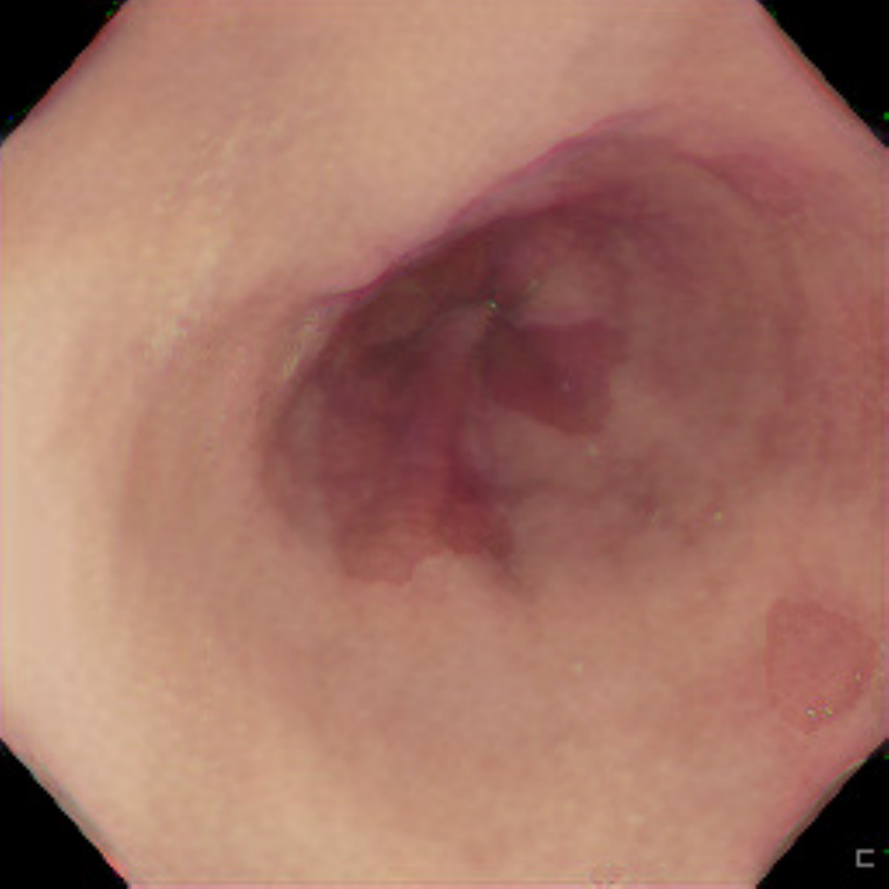}
		\end{minipage}
		\begin{minipage}[b]{0.16\linewidth}
			\includegraphics[width=\linewidth]{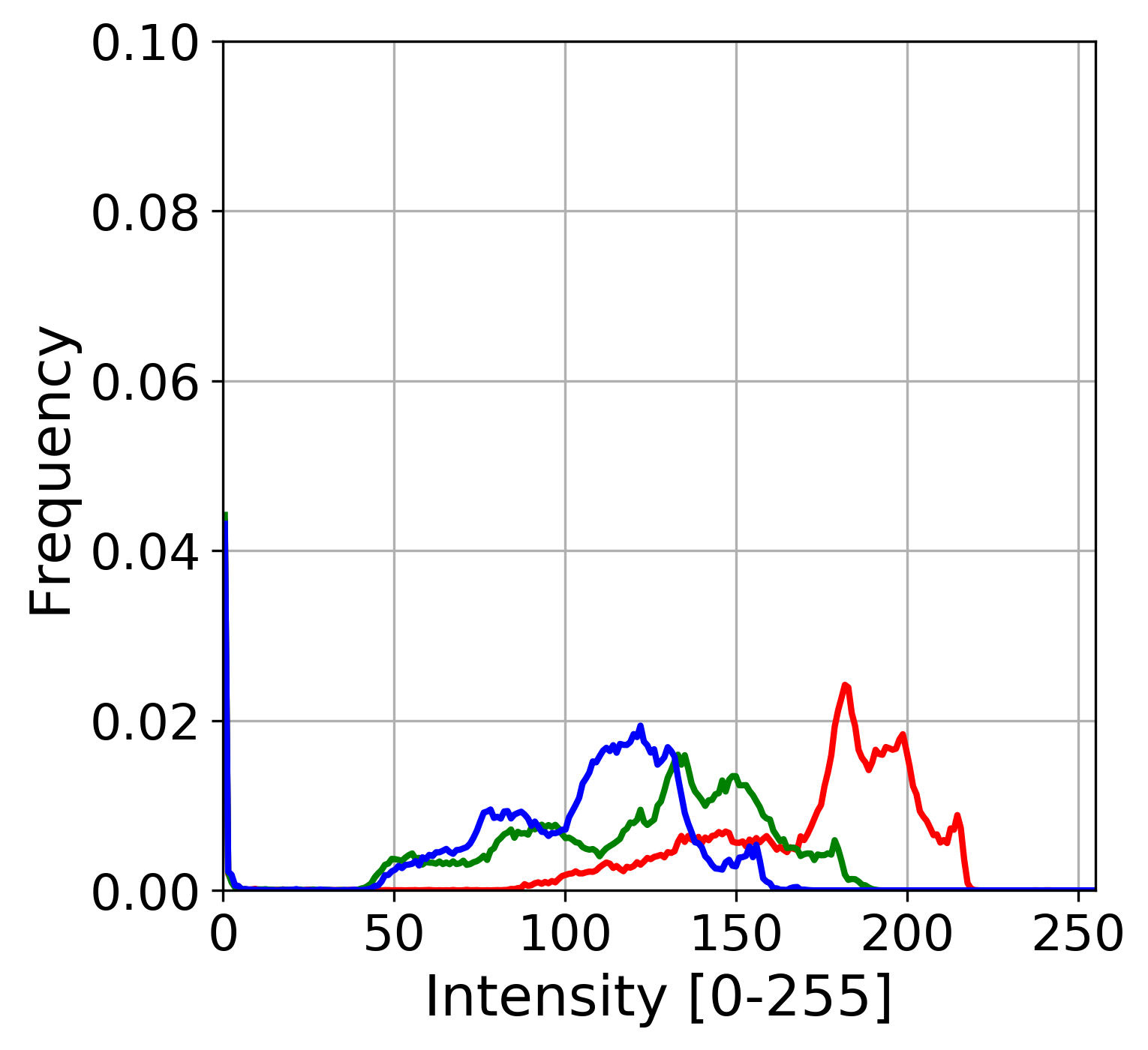}
		\end{minipage}
		\begin{minipage}[b]{0.15\linewidth}
			\includegraphics[width=\linewidth]{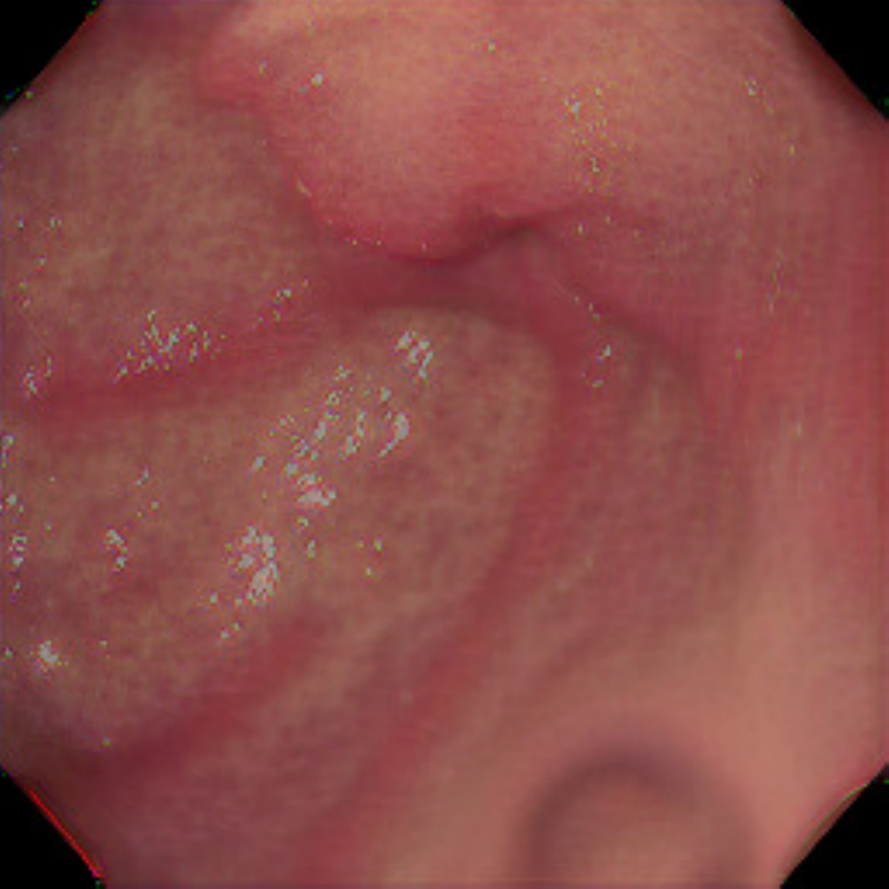}
		\end{minipage}
		\begin{minipage}[b]{0.16\linewidth}
			\includegraphics[width=\linewidth]{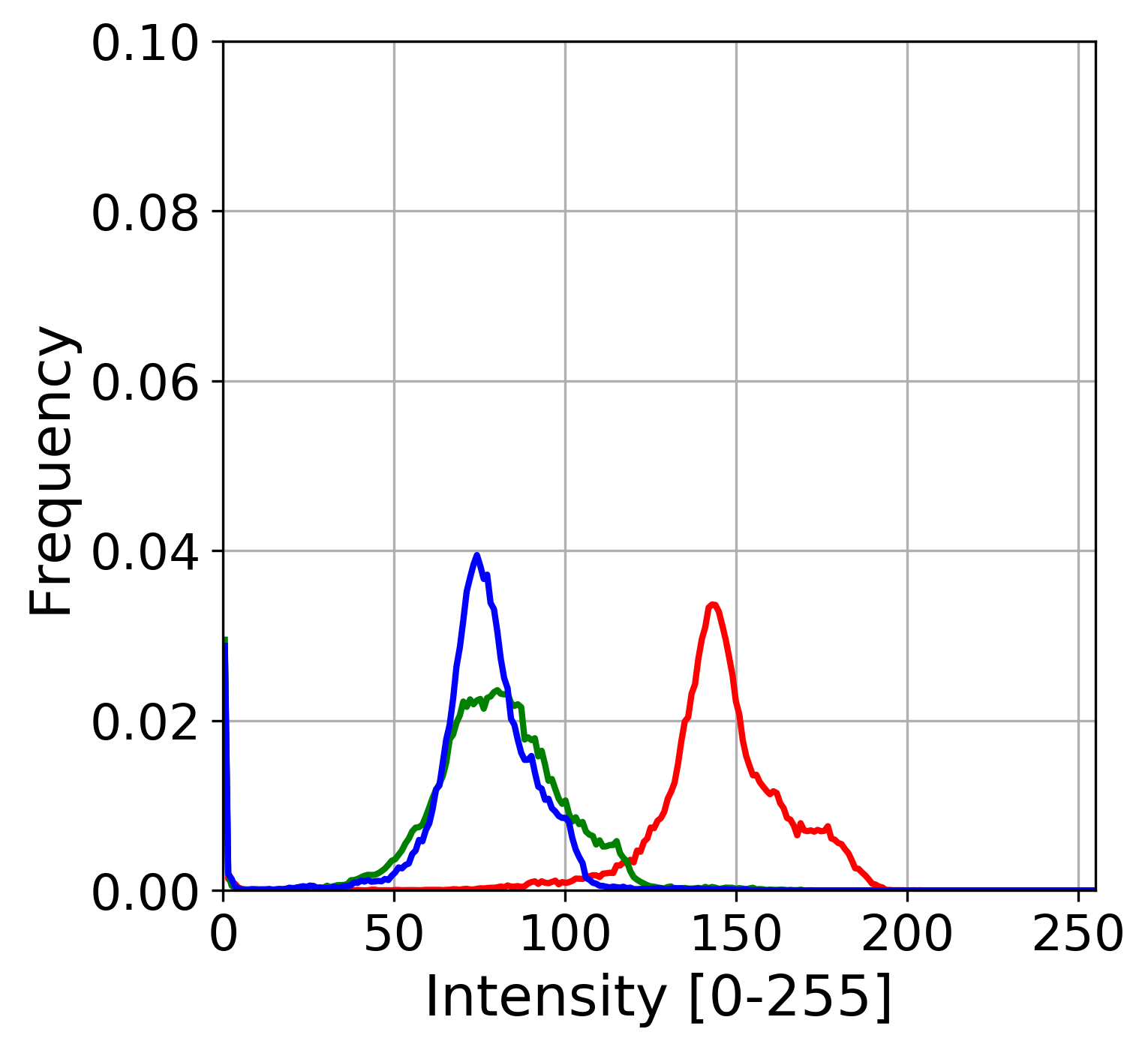}
		\end{minipage}
		
		\centering
		\begin{minipage}[b]{0.15\linewidth}
			\includegraphics[width=\linewidth]{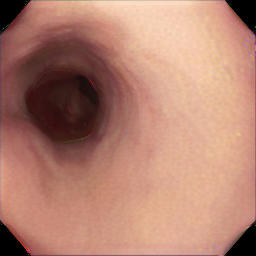}
		\end{minipage}
		\begin{minipage}[b]{0.16\linewidth}
			\includegraphics[width=\linewidth]{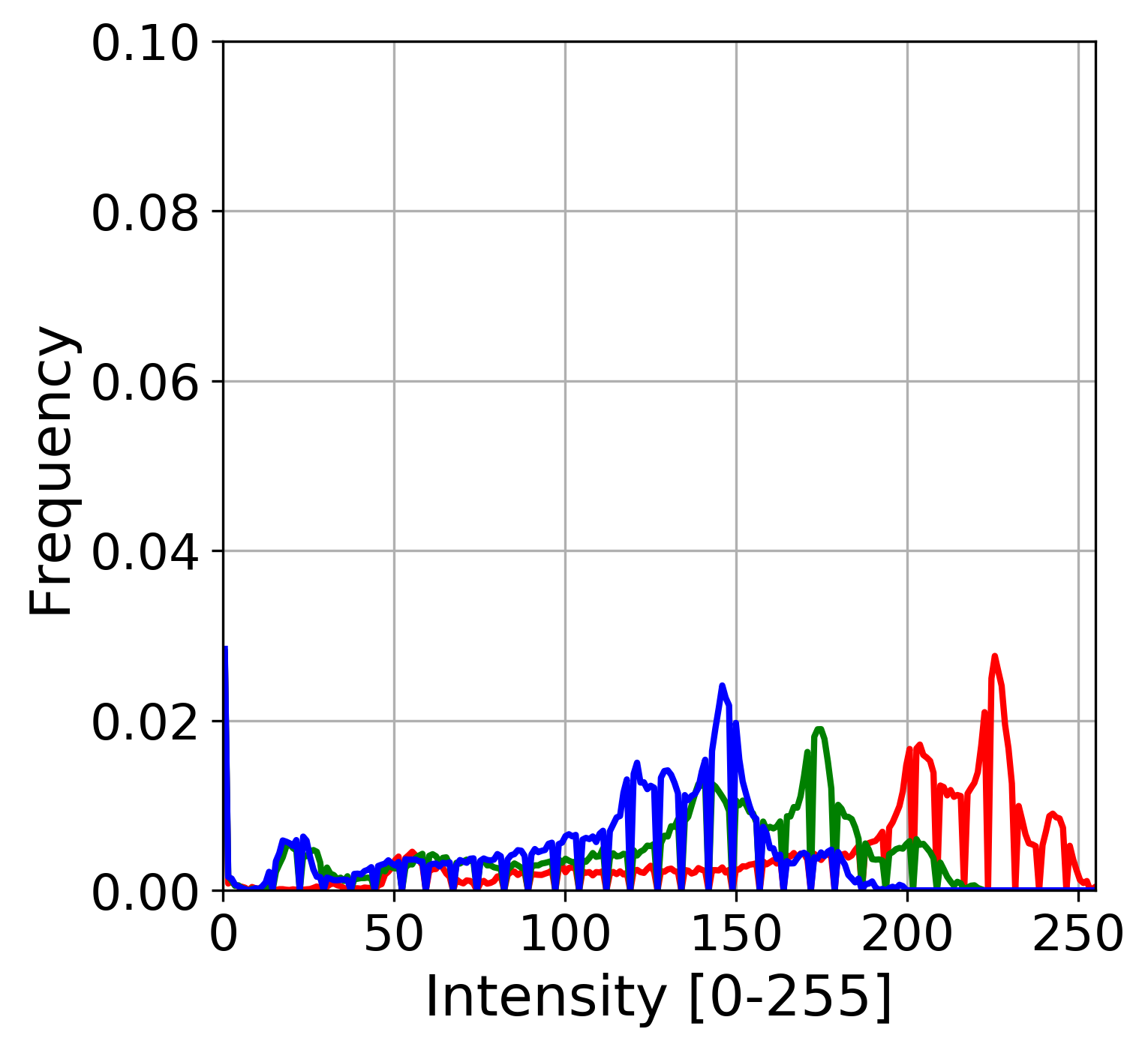}
		\end{minipage}
		\begin{minipage}[b]{0.15\linewidth}
			\includegraphics[width=\linewidth]{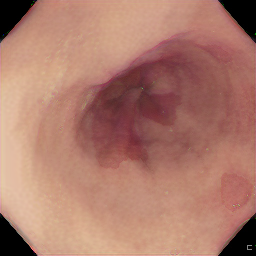}
		\end{minipage}
		\begin{minipage}[b]{0.16\linewidth}
			\includegraphics[width=\linewidth]{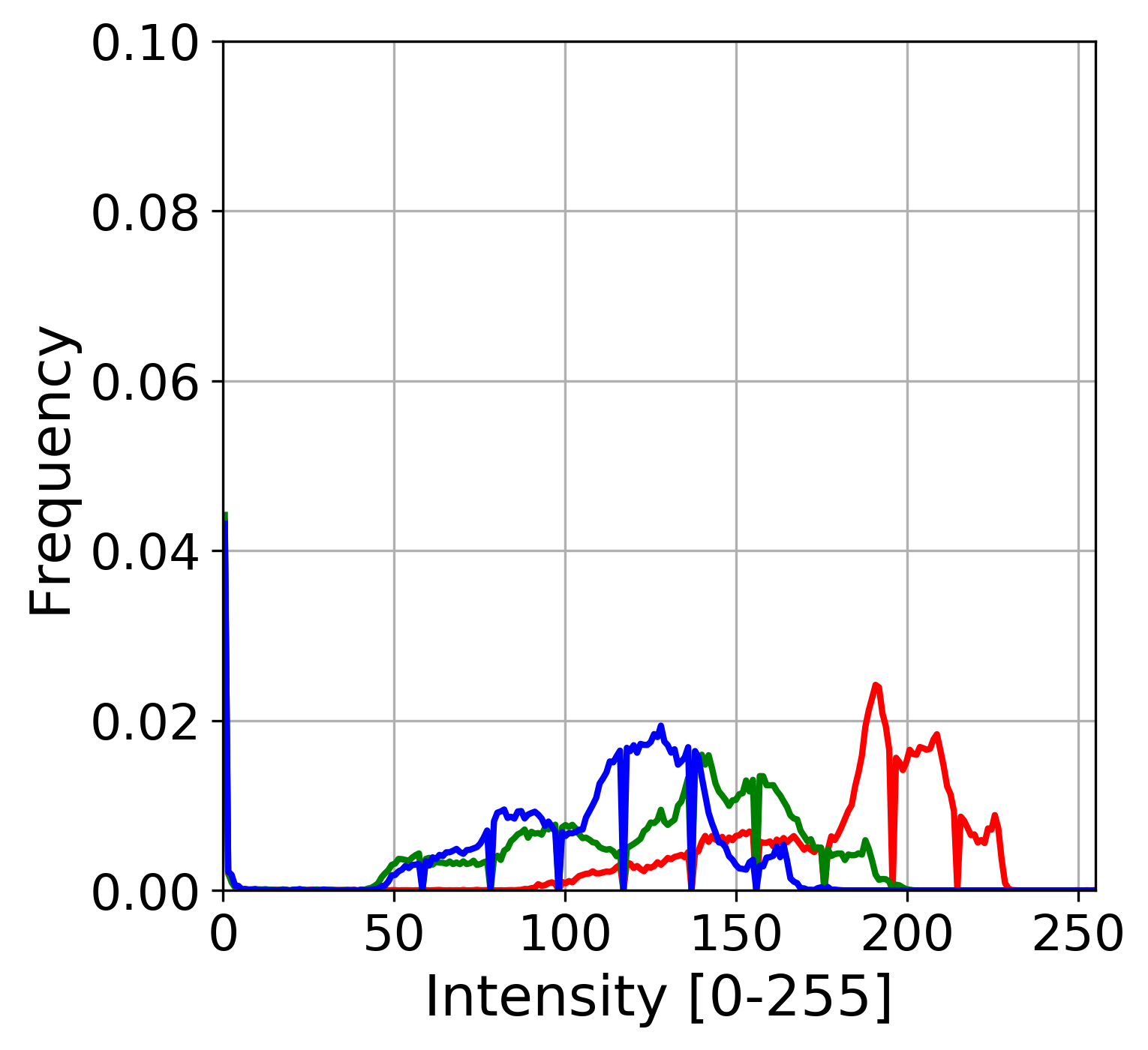}
		\end{minipage}
		\begin{minipage}[b]{0.15\linewidth}
			\includegraphics[width=\linewidth]{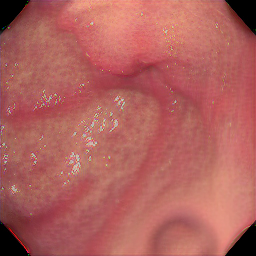}
		\end{minipage}
		\begin{minipage}[b]{0.16\linewidth}
			\includegraphics[width=\linewidth]{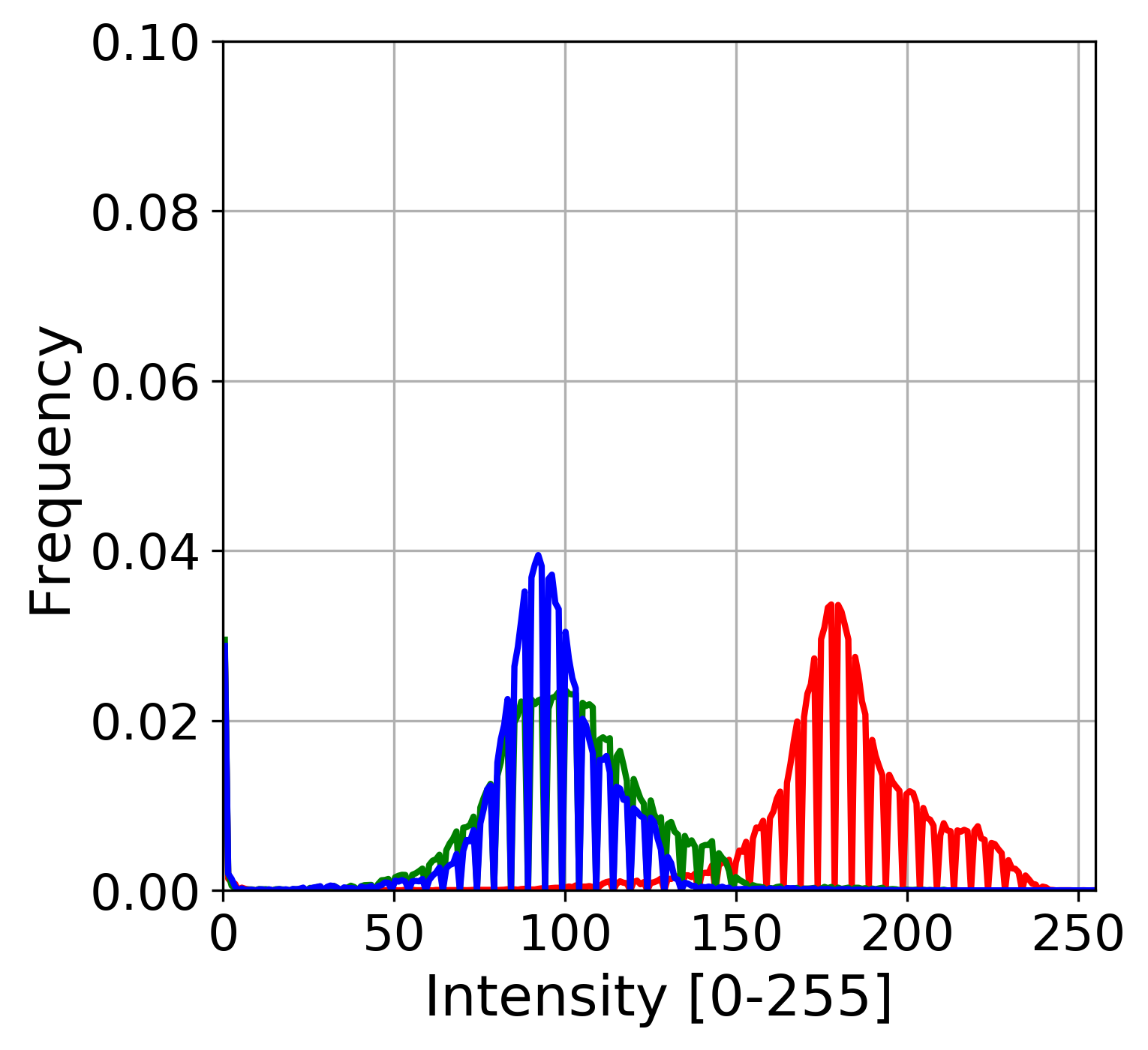}
		\end{minipage}
		
		\centering
		\begin{minipage}[b]{0.15\linewidth}
			\includegraphics[width=\linewidth]{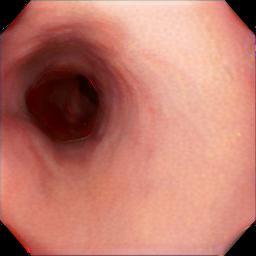}
		\end{minipage}
		\begin{minipage}[b]{0.16\linewidth}
			\includegraphics[width=\linewidth]{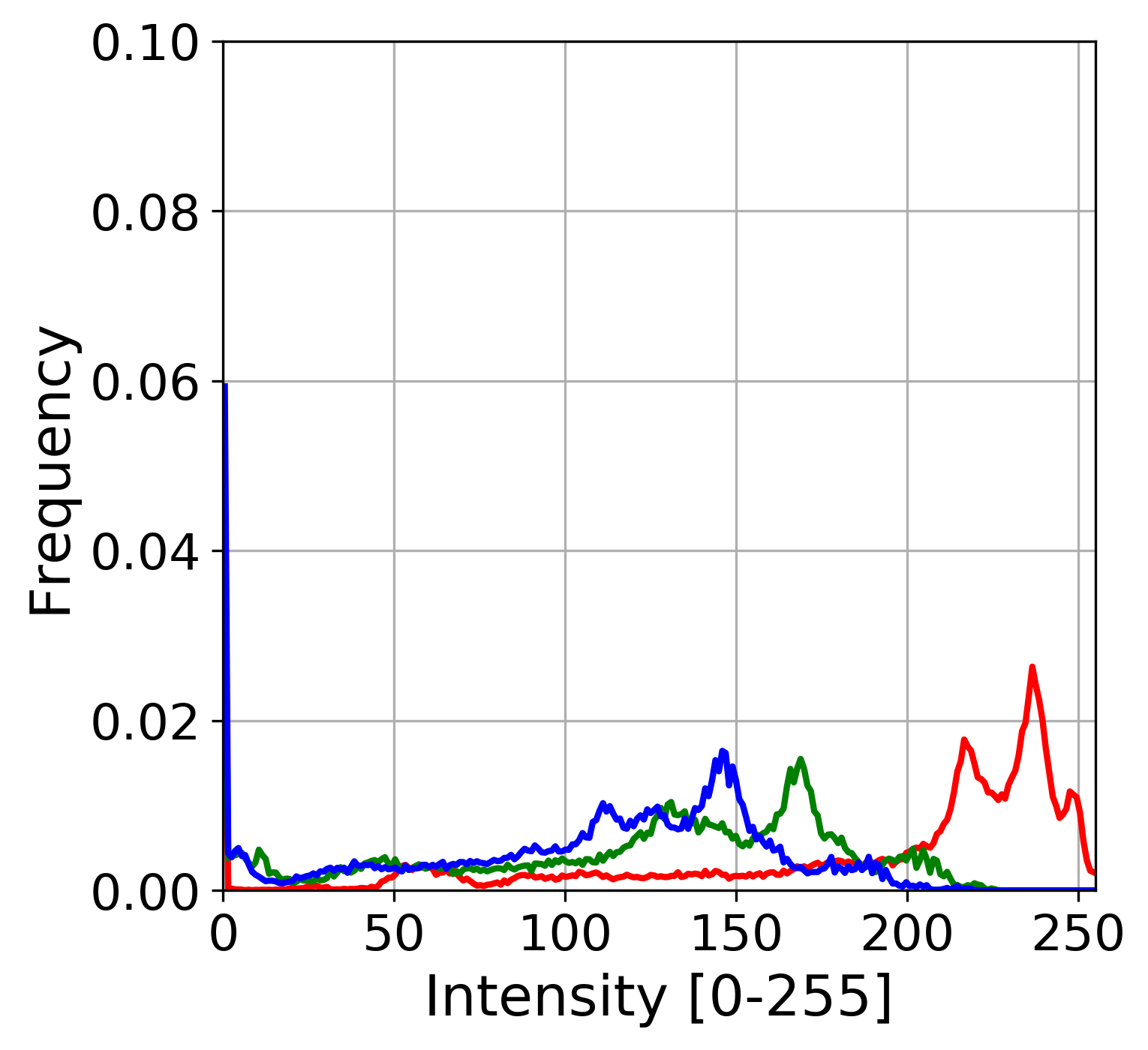}
		\end{minipage}
		\begin{minipage}[b]{0.15\linewidth}
			\includegraphics[width=\linewidth]{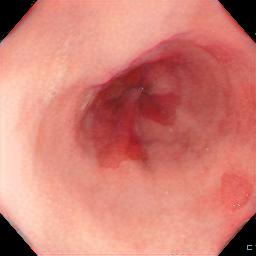}
		\end{minipage}
		\begin{minipage}[b]{0.16\linewidth}
			\includegraphics[width=\linewidth]{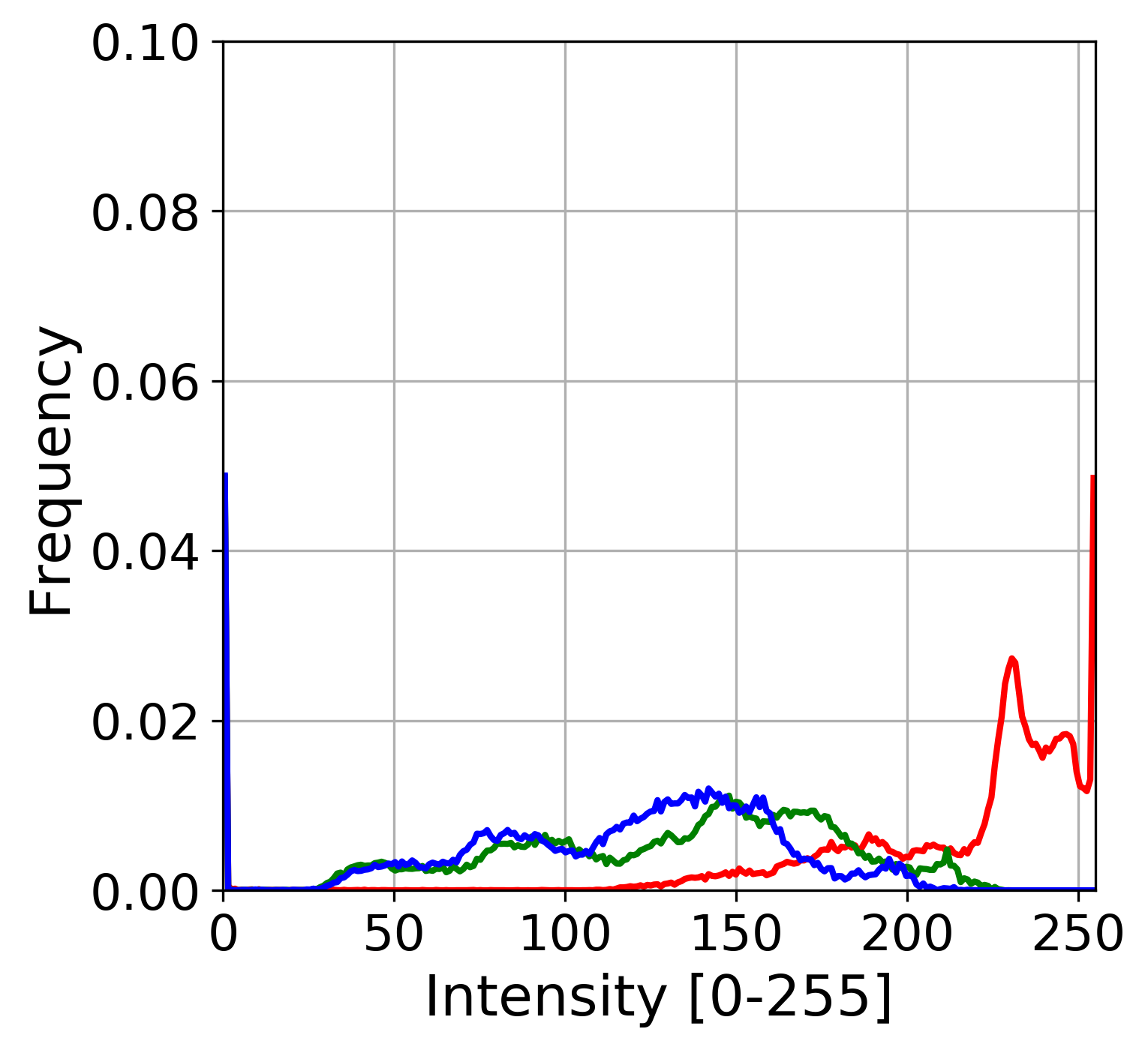}
		\end{minipage}
		\begin{minipage}[b]{0.15\linewidth}
			\includegraphics[width=\linewidth]{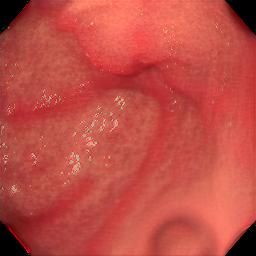}
		\end{minipage}
		\begin{minipage}[b]{0.16\linewidth}
			\includegraphics[width=\linewidth]{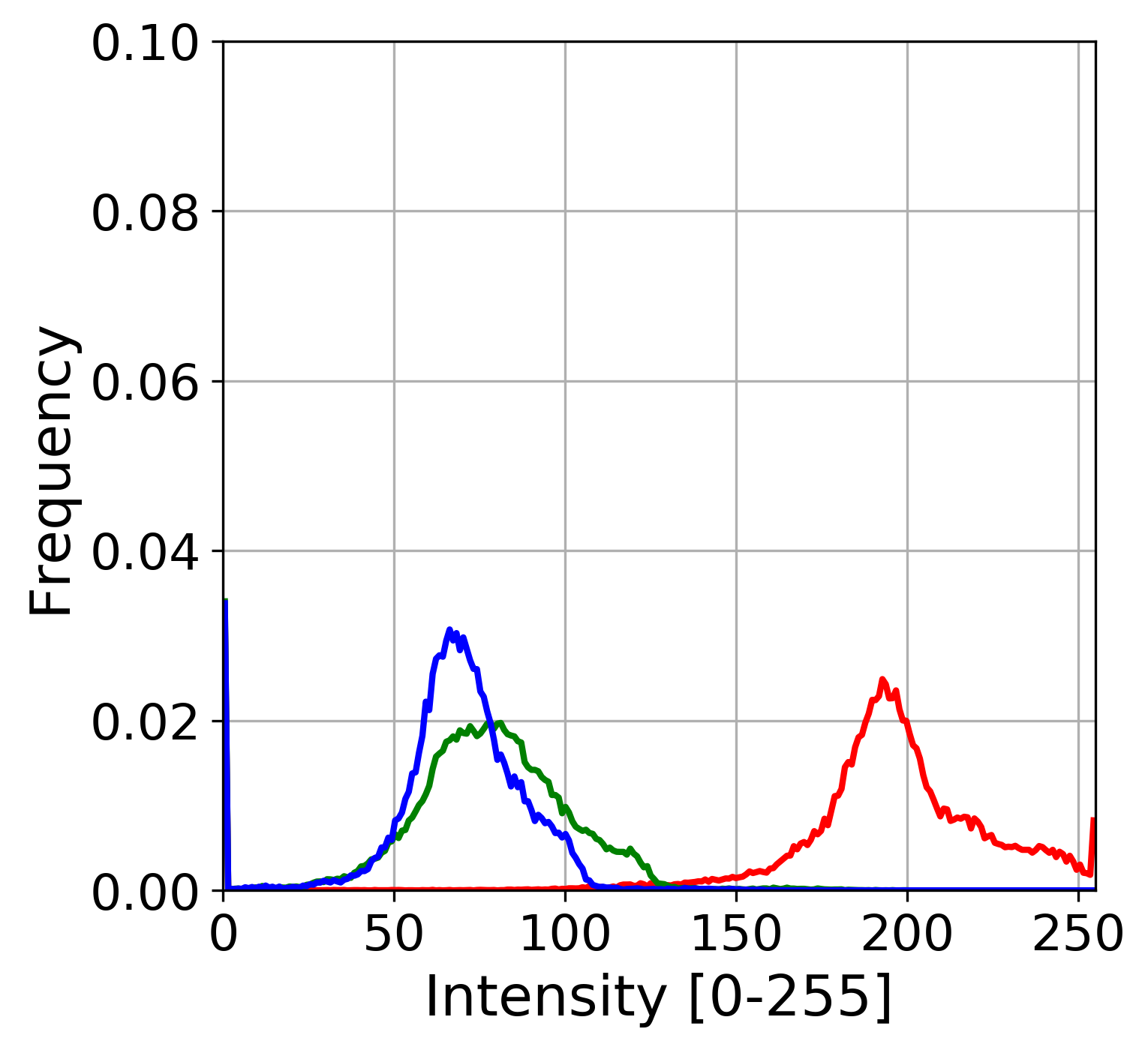}
		\end{minipage}
		\caption{Saturation and specularity correction. Left, center: Saturation of pixels in the region near to the light source (blue area, top), and right: Several specular regions in green area. Corrected images for saturation and specularity removal using trained end-to-end generator are presented on the second row. Result of simple rescaling of the corrected image intensity on the third row and result of using our color-correction instead on the last row. Corresponding RGB histograms is shown to the right of each respective image. ~\label{fig:saturationAndSpecularity}}
	\end{figure}
\end{center}

We present results of treating for saturation removal as a global problem, correcting the entire frame for over exposure as discussed in Section~\ref{subsec:saturation}. Quantitative results are provided in Table~\ref{tab:saturationSeq} for 19 randomly selected saturated frames from our simulated test data set derived from good quality frames (QS$>0.95$). Our restoration model demonstrates increased average values across all tested metrics, (PSNR, SSIM, VIF and RECO). Improvements after color transform for visual quality metrics like RECO (from 1.313 to 1.512), and VIF (from 0.810 to 0.818) illustrates boosted visual quality. This is also evident in qualitative result presented in Fig.~\ref{fig:saturationAndSpecularity}. Largely saturated image patches in the left and central frames are clearly removed by the trained generator whilst preserving the underlying image details (see RGB histograms in Fig.~\ref{fig:saturationAndSpecularity}, second row). The color transform successfully restores the original color consistency in CGAN restored images without introducing new saturation (see Fig.~\ref{fig:saturationAndSpecularity}, last row). Note that simple contrast stretching as shown in Fig.~\ref{fig:saturationAndSpecularity} (third row) by rescaling the CGAN restored frames fails to recover the original color tones. 
\begin{table}[t!]
	\centering
	\begin{tabular}{ |p{1.5cm}|p{1.0cm}|p{1.0cm}p{1.0cm}|  }
		\hline
		\textbf{Metric} &\multicolumn{3}{c|}{\textbf{QA for different stages}}\\
		\cline{2-4}
		&\multicolumn{1}{c|}{CyleGAN}&\multicolumn{1}{c|}{$l2$-contexual}&\multicolumn{1}{c|}{post-process }\\
		&\multicolumn{1}{c|}{simulation}&\multicolumn{1}{c|}{CGAN}&\multicolumn{1}{c|}{CRT}\\
		\cline{2-4}
		\hline
		\footnotesize{$\overline{PSNR}$} &\multicolumn{1}{c|}{{27.892}} &\multicolumn{1}{c|}{\textbf{28.622}} &\multicolumn{1}{c|}{{28.335}}  \\
		\hline
		\footnotesize{$\overline{SSIM}$} &\multicolumn{1}{c|}{{0.905}} &\multicolumn{1}{c|}{\textbf{0.964}} &\multicolumn{1}{c|}{{0.944}}  \\
		\hline
		\footnotesize{$\overline{VIF}$} &\multicolumn{1}{c|}{{0.808}} &\multicolumn{1}{c|}{{0.810}} &\multicolumn{1}{c|}{\textbf{0.818}}  \\
		\hline
		\footnotesize{$\overline{RECO}$} &\multicolumn{1}{c|}{{1.091}} &\multicolumn{1}{c|}{{1.313}} &\multicolumn{1}{c|}{\textbf{1.512}}  \\
		\hline
	\end{tabular}
	\vspace{0.2cm}
	\caption{Average PSNR ($\overline{PSNR}$) and average SSIM ($\overline{SSIM}$) for 19 randomly selected saturated images in our simulated data set using CycleGAN. Quality assessment (QA) for simulated images, $l2$-contexual CGAN, and post-processing using color retransfer (CRT) method are provided.}{\label{tab:saturationSeq}}
\end{table}
%
\vspace{-7mm}
\subsubsection{Specularity and other misc. artifacts removal}
Specularity and other local artifacts are removed based on inpainting (Section~\ref{subsec:specularity} for details). To validate our inpainting methods, we used a set of 25 images (clean) with randomly selected patches covering 5\% and 12\% of the total pixels of $512\times 512$ image size. We compare our CGAN-based model with $l1$-contextual loss model with widely used traditional TV-based and patch-based inpainting methods. We observe in Table.~\ref{tab:inpaintingMetric} that $l1$-contextual CGAN method has the best quality assurance values for both VIF and RECO measures (VIF: 0.95, RECO:0.992 for 5\% masked pixels and VIF: 0.883, RECO:0.983 for 12\% masked pixels). Even though the TV-based inpainting method scored higher PSNR values in both cases, it scored the least RECO values (0.984 and 0.975 respectively for 5\% and 12\% cases) and has the highest computational cost (392 seconds). In contrast, $l1$-contexual CGAN has the least computational time (2s to load the trained model and apply on images on GeForce GTX 1080 Ti). 

\begin{figure}[t!]
	\centering
	\begin{minipage}[b]{0.18\linewidth}
		\includegraphics[width=\linewidth]{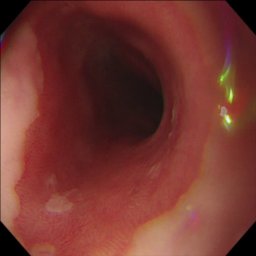}
	\end{minipage}
	\begin{minipage}[b]{0.18\linewidth}
		\includegraphics[width=\linewidth]{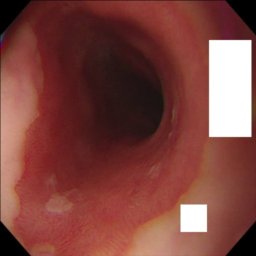}
	\end{minipage}
	\begin{minipage}[b]{0.18\linewidth}
		\includegraphics[width=\linewidth]{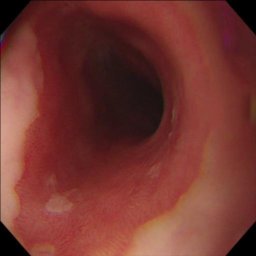}
	\end{minipage}
	\begin{minipage}[b]{0.18\linewidth}
		\includegraphics[width=\linewidth]{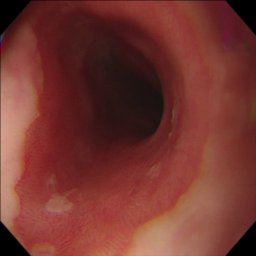}
	\end{minipage}
	\begin{minipage}[b]{0.18\linewidth}
		\begin{minipage}[b]{0.55\textwidth}
			\includegraphics[width=\linewidth]{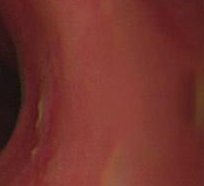}
		\end{minipage}
		\begin{minipage}[b]{0.55\textwidth}
			\includegraphics[width=\linewidth]{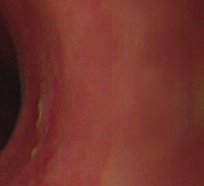}
		\end{minipage}
	\end{minipage}
	\vspace{0.05cm}
	
	\centering
	\begin{minipage}[b]{0.18\linewidth}
		\includegraphics[width=\linewidth]{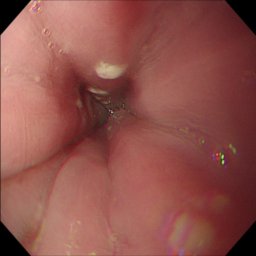}
	\end{minipage}
	\begin{minipage}[b]{0.18\linewidth}
		\includegraphics[width=\linewidth]{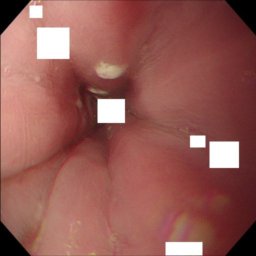}
	\end{minipage}
	\begin{minipage}[b]{0.18\linewidth}
		\includegraphics[width=\linewidth]{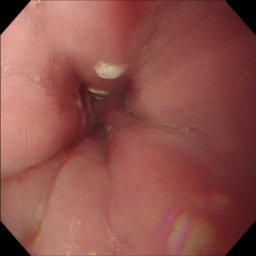}
	\end{minipage}
	\begin{minipage}[b]{0.18\linewidth}
		\includegraphics[width=\linewidth]{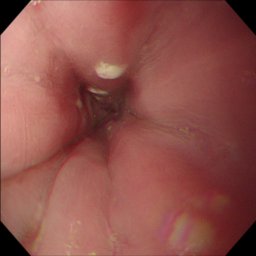}
	\end{minipage}
	\begin{minipage}[b]{0.18\linewidth}
		\begin{minipage}[b]{0.55\textwidth}
			\includegraphics[trim=0.0cm 1.0cm 0.5cm 0.0cm, clip=true,width=\linewidth]{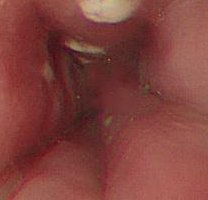}
		\end{minipage}
		\begin{minipage}[b]{0.55\textwidth}
			\includegraphics[trim=0.0cm 1.0cm 0.5cm 0.0cm, clip=true,width=\linewidth]{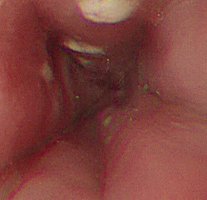}
		\end{minipage}
	\end{minipage}
	\vspace{0.05cm}
	
	\begin{minipage}[b]{0.18\linewidth}
		\includegraphics[width=\linewidth]{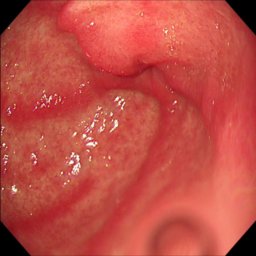}
	\end{minipage}
	\begin{minipage}[b]{0.18\linewidth}
		\includegraphics[width=\linewidth]{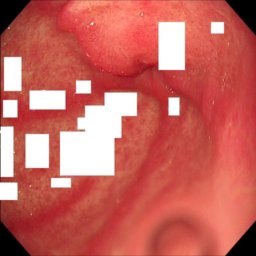}
	\end{minipage}
	\begin{minipage}[b]{0.18\linewidth}
		\includegraphics[width=\linewidth]{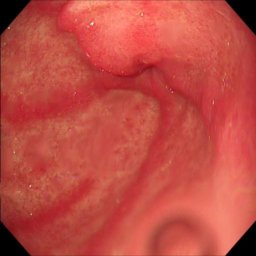}
	\end{minipage}
	\begin{minipage}[b]{0.18\linewidth}
		\includegraphics[width=\linewidth]{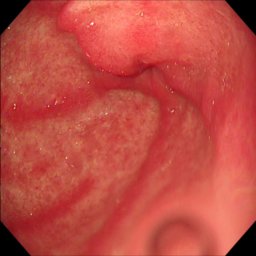}
	\end{minipage}
	\begin{minipage}[b]{0.18\linewidth}
		\begin{minipage}[b]{0.55\textwidth}
			\includegraphics[width=\linewidth,height=0.8cm]{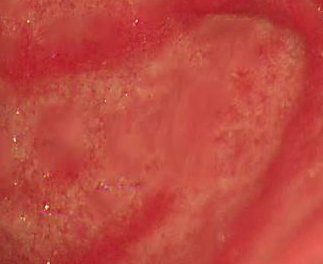}
		\end{minipage}
		\begin{minipage}[b]{0.55\textwidth}
			\includegraphics[width=\linewidth,height=0.8cm]{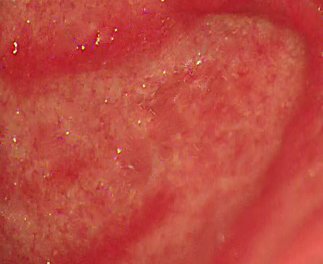}
		\end{minipage}
	\end{minipage}
	\vspace{0.05cm}

	\begin{minipage}[b]{0.18\linewidth}
		\includegraphics[width=\linewidth]{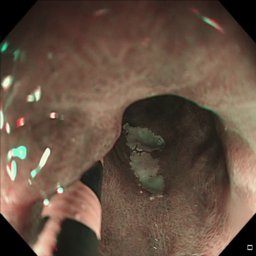}
	\end{minipage}
	\begin{minipage}[b]{0.18\linewidth}
		\includegraphics[width=\linewidth]{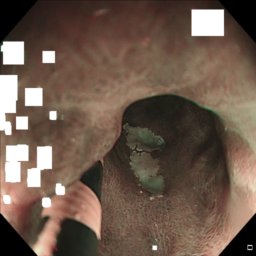}
	\end{minipage}
	\begin{minipage}[b]{0.18\linewidth}
		\includegraphics[width=\linewidth]{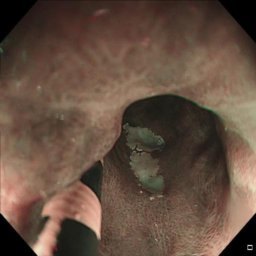}
	\end{minipage}
	\begin{minipage}[b]{0.18\linewidth}
		\includegraphics[width=\linewidth]{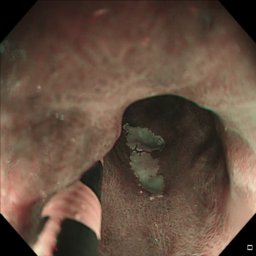}
	\end{minipage}
	\begin{minipage}[b]{0.18\linewidth}
		\begin{minipage}[b]{0.55\textwidth}
			\includegraphics[trim=0.0cm 2.0cm 0.5cm 0.0cm, clip=true,width=\linewidth,height=.8cm]{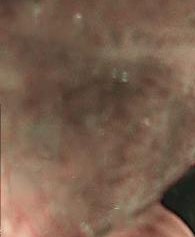}
		\end{minipage}
		\begin{minipage}[b]{0.55\textwidth}
			\includegraphics[trim=0.0cm 2.0cm 0.5cm 0.0cm, clip=true,width=\linewidth,height=0.8cm]{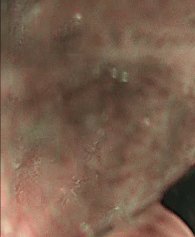}
		\end{minipage}
	\end{minipage}
	
	\begin{minipage}[b]{0.18\linewidth}
		\centerline{\footnotesize{(a)}}\medskip
	\end{minipage}
	\begin{minipage}[b]{0.18\linewidth}
		\centerline{\footnotesize{(b)}}\medskip
	\end{minipage}
	\begin{minipage}[b]{0.18\linewidth}
		\centerline{\footnotesize{(c)}}\medskip
	\end{minipage}
	\begin{minipage}[b]{0.18\linewidth}
		\centerline{\footnotesize{(d)}}\medskip
	\end{minipage}
	\begin{minipage}[b]{0.18\linewidth}
		\hspace{-5mm}
		\centerline{\footnotesize{(e)}}\medskip
	\end{minipage}
	\caption{Image restoration result using inpainting of corrupted areas (specularity, imaging artifacts) detected by our detection method. a) Original corrupted image, b) detected bounding boxes, c) inpainting result using recent TV-based method, d) l1-contexual CGAN, e) top, bottom: Restored area marked with blue rectangle in (b) using TV-based and generative model using l1-contexual CGAN, respectively.~\label{fig:Specularity}}
\end{figure}

Qualitative results for our specularity and local artifact removal on real problematic gastro-oesophageal endoscopic frames are shown in Fig.~\ref{fig:Specularity}. In Fig.~\ref{fig:Specularity} (a), both imaging artifacts (first and fourth rows) and specularities (second and third rows) introduce large deviations in pixel intensities both locally with respect to neighouring pixels and globally with respect to the uncorrupted image appearance.
Using inpainting methods (see Fig.~\ref{fig:Specularity} (c) and (d)), the images have been restored based on the bounding box detections of our artifact detector. The second best TV-based method in Fig.~\ref{fig:Specularity}(c) produces blurry and non-smooth patches during the reconstruction of unknown pixels (refer to Fig.~\ref{fig:Specularity} (b)) compared to CGAN generative model (see Fig.~\ref{fig:Specularity}(d)). A closer look around the unknown regions indicated by blue rectangular boxes in Fig.~\ref{fig:Specularity}(b), Fig.~\ref{fig:Specularity}(e) shows that local image structures are well preserved and smoother transition from reconstructed pixels to the surrounding pixels is present. An immediate noticeable ghost effect can be observed in the second row, Fig.~\ref{fig:Specularity}(e) top using the TV-based method. 
\begin{table*}[t!b!]
	\centering
	\begin{tabular}{ |p{1cm}|p{0.6cm}|p{0.3cm}p{0.3cm}|p{0.3cm}|p{.3cm}p{0.3cm}|p{.05cm}| }
		\hline
		\textbf{Method} & \multicolumn{3}{c|}{\textbf{5\% of total pixels}}  & \multicolumn{3}{|c|}{\textbf{12\% of total pixels}} &  \multicolumn{1}{c|}{$\overline{t}$}\\
		\cline{2-8}
		& {\footnotesize{PSNR}}&\multicolumn{1}{c|}{\footnotesize{VIF}}&\multicolumn{1}{c|}{\footnotesize{RECO}}& \multicolumn{1}{|c|}{\footnotesize{PSNR}}&\multicolumn{1}{c|}{\footnotesize{VIF}}&\multicolumn{1}{c|}{\footnotesize{RECO}} & \multicolumn{1}{c|}s\\
		\hline
		TV-based~\cite{ipoltvdc} & \textbf{45.130} &\multicolumn{1}{c|}{0.947}&\multicolumn{1}{c|}{0.984}& \multicolumn{1}{|c|}{\textbf{40.970}} &\multicolumn{1}{c|}{0.881}&\multicolumn{1}{c|}{0.975} &\multicolumn{1}{c|} {392.0} \\
		\hline
		Patch-based~\cite{Alasdair:IPOL17} & 43.440&\multicolumn{1}{c|}{0.940}&\multicolumn{1}{c|}{0.990}& \multicolumn{1}{|c|}{39.520} &\multicolumn{1}{c|}{0.871}&\multicolumn{1}{c|}{0.984}&\multicolumn{1}{c|} {35.0}\\
		\hline
		$l1$-cont. CGAN & 43.487&\multicolumn{1}{c|}{\textbf{0.950}}&\multicolumn{1}{c|}{\textbf{0.992}}& \multicolumn{1}{|c|}{39.693} &\multicolumn{1}{c|}{\textbf{0.883}}&\multicolumn{1}{c|}{\textbf{0.983}} &\multicolumn{1}{c|}{\textbf{2.5}}\\
		\hline
	\end{tabular}
	\vspace{0.5cm}
	\caption{Average values for PSNR, VIF~\cite{Sheikh:TIP06}, and RECO~\cite{Baroncini:ESPC09}) metrics for restoration of missing pixels for masks covering 5\% and 12\% of total image pixels ($512\times 512$ pixels) with 21 randomly sampled rectangular boxes on 20 randomly selected images from 3 different patient videos. }{\label{tab:inpaintingMetric}}
\end{table*}
%
\subsection{Video recovery and quality assessment}

\begin{figure}[t!]
	\centering
	\begin{minipage}[b]{\linewidth}
		\centering
		\includegraphics[width=0.5\linewidth]{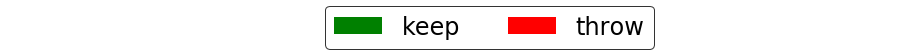}
	\end{minipage}
	\begin{minipage}[b]{.75\linewidth}
		\includegraphics[width=\linewidth]{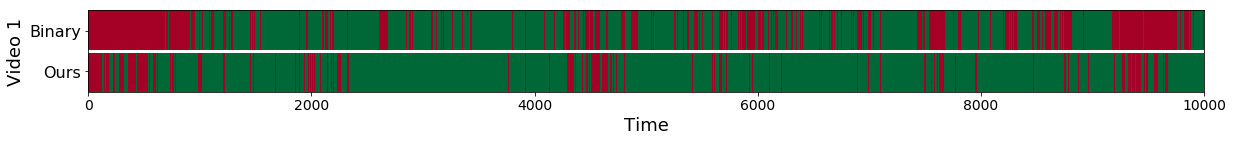}
		\includegraphics[width=\linewidth]{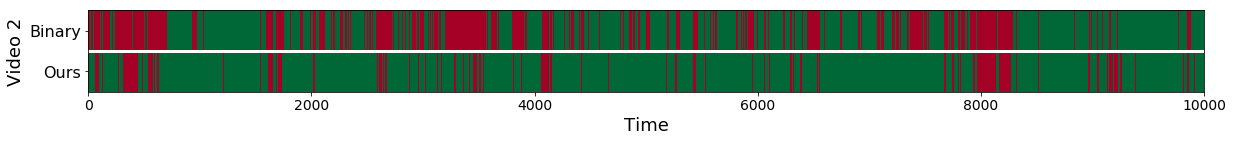}
		\includegraphics[width=\linewidth]{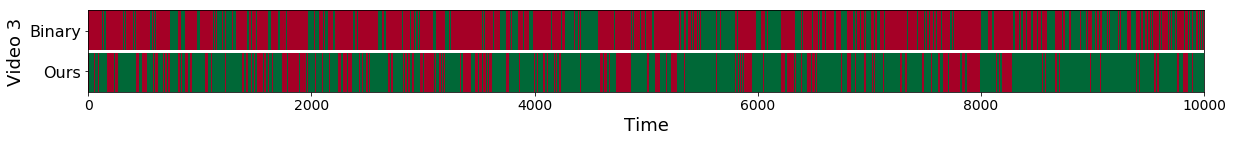}
	\end{minipage}
	\begin{minipage}[b]{.21\linewidth}
		\centering
		\includegraphics[width=\linewidth]{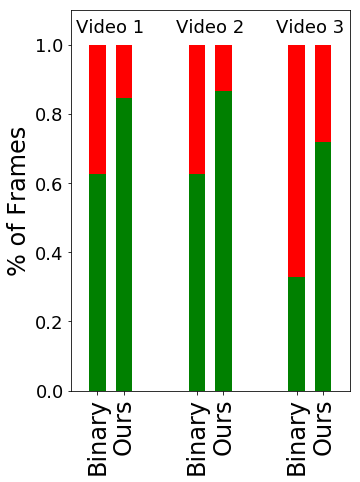}
	\end{minipage}
	
	\begin{minipage}[b]{0.32\linewidth}
		\includegraphics[width=\linewidth]{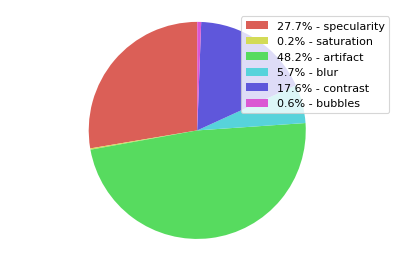}
		\centerline{\footnotesize{Video 1}}\medskip
	\end{minipage}
	\begin{minipage}[b]{0.32\linewidth}
		\includegraphics[width=\linewidth]{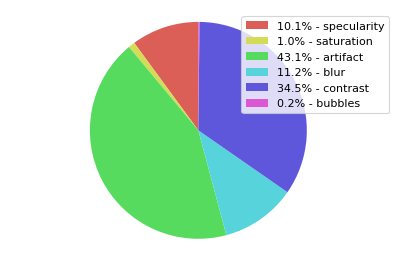}
		\centerline{\footnotesize{Video 2}}\medskip
	\end{minipage}
	\begin{minipage}[b]{0.32\linewidth}
		\includegraphics[width=\linewidth]{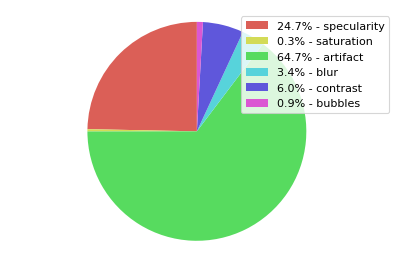}
		\centerline{\footnotesize{Video 3}}\medskip
	\end{minipage}
	\caption{Frame recovery in clinical endoscopy videos. Top: frames and proportion deemed recoverable over a sequence of  and the over a sequence of using a binary deep classifer and our proposed QS score. Bottom: the proportion of each artifact type present in each video.}
	\label{fig:artifact_detection_recovery_results}
\end{figure}

We evaluated our artifact detection and recovery framework on 10 gastroesophageal videos comprising with nearly 10,000 frames each. For artifact detection, an objectness threshold of 0.25 was used to reduce duplication in detected boxes and QS value for restoring the frame was set to $ \geq 0.5$. As a baseline, we also separately trained a sequential 6-layer convolution neural network (layer with 64 filters of sizes ${3\time 3, 5\times 5}$, ReLU activation function and batch normalization) with a fully connected last layer for binary classification on a set of 6000 manually labeled positive and negative images to decide whether to discard or keep a given input video frame. A threshold of 0.75 was set for the binary classifier to keep only frames of sufficient quality. Our framework successfully retains the vast majority of frames compared to a binary decision, Fig.\ref{fig:artifact_detection_recovery_results}. The quality enhanced video was again fed to our CNN-based binary classifier which resulted in lower number of frame rejection than on raw videos. Consequently, the resultant video is more continuous compared to the equivalent binary cleaned video utilizing raw videos. For example, in video 3, the video after frame removal based on the binary classifier directly lead to many distinct abrupt transitions that can be detrimental for post-processing algorithms as only 30\% is kept. Comparatively, our proposed framework retains 70\% of frames, i.e. a frame restoration of nearly 40\%. Quantitatively across all 10 endoscopic videos tested, our framework restored 25\% more video frames, retaining on an average of 68.7\% of 10 videos considered.
%
%
%
%
\subsection{Clinical relevance test}
We corrupted 20 high-quality images selected from 10 test videos with blur, specularity, saturation and misc. artifacts (refer Sec.~\ref{sec:restoration_sec}). Restoration methods were then applied to these images. Two expert endoscopists independently were asked to score these restoration results compared to the original high-quality images and corresponding videos. Scores (0-10) were based on: 1)  Addition of unnatural distortions was assigned a negative score and 2) removal of distortions was assigned a positive score. The obtained mean score were blur: 7.87, specularity or misc. artifacts: 7.7, and saturation: 1.5. A remarkable restoration was obtained for blur and specularity or misc. artifacts. However, saturation correction was not pleasant to experts mostly due to loss of 3D information (according to feedback comments) even though visual coherence was improved.
%
%
%
\section{Conclusion}
\label{sec:conc}
We have presented a novel end-to-end framework for the detection and restoration of inevitable endoscopic video frame artifacts which are embodied in frames linearly, non-linearly or both. Our contribution includes artifact-specific quality assessment and sequential restoration. In particular, as each module in the proposed framework is formulated as a neural network, our framework can fully take advantage of the real-time processing capabilities of modern GPUs. We have proposed several novel techniques for frame restoration that includes an edge-based (high-frequency) loss for recovering blurred images and a color re-transfer scheme to deal with color shifts in generated frames due to model transfer. We have proposed novel regularization schemes based on each artifact class-type for frame restoration yielding high-quality image generation. Through extensive experiments we have validated each step of our framework, from detection to restoration methods. We achieved the highest $mAP_5$ and $mAP_{25}$ with our modulated YOLOv3-spp and the least inference time (88~ms) for real time frame quality scoring. We demonstrated quantitative and qualitative improvements for frame restoration tasks. Notably, we achieved improvements in both PSNR and SSIM metrics for blur and saturation using our proposed models. For specularity and other misc. artifacts removal, we achieved also significant improvements on visual similarity metrics. Finally, our sequential approach was able to restore an average of 25\% of the video frames in 10 randomly selected videos from our database. It is worth noting that for 3 videos used for illustration of the importance of our proposed framework, 40\% of frames which otherwise would be discarded for downstream analysis were rescued. We demonstrated high quality performance on real clinical endoscopy videos for both intra- and inter-patient variabilities and multimodality. Future work will focus on further improving the object detection network and implementing the entire framework as a single end-to-end trainable neural network.     
\bibliographystyle{ieee}
\bibliography{tmi_Ali_Zhou_arXIV_version}
\end{document}